\pdfoutput=1

\documentclass[11pt]{article}

\usepackage[table]{xcolor}
\usepackage{acl}

\usepackage{times}
\usepackage{latexsym}

\usepackage[T1]{fontenc}

\usepackage[utf8]{inputenc}

\usepackage{microtype}

\usepackage{color}
\usepackage{longtable}
\usepackage{arydshln}
\usepackage{booktabs}
\usepackage{graphicx}
\usepackage{subcaption}
\usepackage{float}
\usepackage{amsmath,amsfonts,amssymb,bbm,epsfig,bm}
\usepackage[ruled,noresetcount,linesnumbered]{algorithm2e}
\usepackage{listings}
\usepackage{url}
\usepackage{spverbatim}
\usepackage[frozencache,cachedir=minted-cache]{minted}         %
\usepackage{multirow}
\usepackage{pbox}
\usepackage{tikz}
\usepackage{pifont}
\usepackage{xspace}
\usepackage{physics}
\usepackage[nameinlink,capitalize]{cleveref}
\usepackage{scalerel}
\usepackage[font=small]{caption}
\usepackage{titlesec}
\usepackage{todonotes}
\usepackage[normalem]{ulem}
\usepackage{sidecap}
\usepackage{comment}
\usepackage{tablefootnote}
\usepackage{multicol}
\usepackage{xpatch}  %
\usepackage{tcolorbox}

\newenvironment{cite-comment}{}{}
\tcbuselibrary{skins}
\tcolorboxenvironment{cite-comment}{empty,
  left = 1em, top = 1ex, bottom = 1ex,
  borderline west = {2pt} {0pt} {black!20},
}

\setminted{fontsize=\scriptsize}
\newenvironment{code}{\captionsetup{type=listing}}{}

\titlespacing{\paragraph}{%
  0pt}{%
  0.2\baselineskip}{%
  1em}%

\renewcommand{\tt}[1]{\fontfamily{cmtt}\selectfont #1}

\makeatletter
\Crefname{algorithm}{Algo.}{Algorithms}
\Crefname{table}{Tab.}{Tables}
\crefname{section}{\S\@gobble}{\S\S\@gobble}
\crefname{subsection}{\S\@gobble}{\S\S\@gobble}
\makeatother

\newcommand{\vcenteredinclude}[1]{\begingroup
\setbox0=\hbox{\includegraphics[height=1.0em]{#1}}%
\parbox{\wd0}{\box0}\endgroup}

\newcommand{\xmark}{\vcenteredinclude{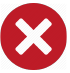}}
\newcommand{\cmark}{\vcenteredinclude{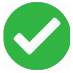}}

\newcommand{\passat}[1]{\textit{pass}@\ensuremath{#1}}
\def\dataset/{\textsc{ARCADE}}
\def\incoder/{\textsc{InCoder}}
\def\codegen/{\textsc{CodeGen}}
\def\palm/{\textsc{PaLM}}
\def\nbmodel/{\textsc{PaChiNCo}}

\def\vc/{$\mathrm{Vanilla~Code}$}
\def\sbs/{$\mathrm{Step}\textrm{-}\mathrm{by}\textrm{-} \mathrm{Step}$}
\def\sbshort/{$\mathrm{SbS}$}

\newcommand\codegensym[3]{\codegen/\ensuremath{_\mathrm{#1}}~\ensuremath{#2\mathrm{#3}}}

\def\numproblem/{$1,082$}  %
\def\numdataset/{$106$}
\def\numnb/{$133$}

\newcommand{\eg}{\hbox{\emph{e.g.}}\xspace}
\newcommand{\ie}{\hbox{\emph{i.e.}}\xspace}

\newcommand\intent{\ensuremath{\bm{u}}}

\newcommand\cell{\ensuremath{c}}

\newcommand\utterance[1]{\textit{#1}}
\newcommand{\ablation}[1]{\raisebox{0.5ex}{\;$\llcorner\,$}#1}
\def\existtasks/{\textsc{PaLM}}
\def\existtasks/{\emph{Existing Tasks}}
\def\newtasks/{\emph{New Tasks}}

\newcommand\adds[1]{\ensuremath{\color{gray} \scriptstyle + #1}}
\newcommand\mi[1]{\ensuremath{\color{gray} \scriptstyle - #1}}

\makeatletter
\AtBeginEnvironment{minted}{\dontdofcolorbox}
\def\dontdofcolorbox{\renewcommand\fcolorbox[4][]{##4}}
\xpatchcmd{\inputminted}{\minted@fvset}{\minted@fvset\dontdofcolorbox}{}{}
\xpatchcmd{\mintinline}{\minted@fvset}{\minted@fvset\dontdofcolorbox}{}{} %
\makeatother

\makeatletter

\newcommand*{\tabminted@finalstrut}[1]{%
  \ifdim\prevdepth>0pt
    \ifdim\dp#1>\prevdepth
      \vskip\dimexpr(\dp#1)-\prevdepth\relax
    \fi
  \else
    \vskip\dimexpr(\dp#1)\relax
  \fi
}
\newcommand*{\@tabmintedend}{%
  \let\@finalstrut\tabminted@finalstrut
}
\makeatother

\title{Natural Language to Code Generation in Interactive \\ Data Science Notebooks}

\author{
    Pengcheng Yin\thanks{~~Correspondence to {\tt pcyin@google.com}},~\quad
    Wen-Ding Li,\quad
    Kefan Xiao,\quad
    Abhishek Rao,\quad
    Yeming Wen,\quad \\
    \textbf{Kensen Shi,}\quad
    \textbf{Joshua Howland,}\quad
    \textbf{Paige Bailey,}\quad
    \textbf{Michele Catasta,}\quad \\
    \textbf{Henryk Michalewski,}\quad
    \textbf{Alex Polozov,}\quad
    \textbf{Charles Sutton}\quad \\
    \vspace{-2mm}\\
    Google Inc. 
}

\begin{document}
\maketitle
\begin{abstract}
Computational notebooks, such as Jupyter notebooks, are interactive computing environments
that are ubiquitous among data scientists to perform data wrangling and analytic tasks.
To measure the performance of AI pair programmers that automatically synthesize programs for those tasks given natural language (NL) intents from users, we build \dataset/, a benchmark of \numproblem/ code generation problems using the {\tt pandas} data analysis framework in  data science notebooks.
\dataset/ features multiple rounds of NL-to-code problems from the same notebook. It requires a model to understand rich multi-modal contexts, such as existing notebook cells and their execution states as well as previous turns of interaction.
To establish a strong baseline on this challenging task, we develop \nbmodel/, a 62B code language model (LM) for Python computational notebooks, which significantly outperforms public code LMs.
Finally, we explore few-shot prompting strategies to elicit better code with step-by-step decomposition and NL explanation, showing the potential to improve the diversity and explainability of model predictions.\footnote{Dataset will be released shortly.}
\end{abstract}

\section{Introduction}

\begin{figure}[t!]
    \centering
    \small
    \includegraphics[width=\columnwidth]{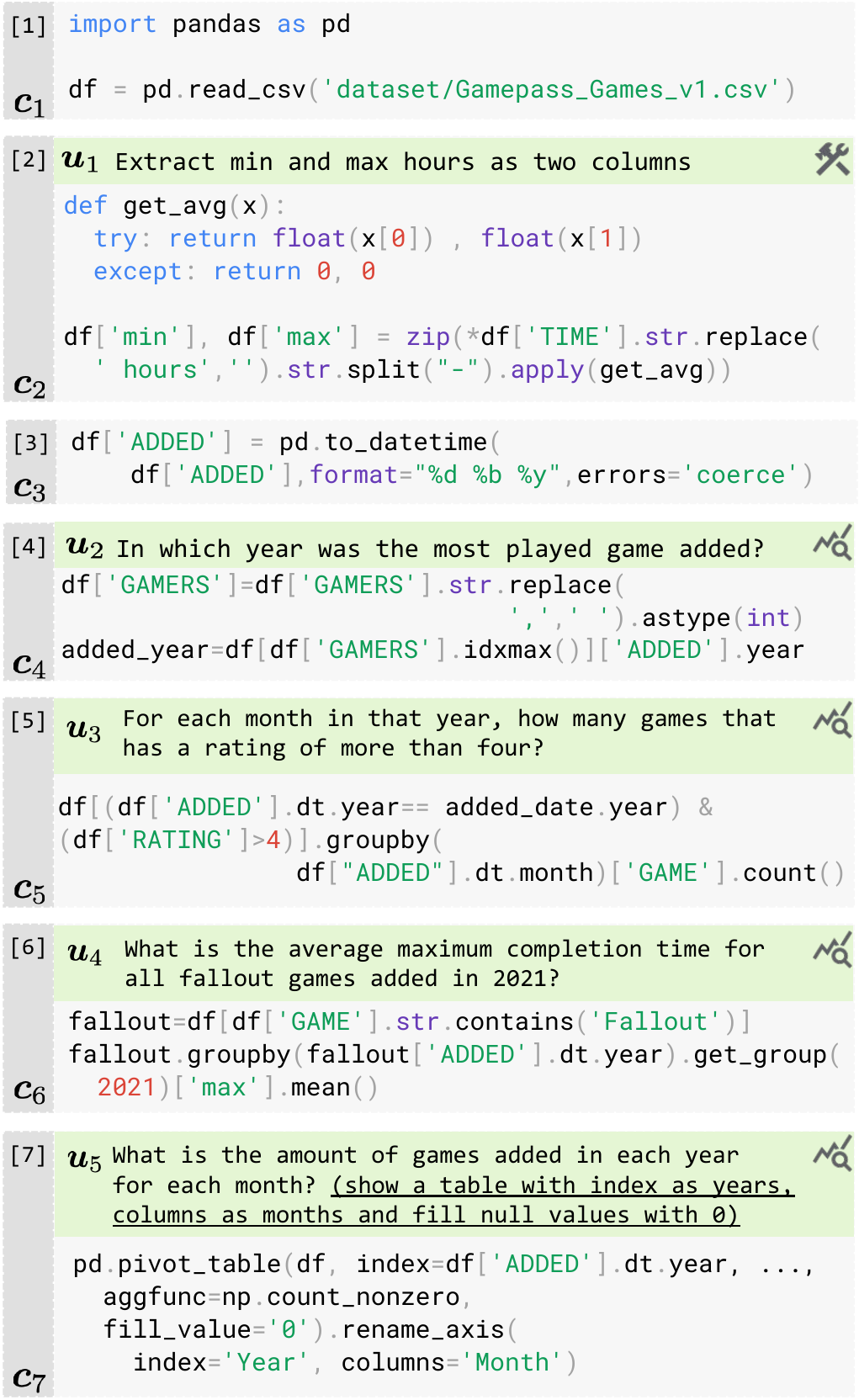}
    \caption{An example of a computational notebook adapted from our dataset, with examples of reading and preprocessing data (cell $c_1$), data wrangling (cell $\cell_2, \cell_3$), and data analysis (cells $c_3 -c_7$). Annotated NL intents are shown in green.}
    \label{fig:intro:teaser_example}
    \vspace{-1em}
\end{figure}

Data science is the process of extracting insights from data~\cite{wang2021autods}, and has become an integral part of  decision making and knowledge discovery~\cite{donoho50yearsds}.
Data scientists and machine learning (ML) practitioners often use \emph{computational notebooks},
which are interactive environments such as Jupyter notebooks~\cite{jupyter16paper} and Google Colab,\footnote{ \url{https://colab.research.google.com/}} in their work.
As illustrated in \cref{fig:intro:teaser_example}, data scientists spend significant amount of time on data wrangling and exploratory data analysis (EDA)~\cite{agashe-etal-2019-juice,wang2022documentation}.
This has motivated research on automating and accelerating the data science workflow in general~\cite{aggarwal2019hicanai,wang2021autods,wang2021howmuchautomationds}, with particular interest in data wrangling and EDA tasks~\cite{Bavishi2019AutoPandasNG,Jain2021JigsawLL,Karmaker2020AutoMLTD,Nazabal2020data,Kandel2011wrangler}.

On the other hand, large language models (LLMs) trained on code can assist developers by translating  natural language (NL) intents into executable programs~\citep[][\emph{inter alia}]{chen2021codex,austin2021lambdacode,chowdhery2022palm,nijkamp2022codegen,fried2022incoder}, with promising applications in synthesizing code for data wrangling and EDA tasks~\cite{Jain2021JigsawLL,rajkumar2022evaluatingllmsfortextsql,cheng2022binding}. 
Computational notebooks also present unique challenges to LLMs, because notebooks freely mix NL, code, graphics, and execution results~\cite{je2021reactive}, and because of their interactivity, notebooks feature multiple turns of related NL-to-code problems~\cite{Heyman2021NaturalLP}.
As illustrated in \cref{fig:intro:teaser_example}, those problems often have interesting dependency structures. They may share common execution context (\eg~{\tt DataFrame} {\tt df}), form  semantically coherent turns (\eg~$\cell_4, \cell_5$), or exhibit non-trivial long range data dependency (\eg~from $\cell_6$ to $\cell_2$, or $\cell_7$ to $\cell_3$). This dependency structure is more complex than existing multi-turn text to code tasks with sequentially dependent problems~\cite{nijkamp2022codegen}.

Several benchmarks have been proposed to evaluate program synthesis of data science programs from NL intents, but these datasets have several limitations.
First, some datasets derive from data science tutorial notebooks~\cite{agashe-etal-2019-juice,chandel2022msdsp}, but this data
tends to contain NL text (\eg~exercise questions) which is more verbose and elaborate than the concise, ephemeral style that developers write when interacting with code LMs~\citep[more in \cref{sec:dataset}]{barke2022groundedcopilot}.
Other datasets assume that the developer provides extra information, such as unit tests or input/output examples~\cite{chandel2022msdsp,jain2022jigsaw}, but such systems pose an extra burden to users who might not normally write such tests or examples during their workflow~\cite{Pimentel2019ALS}.
Finally, existing datasets usually contain independent tasks with isolated contexts~\citep{Lai2022DS1000}, or limited number of contextually dependent problems~\cite{huang2022execution}, rather than having multiple, related tasks with complex dependencies such as in \cref{fig:intro:teaser_example}.
Therefore, there is a need for a benchmark that provides \emph{realistic NL intents, rich notebook context, and multiple interrelated problems}, so as to better reflect real-world usage by data scientists.

To fill this gap, we present \dataset/,\footnote{\underline{A}nswer \underline{R}epository for \underline{C}omputational \underline{A}nalysis and \underline{D}ata \underline{E}ngineering} a new benchmark for code generation for data wrangling and EDA tasks in computational notebooks (\cref{sec:dataset}). 
\dataset/ consists of \numproblem/ problems spanning across 136 notebooks for \numdataset/ ML datasets.
\dataset/ features a series of NL utterances written by professional data scientists with the intention of interacting with an AI pair programmer when working in a notebook (\eg, green texts in \cref{fig:intro:teaser_example}), with high-quality code solutions using the {\tt pandas} library.
To mitigate the risk of data leakage when evaluating LLMs, $65\%$ of the problems and their notebooks are created from scratch, based on recent ML datasets on Kaggle.\footnote{\url{https://www.kaggle.com/}}
\dataset/ also challenges LLMs with grounded language understanding, where a model needs to ground relevant semantic concepts in intents (\eg~``\utterance{min and max}'' in $\intent_1$) to variable states (\eg~the column {\tt df['TIME']}) to understand the intent.
Besides featuring concise NL intents and contextually rich problems, more than $67\%$ of problems in \dataset/ also require complex solutions using $5$ or more {\tt pandas} API calls, making it more challenging than  most existing benchmarks.

To demonstrate how \dataset/ can motivate new research, we evaluate public code LMs, finding that they struggle on this task, especially newly created problems from scratch.
To build an LM tailored for data science, we develop \nbmodel/, a $62\mathrm{B}$ code language model for Python computational notebooks, trained on a mixture of general-domain NL, source code, and 
Jupyter notebooks~(\cref{sec:model}).
\nbmodel/ significantly outperforms  public code LMs on \dataset/.
Further, we explore few-shot prompting strategies to alter the style of model predictions, such as decomposing code into step-by-step solutions and adding inline NL explanations (\cref{sec:experiments:prompting_method}).
Not only is code in this style potentially more understandable to novice data scientists (\cref{sec:prompting_case_study}), 
we also demonstrate that this improves the accuracy of self-consistency reranking~\cite{wang2022selfconsistency}, because prompting the model to explain its solutions improves the diversity of the model's predictions.

\section{Problem: Code Generation in Computational Notebooks}
\label{sec:problem}

\paragraph{Problem Definition}
A computational notebook is an interactive computing environment that allows mixing code, 
text, and graphics. The notebook itself consists of a sequence of markdown or source code cells.
Formally, given a partial notebook context with $k$ cells $\{ c_i \}_{i=1}^k$ and a user-specified intent $\intent$ for the next cell $c_{k + 1}$ (\eg, $\intent_1$ in \cref{fig:intro:teaser_example} for $k=1$), we aim to generate code for $c_{k + 1}$ that fulfills the user's intent~\cite{agashe-etal-2019-juice}.
This process could proceed sequentially with multiple rounds between the user and a system~\cite{Heyman2021NaturalLP}.
To answer subsequent intents (\eg, $\intent_4$), a system will leverage the updated notebook context (\eg, $\{ c_i \}_{i=1}^5$) which includes previous problems (\eg, $\langle \intent_1 \sim \intent_3 \rangle$).

\paragraph{Tasks and Programming Library}
We focus on synthesizing programs for data wrangling and exploratory data analysis tasks.
Specifically, \textbf{data wrangling} refers to the process of cleaning and transforming data to allow for downstream analysis and machine learning, while \textbf{exploratory data analysis} involves querying the data for insights to support later statistical analysis and decision making.
We focus on generating code using {\tt pandas}, a popular Python library for performing wrangling and EDA tasks in computational notebooks, with built-in objects such as {\tt DataFrame}s and {\tt Series} to model tabular data types (tables and columns, respectively) .

\section{\dataset/: A Benchmark of {\tt \fontseries{b}\selectfont pandas} Data Science Code Generation}
\label{sec:dataset}

\subsection{Constructing \dataset/}
\label{sec:dataset:construction}

\dataset/ consists of \numproblem/ NL-to-code problems from both existing data science notebooks on GitHub (\cref{sec:dataset:construction:existingtasks}) and new ones curated from scratch (\cref{sec:dataset:construction:newtasks}).
In this section we elaborate on details to create annotated examples from both sources.

\subsubsection{Mining Examples from Existing Notebooks}
\label{sec:dataset:construction:existingtasks}

We identify candidate code cells performing data wrangling and EDA tasks from existing high-quality data science notebooks, and then manually annotate these cells with NL intents.

\paragraph{Collecting Notebooks for Annotation}
To form a pool of candidate notebooks, we use JuICe~\cite{agashe-etal-2019-juice}, a collection of Jupyter notebooks from GitHub, together with additional notebooks from \textsc{BigQuery}, yielding over $1.5\mathrm{M}$ notebooks in total.\footnote{\url{https://cloud.google.com/bigquery/public-data/}}
These notebooks are first filtered and near-deduplicated, similar to \nbmodel/'s training data preprocessing step in \cref{sec:model}.
We then identify candidate code cells from the remaining notebooks for annotation.
Specifically, we select code cells that are either 
(1) contain pandas programs with at least 3 {\tt pandas} API calls,
or (2) preceded by a markdown cell with a short question as its content (\eg, \utterance{What are the top 10 producers?}). 
The first heuristic is useful to identify complex wrangling tasks, while the second one is particularly effective in finding interesting dataset-specific EDA tasks, and the existing markdown texts also provide reference for labeling intents later.
Next, we group the notebooks with at least one candidate cell based on their underlying ML datasets (\eg~imported using {\tt pd.read\_csv()}),
and then select the top 5 notebooks with the most number of candidate cells from a curated set of 36 dataset groups for annotation.\footnote{This set contains ML datasets from a variety of domains and schema.}
We favor notebooks with more candidate cells so that we could extract multiple NL-to-code problems within the same notebook.

\paragraph{Annotating Cells with NL Intents}
We hired a group of data scientists to annotate the notebooks selected above.\footnote{Eight freelancers reported skill in {\tt pandas} are hired from Upwork, with an average of 3 years of experience.}
Annotation primarily consists of judging the quality of candidate code cells, fixing any errors, and creating NL intents summarizing the code.
Specifically, the annotators are instructed to frame their intents the way they prefer when interacting with an AI system to help them implement the existing code solution, while keeping the intents natural, concise without redundant elaboration, such as line-by-line explanation.
In addition, to make the intents more challenging, we encourage annotators to refer to entities and variables in intents using semantic rewrites without introducing ambiguity (\eg, use ``\utterance{convert all \underline{binary columns} to bool}'' instead of listing columns verbatim), reminiscent of synonym substitution for labeling utterances in text-to-SQL~\cite{gan-etal-2021-towards}.

\paragraph{Mitigate Ambiguity in NL Intents}
Creating succinct NL intents without ambiguity could be non-trivial in this open-domain code generation setting, especially when there could be multiple plausible interpretations of an intent.
For example, without the \underline{\textit{underlined part}} of $\intent_5$ (\cref{fig:intro:teaser_example}), a programmer or a system may propose alternative solutions using different table schema.
Therefore, for such open-ended problems where there could be multiple alternative ways to present the answer, we ask annotators to provide extra specification in their intents about the desired output (\eg~schema of output {\tt DataFrame}, such as the \textit{\underline{underlined part}} in $\intent_5$).
Even with these additional semantic constraints, empirically we observe around $45\%$ intents are still under-specified, making \dataset/ a challenging benchmark for handling realistic NL intents with uncertainty.
We present more analysis in \cref{sec:dataset:analysis}, while introducing a robust evaluation metric that mitigates this issue in \cref{sec:dataset:evaluation_metric}. 

\begin{table*}[t!]
    \centering
    \small
    \setlength{\tabcolsep}{3.3pt}
    \resizebox{\textwidth}{!}{\begin{tabular}{lccccccccr@{\hspace{2pt}}lc}
    \toprule
        \multirow{2}{*}{Dataset} & \multirow{2}{*}{Source} & Execu- & Evaluation & No. & No. & Problems & \multirow{2}{*}{Intents Type} & Intent & \multicolumn{2}{c}{AST Size$^\star$} & No. \\ %
        & & table? & Method & Notebooks & Problems & per N.B. & & Length & \multicolumn{2}{c}{All /\hspace{2pt}{\tt pandas}} & API$^\triangle$
        \\
        \rowcolor{black!10!} \multicolumn{12}{c}{\textbf{Existing Datasets}} \\
        JuICE (Eval) & GH & \xmark & EM$+$\textsc{BLEU} & 1,457 & 3,946 & 2.7 & Markdown & 60.2 & 21.2 &/ 24.3$^\ddagger$ & 2.5 \\
        MS DSP & GH & \cmark & Unit Tests & 305 & 1,096 & 3.6 & Markdown$+$Unit Tests & 54.3 & 28.7 &/ 34.8$^\ddagger$ & 3.1 \\
        ExeDS & GH & \cmark & Output Match & 277 & 534 & 1.9$^\dagger$ & Annotated NL & 20.0 & 9.0 &/ 10.7 & 2.4 \\
        NLGP & GH & \xmark & BLEU$+$Manual Eval & 150 & 201 & 1.3 & Annotated NL & 7.7 & 13.5 &/ 15.1 & 2.1 \\
        DS-1000$^\diamond$ & SO & \cmark & Constraints$+$Unit Tests & N/A & 1,000 & N/A & Annotated NL & 166.5 & 27.3 &/ 41.6$^\ddagger$ & 5.0 \\
        \rowcolor{black!10!} \multicolumn{12}{c}{\textbf{This Work: \dataset/}} \\
        \ablation{Existing Tasks} & GH & \cmark &Fuzzy & 63 & 422 & 7.7 & \multirow{2}{*}{Annotated NL} & 14.1 & \multicolumn{2}{c}{19.8} & 4.2 \\
        \ablation{New Tasks} & New & \cmark & Output Match  & 70 & 660 & 9.4 & & 18.4 & \multicolumn{2}{c}{27.2} & 5.8 \\
    \bottomrule
    \end{tabular}}
    \caption{Summary statistics of \dataset/ compared with existing work.
    Source: GitHub (\textbf{GH}) or StackOverflow (\textbf{SO}).
    $^\triangle$Number of {\tt pands} API calls calculated on the subset of examples using the library.
    $^\star$Metrics are averaged over all and ({\tt /}) the subset of examples using {\tt pandas}.
    $^\ddagger$AST sizes for existing datasets are for reference only because some code snippets also contain extra boilerplate code (\eg~function/class sketches) which is not part of solution.
    $^\dagger$Most problems in ExeDS actually do not use preceding examples in the same notebook as contexts for evaluation.
    $^\diamond$DS-1000 derived StackOverflow problems without multi-cell notebook context, and is not directly comparable with other work.
    }
    \label{tab:datasets_stat:comparision}
    \vspace{-1em}
\end{table*}

\paragraph{Other Considerations}
Indeed, re-purposing notebooks in the wild for our benchmark is not an easy task.
As an example, many notebooks in JuICe are data science tutorials, which often contains documentation that includes background knowledge, reference materials, and even solution hints.
Those extra information makes the code generation task easier, and may not reflect the style of ordinary notebooks authored by data scientists in their day-to-day work.
We therefore ask the annotators to clean the notebook and remove such extra information whenever possible.
We compile a 35-page annotation guideline to cover corner cases like this.
Refer to \cref{app:annotation_guideline} for more details about the guideline.

\subsubsection{Creating Notebooks with Examples from Scratch}
\label{sec:dataset:construction:newtasks}

The problems derived from high-quality GitHub notebooks in \cref{sec:dataset:construction:existingtasks} capture realistic tasks and notebook contexts, but may result in artificially high evaluation accuracies due to potential leakage of evaluation notebooks to the training data of LLMs, which is a common issue in LLM evaluation~\cite{brown2020language}.\footnote{Since JuICe and BigQuery primarily contain source files before or in 2019, this issue could be more problematic.}
To defend against this data contamination, we additionally annotated 660 problems by creating notebooks from scratch.

\paragraph{Sourcing Novel ML Datasets}
To ensure that those newly-created examples can be used to evaluate the generalization ability of code LMs on unseen ML datasets, we create notebooks targeting data wrangling and EDA tasks for $70$ tabular ML datasets that have been recently uploaded to the Kaggle data science platform since February 2022.
Those short-listed datasets are manually selected from a pool of $600$ datasets with reasonably complex schema (\eg, having columns with diverse data type), and are verified by our annotators that no older-versioned datasets with similar schema appeared before.

\paragraph{Creating Notebooks from Scratch}
For each ML dataset, the annotators were asked to create one notebook with a series of wrangling and EDA tasks annotated with NL intents.\footnote{We only invite the top-3 performers for this task since it is harder than labeling existing notebooks.}
Specifically, we ask annotators to come up with tasks that they would like to perform in order to gain insights into these recently appeared ML datasets in order to build models for them.
We follow the same standard to create intents as in \cref{sec:dataset:construction:existingtasks}.
To make the problems more challenging, annotators are encouraged to create harder tasks whose code solutions require at least $5$ {\tt pandas} API calls.
For quality assurance, each notebook is peer reviewed by another annotator, before a final round of review by the first author.
Since the annotators have already worked on the prior task of creating examples from existing notebooks, they are fairly familiar with the requirement, and are able to create each problem in $13$ minutes on average.

\subsection{Dataset Analysis}
\label{sec:dataset:analysis}

\dataset/ consists of \numproblem/ NL-to-code problems from \numnb/ notebooks based on \numdataset/ unique ML datasets.
We first present some analysis on our dataset, before comparing the statistics of \dataset/ with existing datasets in \cref{tab:datasets_stat:comparision}.

\paragraph{NL Intents are often Under-specified}
Since \dataset/ aims to evaluate code LMs in the real-world scenario where data scientists provide succinct NL intents without extra specification (\eg~I/O examples), the intents we collected 
are often under-specified and may not contain sufficient information to generate a solution that executes to the exact reference output. 
To understand the patterns of semantic ambiguity in user-issued intents, we conduct a case study, in which we manually examine a small group of random samples.
Among those samples, around $55\%$ of them are unambiguous, \ie, they are precise and sufficient to infer the target outputs. 
Those intents are often numerical queries with limited variety in output type (\eg, \utterance{How many customers receive an income of more than 95 percentile?}), or contain sufficient descriptions of output specification (\cref{sec:dataset:construction:existingtasks}, \eg, \utterance{List the 3 largest towns, \underline{return the name and area}}). The remaining
45\% of intents are under-specified.
Of these, $30\%$ lack a description of target columns in output {\tt DataFrame}s (\eg,~$\cell_4$, \cref{fig:intro:teaser_example}).
An additional $25\%$ of intents imply that the desired {\tt DataFrame}s should have a complex schema, such as a nested row index or table header (\eg, \utterance{Show the \underline{time of the day} and the fare price \underline{for each airline}}), which is difficult to predict without additional information.
Another $25\%$ of ambiguous intents have lists of entities as their answers, and do not specified the desired container type (\eg~{\tt List} or {\tt pandas.Series}).
Finally, the remaining $20\%$ cases require outputs with more complex structures (\eg, multiple variables), or imply additional post-processing steps such as data imputation.

\paragraph{Notebook Context help Disambiguate Intents}
Interestingly, while nearly half of the intents are under-specified, we observe that $\sim 30\%$ of those cases can be disambiguated by referring to intents or code snippets from prior rounds of problems with similar query patterns in the notebook context.
A common scenario is follow-up EDA queries  (\eg~\utterance{How many of them are $\ldots$}) that refine the previous query step (\eg~\utterance{Show me all $\ldots$}) by narrowing the down search condition, while the output {\tt DataFrame} structure is the same as previous turns, reminiscent of similar thematic relation patterns in contextual semantic parsing~\cite{Iyyer2017SearchbasedNS,Yu2019SParCCS}.

\begin{figure}[t]
    \centering
    \includegraphics[width=\columnwidth]{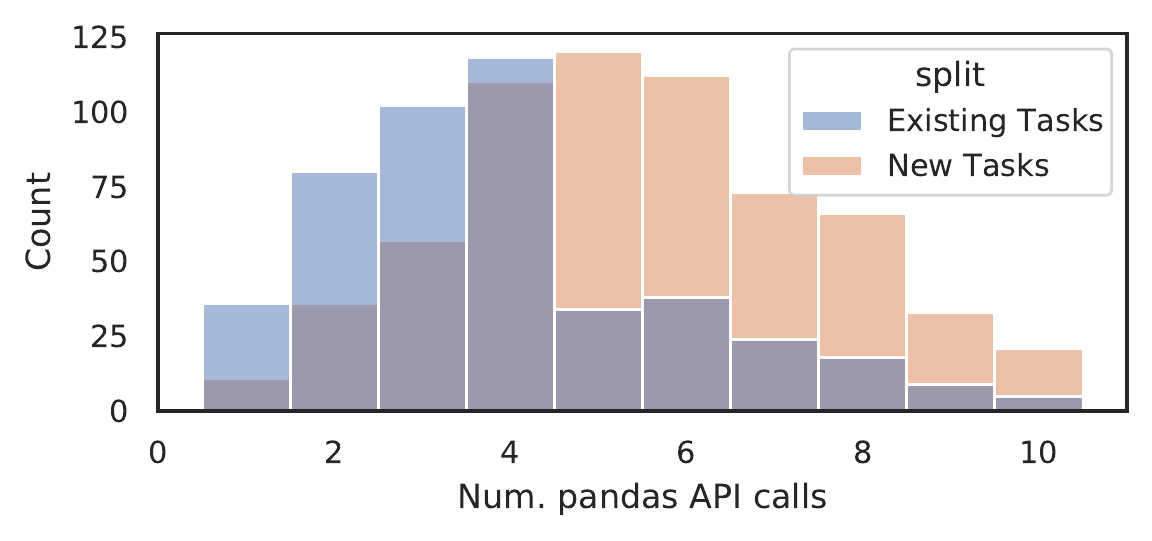}
    \caption{Histogram of the number of {\tt pandas} API calls in \dataset/.}
    \label{fig:dataset:compare_two_splits}
    \vspace{-1em}
\end{figure}

\paragraph{Comparing Existing and New Tasks}
Looking closer into the problems in the \textbf{\existtasks/} (\cref{sec:dataset:construction:existingtasks}) and \textbf{\newtasks/} (\cref{sec:dataset:construction:newtasks}) splits, the \newtasks/ split has more challenging problems than \existtasks/, as measured by the number of {\tt pandas} API invocations and the AST size in reference code solutions (\cref{tab:datasets_stat:comparision}, \emph{Bottom}).
\cref{fig:dataset:compare_two_splits} plots the histogram of pandas API usage, which shows $67\%$ problems in \newtasks/ require at least 5 API calls to solve them.
As we show in \cref{sec:experiments}, with more complex held-out problems targeting for recent ML datasets, the \newtasks/ split serves both as a more robust and challenging benchmark for state-of-the-art code LLMs.

\paragraph{Comparing with Existing Datasets}
\cref{tab:datasets_stat:comparision} compares \dataset/ with existing datasets for data science code generation in computational notebooks.
Specifically, \textbf{JuICe}~\cite{agashe-etal-2019-juice} contains exercise problems in assignment notebooks from data science tutorials or coures, where the NL intents are usually elaborative assignment problem definitions.
Notebooks in JuICe are not executable so evaluation is performed by surface-level matching (exact match or \textsc{Bleu}) between reference and predicted programs.
\textbf{DSP}~\cite{chandel2022msdsp} contains problems from a filtered set of JuICe notebooks that are executable and also associated with unit tests  for auto-grading.
Hence the intents in DSP follow similar patterns as those in JuICe.
To ensure that the free-form model-predicted code is compatible with unit tests, DSP uses the unit test code iteself as extra model input besides NL intents to constrain the model to generate code that could be directly consumed by the tests.
Next, \textbf{ExeDS}~\cite{huang2022execution}, a concurrent work to this paper, is another set of filtered problems from JuICe.
Similar to this work, ExeDS uses hand-annotated intents, and compares the execution output between reference and predicted code for evaluation instead of relying on unit tests (\cref{sec:dataset:evaluation_metric}).
Moreover, \textbf{NLGP}~\cite{Heyman2021NaturalLP} is another collection of the NL-to-code problems in Jupyter notebooks with short annotated intents for simple data manipulation tasks, where most notebooks have one associated problem.
Finally, for the sake of completeness, we also list another concurrent work \textbf{DS-1000}~\cite{Lai2022DS1000}, a collection of data science problems derived from StackOverflow questions.
It primarily features problems using synthetic contexts with minimal working examples, and therefore does \emph{not} concerns with code generation in notebooks with interrelated problems on general ML datasets.

We remark that \dataset/ is the only benchmark that satisfies \emph{all} the following criteria:
\emph{First}, our dataset consists of manually annotated, concise NL intents (``Intents Type'' column in \cref{tab:datasets_stat:comparision}).
Those utterances are significantly shorter (``Intent Length'' column) than verbose markdown texts often found in tutorial notebooks (\textit{c.f.~}JuICe, DSP), and do not require a model to rely on extra specification such as unit tests (\textit{c.f.~}DSP).
Those succinct intents are also under-specified, requiring a more robust evaluation metric to consider alternative answers (later in \cref{sec:dataset:evaluation_metric}).
This may also motivate future research on generating more diverse predictions to cover possible interpretations of a problem --- which we will explore in \cref{sec:experiments} --- or even models to explicitly capture such uncertainty in NL intents~\cite{lin22uncertainty}.
\emph{Second}, \dataset/ contains more interrelated problems in a single notebook (``Problems per N.B.'') with complex dependencies (\cref{fig:intro:teaser_example}), capturing the essence of interactive computing.
This makes our dataset useful in testing an LLM's skill at understanding rich contexts, including existing user-written code and markdown cells, as well as preceding problems and their solutions (\cref{sec:problem}).
\emph{Third}, \dataset/ challenges LLMs with grounded language understanding, where the model needs to link relevant concepts (\eg~``\utterance{max and min}'' in $\intent_1$) in an intent to the corresponding variable execution states in the context (\eg~content of the {\tt TIME} column in {\tt df}).
Such necessity of understanding semi-structured tabular data contents~\cite{Pasupat2015CompositionalSP} and performing data transformations using an open domain programming language (Python) potentially makes NL grounding in \dataset/ more challenging than other applications like database semantic parsing~\cite{Yu2019SParCCS}.
\emph{Fourth}, \dataset/ has problems with complex data transformations and richer usage of real-world data science APIs.
As evidence for this, note that the number of {\tt pandas} APIs used in each problem (``No.~API'' in \cref{tab:datasets_stat:comparision})\footnote{Calculated by counting function names in a predefined list of {\tt pandas} functions, including advanced array indexing operations. Functions with similar names from other libraries (\eg~{\tt numpy}) may also be included.} is on par with DS-1000 and significantly higher than ExeDS, DSP and NLGP.
\emph{Finally,} nearly $70\%$ of problems and their notebooks (\newtasks/) are curated from scratch targeting for recent ML datasets, which mitigates the risk of training data contamination of LLMs, and also makes \dataset/ a more reliable benchmark to test the generalization ability of LMs in understanding held-out tabular datasets~\cite{Lee2021KaggleDBQARE}.

To summarize, \dataset/ features \emph{more realistic NL intents} and \emph{multiple related problems in the same notebook context} to better reflect the real-world scenario where data scientists prompt LLMs using ephemeral NL comments for assistance with complex data wrangling or EDA tasks~\cite{barke2022groundedcopilot}, therefore making it ideal to evaluate LLMs in this realistic setting.
Nevertheless, we remark that existing datasets in \cref{tab:datasets_stat:comparision} usually contain broader types of problems other than the wrangling and EDA tasks (\eg~fitting ML models) considered in this paper.
We leave expanding the task spectrum as important future work.

\subsection{Evaluation by Fuzzy Output Matching}
\label{sec:dataset:evaluation_metric}

Recent code generation datasets have shifted from surface form evaluation metrics like \textsc{Bleu}~\cite{yin-neubig-2017-syntactic} to functional correctness, where the execution results of predicted programs on some input examples are matched with the reference output verbatim~\cite{hendrycksapps2021,chen2021codex,austin2021lambdacode}, or approximately \citep[for complex output like plots, see][]{chandel2022msdsp} using unit tests. 
Such evaluation methods assume a model generates code that strictly follows the signature of unit tests, which is impractical in data science notebooks with free-form code.

In this paper we aim to evaluate accuracy of code LLMs in synthesizing code for data science notebooks in the wild, \ie, the model is only conditioned on preceding notebook context besides a short user-issued intent (\cref{sec:problem}).
As discussed in \cref{sec:dataset:analysis}, those intents are often under-specified and have multiple alternative solutions.
We therefore use a set of heuristics to approximately match the execution output\footnote{For code that in-place modifies a variable (\eg~{\tt df} in $\cell_2$), we treat the modified variable as the output.} of a predicted program with the annotated reference to determine if they are functionally equivalent.
These heuristics primarily cover two scenarios.
First, if the output variable is a container type,\footnote{{\tt List}, {\tt Tuple}, {\tt Set}, {\tt numpy.ndarray}, {\tt pandas.Series} and single-column {\tt pandas.DataFrame}.} we canonicalize variables to the same type.
Second, we allow for partial matching between complex {\tt DataFrame} variables.
Formally, consider a reference {\tt DataFrame} $\bm{v}$ with a set of columns $\{ v_i \}$, where each $v_i$ is a vector of cell values for the $i$-th column.
We define that $\bm{v}$ is equivalent with the output {\tt DataFrame} of a predicted program $\hat{\bm{v}}$, \textit{iff} for any $v_i \in \bm{v}$, we have $v_i \in \hat{\bm{v}}$.
Intuitively, we consider a predicted program as correct if its output {\tt DataFrame} contains all the columns (and cell entries) in the reference frame.
Intuitively, a predicted program is already useful as long as it covers all the necessary information in the reference.
In cases of {\tt DataFrame}s, a user could easily create a more compact view of the frame by selecting a subset of target columns.

Empirically, we find our evaluation metric is reliable in identifying solutions with alternative output structures, with a relatively low false-negative rate (under $10\%$). 
We present an error analysis in \cref{sec:error_analysis}.

\section{\nbmodel/: Adapting Code LMs to Computational Notebooks}
\label{sec:model}

In this section we elaborate on \nbmodel/, our code LM for Python notebooks.

\paragraph{Base LM}
\nbmodel/ is based on \palm/, a family of Transformer-based LMs developed for few-shot learning for general natural language tasks~\cite{chowdhery2022palm}.
Specifically, we use the 62B \palm/ model trained on $1.3$T tokens with a mixture of conversational, webpages and code data (Section F, \newcite{chowdhery2022palm}).
Starting with 62B \palm/ as the base model, we perform two stages of  fine-tuning, first on Python source code data, followed by another fine-tuning run on a collection of Jupyter notebooks.

\paragraph{Fine-tuning on Python Code}
We first fine-tune \palm/ 62B on a large corpus of Python source code with 64B tokens.
This corpus consists of deduplicated Python source files with permissive license collected from GitHub.
We finetune \palm/ for 1 epoch with a batch size of 256 while keeping the other hyper parameters consistent with \citet{chowdhery2022palm}.
As we show in \cref{sec:experiments} and also in \cref{app:palm_python_finetuned_performance}, this Python-finetuned \palm/ 62B model is already a strong code LM, registering significant improvements on \dataset/ over the base model, while even outperforming the larger code LM \palm/-Coder 540B on existing benchmarks.

\paragraph{Continued Fine-tuning on Notebooks}
After fine-tuning on Python code, we perform a second stage of fine-tuning on a large collection of Jupyter notebooks.
Our Jupyter notebook dataset is created similarly as the Python code corpus from GitHub, with additional domain-specific pre-processing steps, such as filtering out notebooks with fewer than 4 cells.
In addition, to mitigate the risk of having notebooks similar to the evaluation notebooks from GitHub in the \existtasks/ split leaked into the training data, we perform near de-duplication against notebooks in \existtasks/ at the cell level.
Specifically, we cluster the cells of notebooks in both the evaluation and training sets based on a fuzzy-matching similarity metric, and remove any training notebooks that has one cell that falls into the same cluster as a cell from one of the evaluation notebooks.
This process eliminates ${\sim}350\mathrm{K}$ notebooks from the fine-tuning data.
Our final training set consist of ${\sim}3.8\mathrm{M}$ notebooks and ${\sim}9.6\mathrm{B}$ tokens in total.
We use {\tt nbconverter} to linearize Jupyter notebooks to Python code, where cells in a notebook are concatenated using the special delimiter `{\tt \# In[]:}', with markdown texts converted to comments.
Refer to \cref{app:palm_fine_tuning} for details about fine-tuning, and \cref{app:datacard} for a data card.

\section{Experiments}
\label{sec:experiments}

\subsection{Models and Setup}

To demonstrate the challenges of \dataset/, we compare the performance of \nbmodel/ with existing large code language models.

\noindent \textbf{\codegen/} \cite{nijkamp2022codegen} is a family of code LMs ($350\mathrm{M}$, $2.7\mathrm{B}$, $6.1\mathrm{B}$ and $16.1\mathrm{B}$ parameters) trained using a combination of NL and code data scraped from GitHub \citep[\textsc{ThePile},][]{pile}.
We evaluated on both the \textbf{multi}lingual version trained on \textsc{BigQuery} and the \textbf{mono}lingual version trained on $71\mathrm{B}$ Python tokens.

\noindent \textbf{\incoder/} \cite{fried2022incoder} is a code LM ($1.3\mathrm{B}$ and $6.7\mathrm{B}$ parameters) for both left-to-right generation and infilling tasks.
The LM is trained on a mixture of StackOverflow (SO) posts and $159\mathrm{GB}$ of multilingual code data.
In particular, the corpus contains a dedicated portion of Jupyter notebooks ($\sim 5\mathrm{GB}$), making it appealing to evaluate on \dataset/.

\noindent \textbf{\palm/} Besides \nbmodel/, we also evaluate the base \palm/ $62\mathrm{B}$ LM, as well as the Python code LM obtained from the first-stage of code fine-tuning (\cref{sec:model}).

\phantom{xx}

\paragraph{Inference}
For each problem, we convert the problem into a prompt (\cref{sec:experiments:prompting_method}) and draw random samples from the LMs using nucleus sampling with a top probability of $0.95$ and a temperature $0.8$. Refer to \cref{app:inference_setup} for details.

\paragraph{Metrics}
Following prior work~\cite{chen2021codex,austin2021lambdacode}, we measure model performance using the \passat{k} metric, defined as the fraction of problems with at least one correct sample given a sample size $k$.
To reduce variance, we estimate \passat{k} ($k \le 30$) by drawing $50$ samples for each problem \cite{chen2021codex}.

\begin{figure}
    \centering
    \small
    \includegraphics[width=\columnwidth]{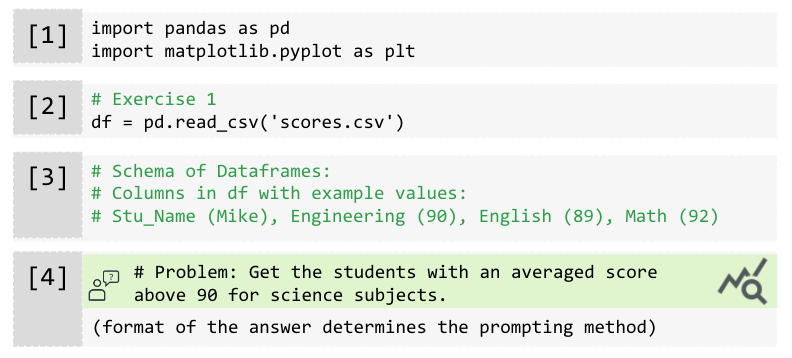}
    \caption{%
    An example problem. Cells 1-3 encode the notebook context. Cell 4 contains the NL intent as problem statement.}
    \label{fig:score}
\end{figure}

\begin{figure}
    \centering
    \small
    \includegraphics[width=\columnwidth]{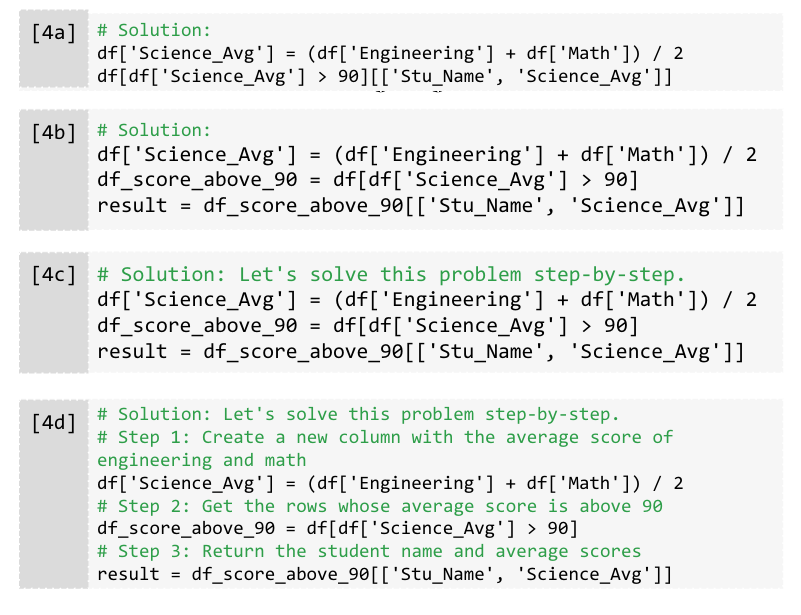}
    \caption{Example completions of Cell 4 from \cref{fig:score} which illustrate the desired coding style for different few-shot prompting strategies.}
    \label{fig:score_completions}
\end{figure}

\subsection{LM Prompting Strategies}
\label{sec:experiments:prompting_method}

For a given problem, we conducted two types prompting experiments: prompting using only the notebook context of the problem (\cref{sec:exp:results}),
and few-shot prompting with extra exemplars as prompt prefix before notebook context (\cref{sec:exp:few_shot_prompting}), in order to impose more control to the style of predicted code.

\subsubsection{Prompting using Notebook Context}
\label{sec:exp:prompt_prefixes}

In our default setting (\cref{sec:exp:results}), for a given problem we turn its notebook context to a prompt to query LLMs. 
\cref{fig:score} gives an example.
The context consists of preceding cells before the problem, which may include previous rounds of interactions (\cref{sec:problem}), and is followed by the current NL intent.
To help LLMs understand intents using variable state information (\cref{sec:dataset:analysis}), we also add NL descriptions of the schema of the dataset imported to the notebook (\textbf{schema description}, \eg~cell 3, \cref{fig:score}), which consist of columns and example values in the {\tt DataFrame}.
Representing structured schema information in NL is a common strategy when prompting LLMs to solve tasks requiring understanding structured data~\cite{Xie2022UnifiedSKGUA}.
Next, for the following problems after cell 4, we include reference solutions to previous problems in their prompts, following the multi-turn task-oriented dialogue setting in \citet{Andreas2020TaskOrientedDA}.
See \cref{app:prompts} for a complete example.

\subsubsection{Few-shot Prompting}
\label{sec:exp:prompt_prefixes}

Besides the basic setting using the notebook context of a problem, we also explored prompting using additional NL-to-code exemplars as prompt prefix before the notebook context. The motivation is to nudge the LM to generate code solutions following different coding and commenting styles, such as documenting the ``reasoning path'' used by the code.
We develop four prompting strategies:
\begin{enumerate}
    \item {\bf Vanilla code strategy} as in \cref{fig:score_completions}, completion 4a. The predicted code follows the common practice of chaining multiple API calls in a single line.
    \item {\bf Step-by-step (SbS) code strategy} as in \cref{fig:score_completions}, completion 4b, which leads to code with fine-grained decomposition structure.
    \item {\bf SbS code with preamble}, as in \cref{fig:score_completions}, completion 4c, which could further elicit decomposition in predictions.
    \item {\bf SbS code with preamble and explanation}, as in \cref{fig:score_completions}, completion 4d, with inline NL explanations per step.
\end{enumerate}

\begin{table*}[ht]
    \small
    \centering
    \resizebox{\textwidth}{!}{\begin{tabular}{l cccc : cccc}
    \toprule
        \multirow{2}{*}{\passat{k}} & \multicolumn{4}{c}{\bf \textit{Existing Tasks}} & \multicolumn{4}{c}{\bf \textit{New Tasks}} \\
         & 5 & 10 & 20 & 30 & 5 & 10 & 20 & 30 \\ \midrule
        \incoder/ $1\mathrm{B}$ & $30.1$ & $36.4$ & $42.6$ & $46.0$ & $3.8$ & $5.8$ & $8.4$ & $10.0$ \\
        \incoder/ $6\mathrm{B}$ & $41.3$ & $48.0$ & $54.0$ & $57.2$ & $7.0$ & $10.0$ & $13.5$ & $15.8$ \\ 
        \codegensym{multi}{350}{M} & $13.3$ & $16.0$ & $18.8$ & $20.4$ & $1.0$ & $1.6$ & $2.5$ & $3.1$ \\
        \codegensym{multi}{2}{B} & $25.0$ & $29.9$ & $34.8$ & $37.7$ & $2.7$ & $4.0$ & $5.5$ & $6.5$ \\
        \codegensym{multi}{6}{B} & $28.0$ & $34.0$ & $39.9$ & $43.3$ & $3.0$ & $4.6$ & $6.7$ & $8.0$ \\
        \codegensym{multi}{16}{B} & $31.2$ & $37.6$ & $43.8$ & $47.2$ & $4.6$ & $6.8$ & $9.5$ & $11.3$ \\
        \codegensym{mono}{350}{M} & $18.9$ & $24.3$ & $30.1$ & $33.4$ & $1.9$ & $2.8$ & $4.0$ & $4.9$ \\
        \codegensym{mono}{2}{B} & $35.8$ & $43.3$ & $50.7$ & $54.9$ & $6.5$ & $9.8$ & $13.8$ & $16.3$ \\
        \codegensym{mono}{6}{B} & $42.1$ & $49.4$ & $56.3$ & $59.7$ & $8.9$ & $13.1$ & $18.3$ & $21.6$ \\
        \codegensym{mono}{16}{B} & $46.7$ & $54.5$ & $60.9$ & $64.0$ & $12.0$ & $17.1$ & $22.7$ & $26.3$ \\
        \hdashline
        \palm/ $62\mathrm{B}$ ($1.3\mathrm{T}$ Tokens) & $49.7$ & $57.5$ & $64.4$ & $67.8$ & $12.5$ & $17.4$ & $22.8$ & $26.0$ \\
        ~~~$+$ Python Code & $58.8$ \adds{9.1} & $65.7$ \adds{8.2} & $71.7$ \adds{7.3} & $74.7$ \adds{6.9} & $21.4$ \adds{8.9} & $28.4$ \adds{11.0} & $35.6$ \adds{12.8} & $39.8$ \adds{13.8} \\
        ~~~$+$ Notebooks (\nbmodel/) & $64.6$ \adds{7.8} & $71.0$ \adds{5.3} & $76.0$ \adds{4.3} & $78.3$ \adds{3.6} & $30.6$ \adds{9.2} & $38.0$ \adds{9.6} & $45.0$ \adds{9.4} & $48.6$ \adds{8.8} \\
        ~~~~~~~$-$ Schema Description & $60.5$ \mi{4.1} & $67.1$ \mi{3.9} & $72.8$ \mi{3.2} & $75.5$ \mi{2.8} & $22.7$ \mi{7.9} & $28.5$ \mi{9.5} & $34.1$ \mi{10.9} & $37.2$ \mi{11.4} \\
    \bottomrule
    \end{tabular}}
    \caption{\passat{k} evaluation on \dataset/ using notebook context as prompts. Grey figures indicate difference to previous line.}
    \label{tab:exp:end2end_results}
    \vspace{-1em}
\end{table*}

\noindent Each strategy has four exemplars in its prompt prefix, as listed in \cref{app:prompts}.
Those extra exemplars before notebook context could help the model better understand the task~\cite{brown2020language}.
More importantly, we could also use additional exemplars to impose more control on the inference process of LMs, with each strategy imposing a slightly different level of granularity of the expected completion. 
Specifically, we focus on prompting LMs to generate code that follows a multi-line, step-by-step decomposition structure (\sbshort/), with NL explanations of the underlying data transformation for each step.
Prompting LMs to explain its own solution in such a step-by-step fashion has been found effective in improving accuracy of natural language reasoning \cite{Wei2022ChainOT,Gao2022PALPL} and program induction \cite{Nye2021ShowYW} tasks.
In \cref{sec:exp:few_shot_prompting}, we show it effectively improves diversity of predicted code solutions, which is beneficial for self-consistency reranking.

\subsection{Results}
\label{sec:exp:results}

In our first set of experiments, we evaluate the performance of state-of-the-art code LMs on \dataset/.
Results are listed in \cref{tab:exp:end2end_results}.
As in \cref{sec:experiments:prompting_method}, inputs to those models are only the notebook context (no few-shot prompting). We truncate the prompt up to $900$ tokens using the tokenizer of \palm/.

\paragraph{Comparing \palm/ Models}
Our \nbmodel/ model registers the best performance on both the \existtasks/ split based on high-quality GitHub notebooks as well as on the \newtasks/ split with held-out ones, thanks to its larger size and domain-specific fine-tuning on notebooks.\footnote{See \cref{sec:app:pass_at_one} for \passat{1} results.}
It is worth noting that the base NL-focused \palm/ 62B model already outperforms most public code LMs and is on par with \codegensym{mono}{16}{B}. 
As the LM is fine-tuned on Python ($+$ Python Code) and notebooks ($+$ Notebooks) data, the domain gap is closing, and \passat{k} continues improving.
Indeed, the Python fine-tuned \palm/ is already a strong model for Python code generation, achieving state-of-the-art results on benchmarks such as \textsc{HumanEval} and \textsc{MBPP} (\cref{app:palm_python_finetuned_performance}).
Meanwhile, the absolute gain after fine-tuning on Python code is higher than continued training on notebooks, which is not surprising since the semantic gap between NL data and Python code is larger than that between code in Python and notebooks.

\paragraph{Comparing Public Code LMs}
Next, we discuss results on public code LMs.\footnote{We also evaluate \textsc{Codex}. Results are in \cref{app:codex_results}.} 
First, among models at similar size and trained with roughly similar amount of Python code data, namely \incoder/ 6B and \codegen/$_\textrm{multi}$ 6B, \incoder/ 6B performs better. This is likely due to that \incoder/ contains dedicated portion of Jupyter notebooks (8GB) in its training data, which makes the model more suitable for \dataset/.
With $4 \times$ more Python data, \codegen/$_\textrm{mono}$ 6B takes over, which is in line with the the performance pattern of these two model families on existing code generation datasets~\citep{fried2022incoder}.
Moreover, comparing \codegen/ models at different size, we observe that \passat{k} scales linearly with model size. 
As we discuss more in~\cref{sec:app:dataset_model_scaling_curve}, the scaling curve for \dataset/ (\newtasks/) is also more flatten than that for other code generation datasets, which might suggest that it is more challenging for \codegen/.

\paragraph{Comparing \existtasks/ and \newtasks/}
It is interesting to compare results between the \existtasks/ and \newtasks/ splits, where we observe two patterns.
First, results on \existtasks/ are significantly higher across all models. 
Second, comparing the improvements after Python and notebook-specific fine-tuning of \palm/ 62B, the gain on \newtasks/ is higher.
These results are likely due to two factors.
First, an obvious reason is that problems in \existtasks/ are overall simpler than \newtasks/  (\cref{sec:dataset:analysis}). 
Second, it is likely that some code data similar to our evaluation notebooks in \existtasks/ split is leaked into the training data of those LMs.
Specifically for \palm/, its training mixture contains general Web documents and sampled Python source files.
Despite the fact that we went through significant amount of effort to deduplicate the training data against \dataset/ during the second-stage notebook fine-tuning (\cref{sec:model}), 
given that some of the notebooks used to create \existtasks/ are popular data science tutorials on GitHub, 
it is still possible \palm/ has already seen similar code data from the Web or notebook-converted Python source files.
This suggests the importance of evaluating code LMs on held-out data, which is the purpose of \newtasks/.
We leave investigating training data contamination as important future work.

\paragraph{Impact of Schema Descriptions}
Our prompts contain NL descriptions of schema for {\tt DataFrame}s imported to a notebook (\cref{sec:exp:prompt_prefixes}).
As in \cref{sec:dataset:analysis}, variable state information like {\tt DataFrame} schema is crucial for understanding NL intents.
The last row in \cref{tab:exp:end2end_results} ($-\bm{\mathrm{Schema~Description}}$) reports an ablation where we remove such NL schema descriptions from prompts, which leads up to significantly worse results, especially on the \newtasks/ split.
Encoding the states for more intermediate variables could potentially further improve performance, which we leave as interesting future work.

\begin{figure}[t]
    \centering
    \small
    \includegraphics[width=0.95 \columnwidth]{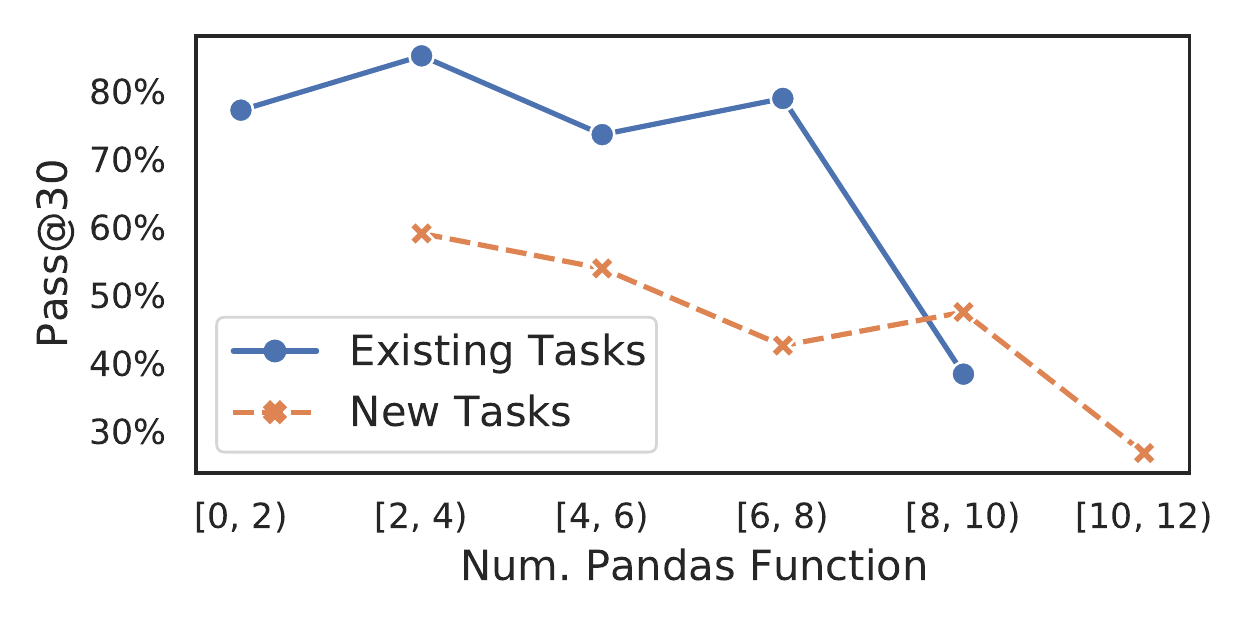}
    \caption{\passat{k} of \nbmodel/ w.r.t~problem complexity}
    \label{fig:exp:pass_at_k_wrt_task_complexity}
\end{figure}

\paragraph{Accuracy with Problem Complexity}
To better understand \nbmodel/'s performance on problems at different levels of complexity, we plot \passat{30} with respect to the number of {\tt pandas} function calls in the annotated reference solutions, as shown in \cref{fig:exp:pass_at_k_wrt_task_complexity}. 
For problems with similar complexity, \nbmodel/ generally achieves higher pass rate on \existtasks/, again suggesting that the \newtasks/ split is still more challenging even after controlling problem complexity.
Refer to \cref{app:pass_at_k_wrt_task_complexity:ast} for a similar plot w.r.t~AST size.

\paragraph{How Much Notebook Context is Useful?}

\begin{figure}[t]
    \centering
    \small
    
    \includegraphics[width=0.95 \columnwidth]{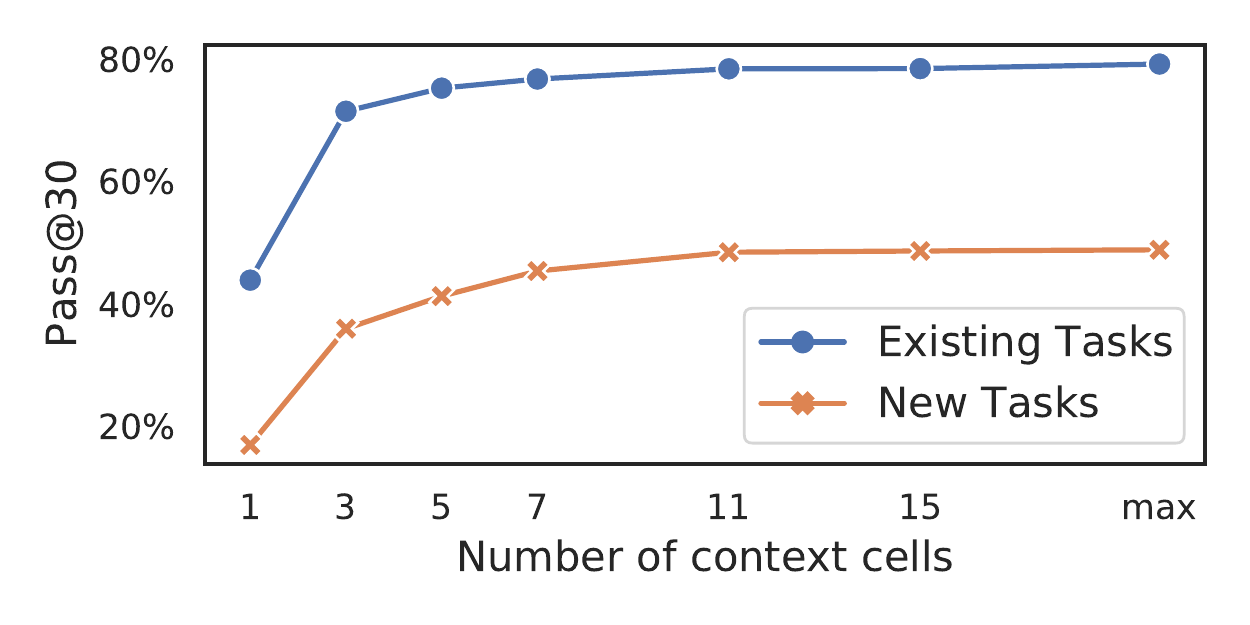}
    \caption{\passat{30} as w.r.t~the number of context cells from the notebook.
    }
    \label{fig:exp:ctx_size:vary_notebook_cell_num}
\end{figure}

\dataset/ requires a model to leverage rich programmatic and NL context in test notebooks to generate code solutions for the current cell.
To study \nbmodel/'s performance with varying amount of available notebook context, we control the number $d$ of context cells $\{ \cell_i \}_{i=k - d}^{k - 1}$ (\cref{sec:problem}) when generating code for each problem (at cell $\cell_k$) in our dataset.
\cref{fig:exp:ctx_size:vary_notebook_cell_num} depicts \passat{30} as a function of the context size $d$.
Since we use the first preceding cell $\cell_{k-1}$ to store the NL intent $\intent_k$ for $\cell_{k}$ (\cref{app:prompts}), having only one context cell is equivalent to the ``cold-start'' setting of only using $\intent_k$ (besides schema description) to predict $\cell_{k}$.
\nbmodel/ achieves a pass rate of $44\%$ (existing tasks) and $17\%$ (new tasks) in this challenging setting ($d=1$), with errors mostly due to failure in referring to variables that the solution relies on, whose information is not present in the short context.
Indeed, including additional context cells is crucial for good performance.
In particular, having 3 context cells could already lift the \passat{30} to $72\%$ and $36\%$ on the two splits --- $1.6 \sim 2 \times$ higher than $d=1$.
The results also start to plateau after including $5 \sim 7$ context cells, with diminishing returns after including more cells, which is in line with findings in~\citet{agashe-etal-2019-juice}.
\footnote{Prior work suggests that the plateau point is around three neighboring cells, while in our case, the number if approximately doubled since we need extra cells to include intents in previous turns (\cref{app:prompts}).}
Empirically, we observe that using more context helps to reduce schema understanding errors (\eg~using undefined columns in {\tt DataFrame}s).
Refer to \cref{sec:app:error_breakdown_wrt_ctx_size} for more details.

\begin{figure*}[t]
    \centering
    \includegraphics[width=\textwidth]{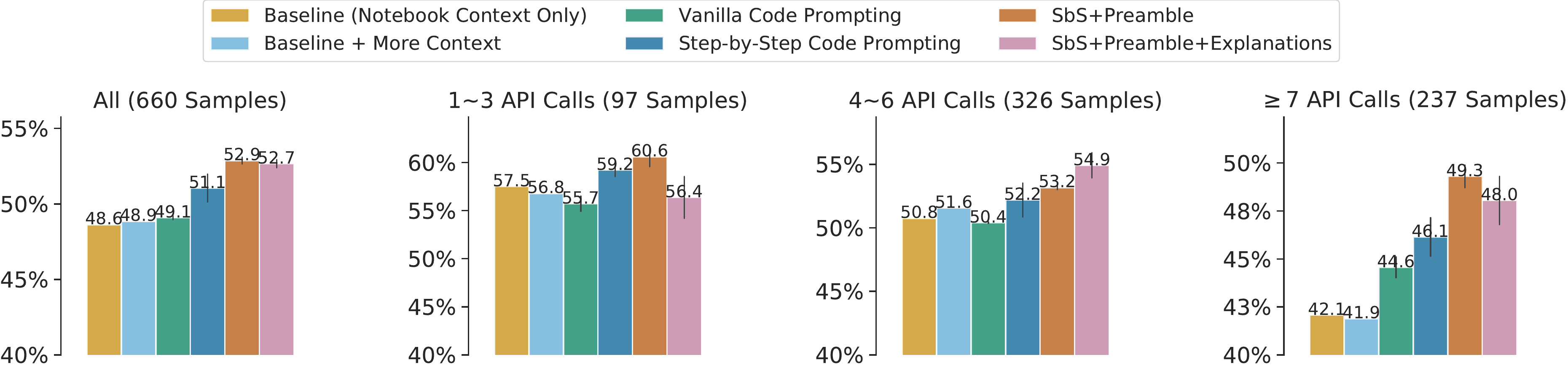}
    \caption{\passat{30} for few-shot prompting on \newtasks/. Results averaged over three runs with different prompt prefixes.}
    \label{fig:exp:prompting:passatk}
\end{figure*}
\begin{figure*}[t]
    \centering
    \includegraphics[width=\textwidth]{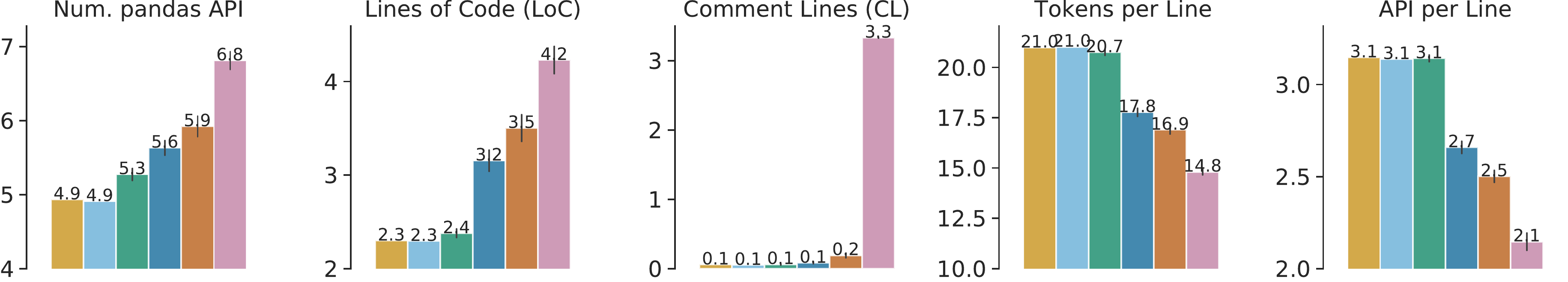}
    \caption{Few-shot prompting code style metrics}
    \label{fig:exp:prompting:code_style}
\end{figure*}

\begin{figure}[t]
    \centering
    \includegraphics[width=\columnwidth]{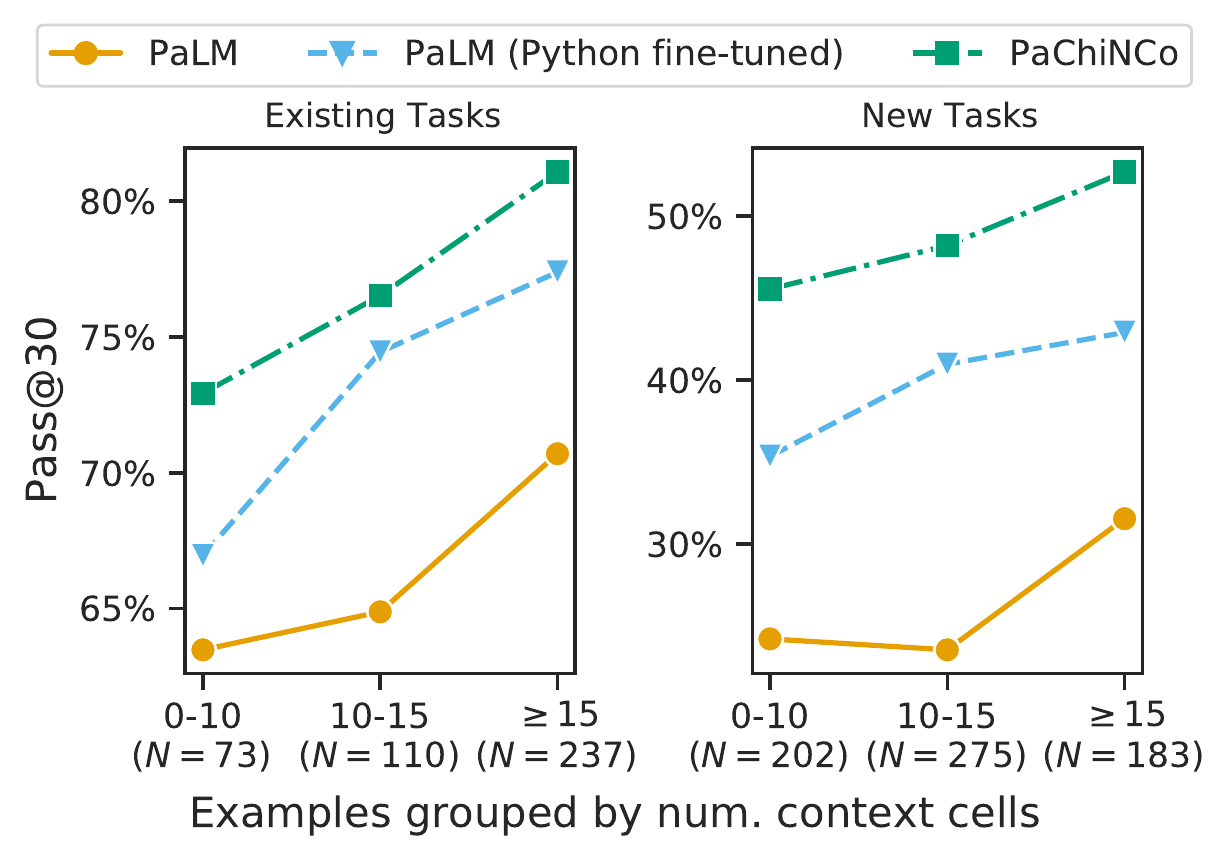}
    \caption{Pass rate on examples grouped by context size. We also report the number of examples in each group.}
    \label{fig:exp:ctx_size:split_by_ctx_cell_num}
\end{figure}

\paragraph{Does Problem Location Impact Performance?}
Another interesting angle to study the effect of context is through the lens of  model accuracy when solving problems $c_k$ at different locations.
Intuitively, problems located later in a notebook ($k$ is larger) would have more context available, therefore they could be easier to answer~\cite{wang-cho-2016-larger}.
\cref{fig:exp:ctx_size:split_by_ctx_cell_num} shows \passat{30} on problems grouped by their preceding context size, which shows increased task success rate when solving problems with more context, confirming the prior intuition.\footnote{We remark that our setup is different from multi-turn semantic parsing where later turns are conditioned on the \emph{predictions} of prior turns, while we use reference solutions (\cref{sec:experiments:prompting_method}). See \cref{sec:related_works} for discussion.}

\subsection{Few-shot Prompting Beyond Test Notebook Contexts}
\label{sec:exp:few_shot_prompting}
Our previous set of experiments demonstrate diminishing returns after including more notebook context cells.
As motivated in \cref{sec:experiments:prompting_method}, besides test notebook context, we use few-shot prompting with additional exemplars to teach the model to perform this challenging task, while generating code in different styles.
Here, we report both functional correctness (\cref{fig:exp:prompting:passatk}) as well as metrics evaluating the style of model-predicted code (\cref{fig:exp:prompting:code_style}) for problems in \newtasks/.
We only evaluate \nbmodel/ due to that the prompt length (maximal $2,100$ sub-tokens) exceeds the limit of public code LMs.

With \textbf{Step-by-Step Prompting} ($\bm{\mathrm{SbS}}$), we observe an improvement on \passat{30} over the baseline using only notebook contexts (\cref{tab:exp:end2end_results}). 
This is consistent with the general results on other tasks when comparing few- vs zero-shot learning with LLMs~\cite{brown2020language}, while we remark that in our case, the ``zero''-shot prompting setting (\cref{tab:exp:end2end_results}) still contains  notebook cells as rich contexts.
Nevertheless, as \cref{fig:exp:prompting:passatk} suggests, $\mathrm{SbS}$ prompting is quite helpful for improving pass rate.
Empirically, we observe $\mathrm{SbS}$ prompting is especially helpful for problems (at code cells $\cell_k$) without adequate preceding notebook context ($k$ is small),
For instance, $\mathrm{SbS}$ prompting results in 6$\%$ absolute improvement for the first two rounds of problems. 
Interestingly, even if we include more test notebook context in the non-exemplar baseline such that the prompt length match with $\mathrm{SbS}$ prompting using extra exemplars ($\bm{\mathrm{Baseline}+\mathrm{More~Context}}$), $\mathrm{SbS}$ prompting is still better, which again suggests the value of extra exemplars to compliment the information in notebook context.

In line with our intuition, $\mathrm{SbS}$ prompting yields code predictions that are better decomposed into more lines of code (LoC$_\uparrow$), with each line being simpler (Tokens/API per Line$_\downarrow$).
It also appears to generate slightly more complex solutions, as indicated by the increased {\tt pandas} API usage.
Our finding demonstrates the effectiveness of $\mathrm{SbS}$ prompting in improving accuracy and also code style.
In contrast, we also report results from prompting the model using exemplars with ``vanilla''-styled code following the common practice of chaining multiple pandas API calls in a single line ($\bm{\mathrm{Vanilla~Code}}$), which only leads to marginal improvements on \passat{k} over the non-exemplar baseline, while the code style remains consistent.

\begin{figure}[t]
    \centering
    \subfloat{%
        \includegraphics[width=0.95 \columnwidth]{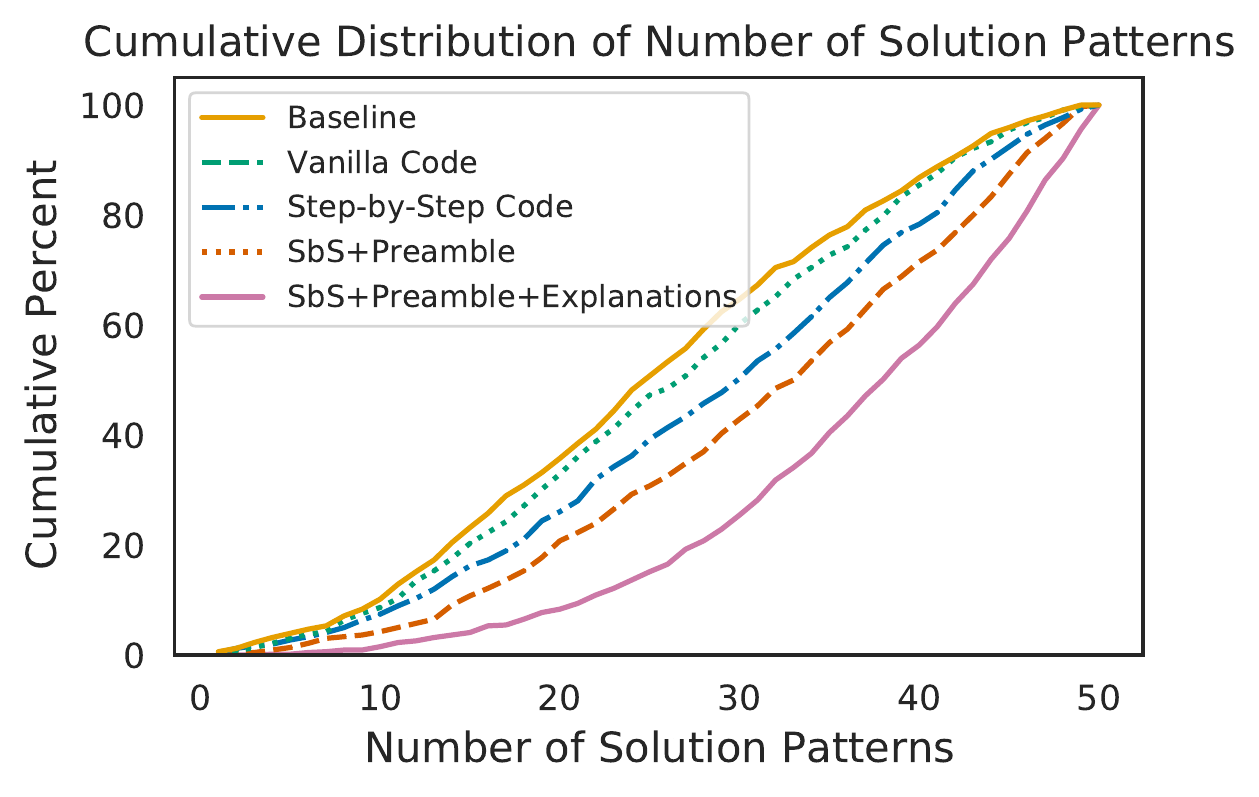}%
    }
    
    \subfloat{%
        \includegraphics[width=0.95 \columnwidth]{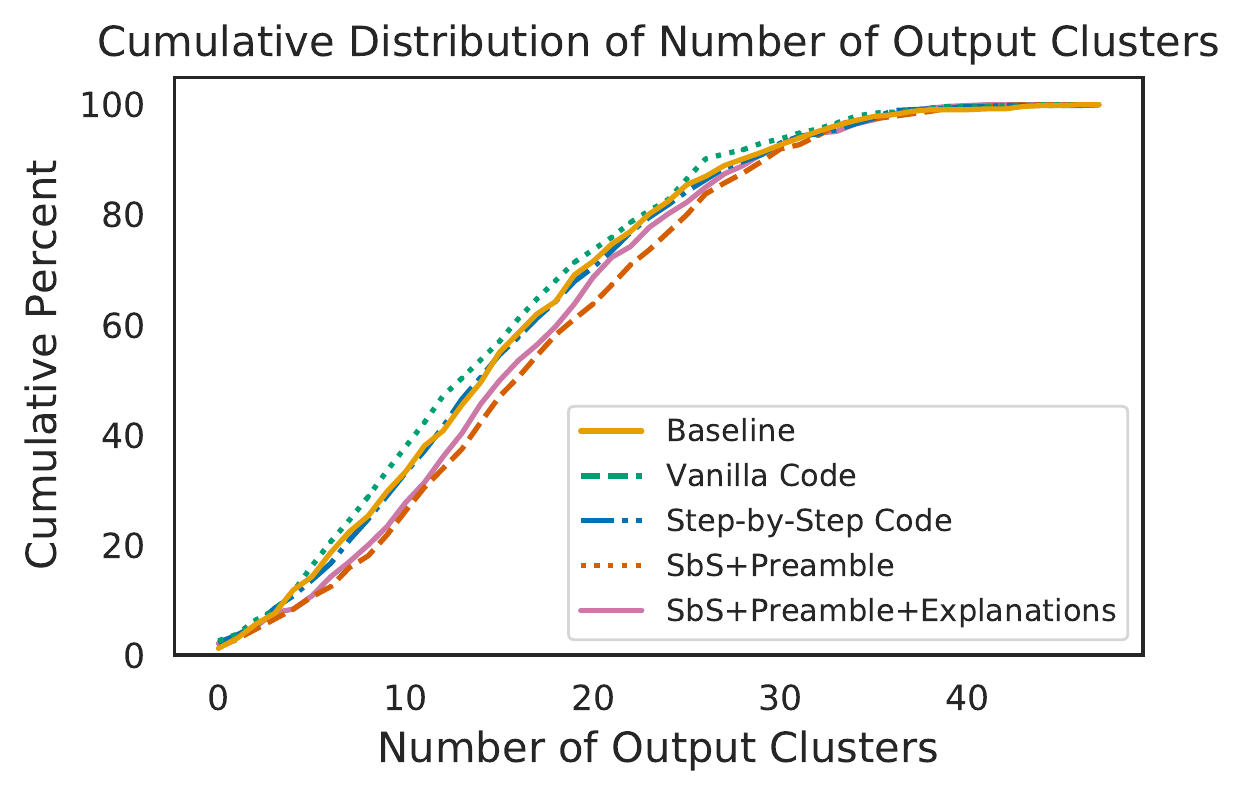}%
    }
    \caption{Cumulative distributions of the number of unique API sequences (top) and output clusters (bottom) extracted from \nbmodel/'s $50$ predictions on the \textit{New Tasks} split. Curves that appear more to the right represent prompting methods with greater diversity in the samples. Step-by-step prompting leads to much greater diversity than the baselines.}
    \label{fig:exp:prompting:solution_diversity_hist}
\end{figure}

Next, on top of $\mathrm{SbS}$ prompting, using a preamble at the beginning of a target cell to further encourage the model to solve the problem in a step-by-step fashion ($\bm{\mathrm{SbS}+\mathrm{Preamble}}$) improves \passat{k}, as well as the level of decomposition in predicted code (LoC$_\uparrow$, Tokens/API per Line$_\downarrow$).
This finding is inline with \citet{Kojima2022LargeLM} where using preambles to nudge the model to generate reasoning steps can still be useful together with few-shot step-by-step demonstrations.
More surprisingly, with additional inline NL explanations for each step ($\bm{+\mathrm{Preamble}+\mathrm{Explanation}}$), \nbmodel/ produces even more decomposed code while maintaining similar accuracy.
Those predictions also come with rich NL comments as explanations, with the number of comment lines nearly equal to the number of code lines.
Intuitively, step-wise NL explanations could be potentially helpful for developers to better understand predictions from code LMs, as we demonstrate later in \cref{sec:prompting_case_study}.

We also report breakdown results of pass rate on problems with varying level of complexity.
$\mathrm{SbS}$ prompting and its variants are helpful across the board, especially for harder tasks with more than 7 {\tt pandas} function calls.
This might suggest the value of step-by-step decomposition when synthesizing complex programs.
Next, we provide more insight into the effectiveness of $\mathrm{SbS}$ prompting.

\begin{figure}[t]
    \centering
    \includegraphics[width=\columnwidth]{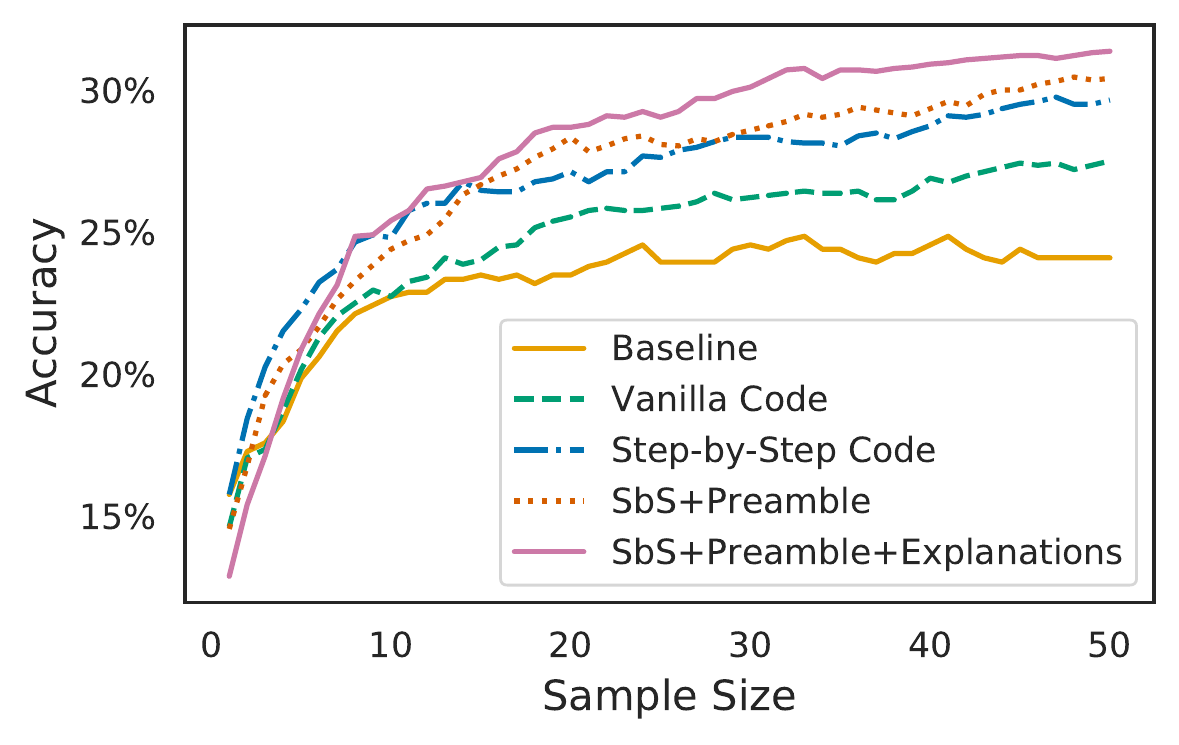}
    \caption{Self-consistency decoding accuracy (using 1 reranked sample) on \textit{New Tasks}}
    \label{fig:exp:prompting:clustering:pass_rate}
\end{figure}

\paragraph{Step-by-Step Prompting Produces More Diverse Solutions}
Given the effectiveness of $\mathrm{SbS}$ prompting in improving accuracy, code style, and explainability (\cref{sec:prompting_case_study}), next we explore why it is helpful by analyzing patterns in model-predicted solutions.
One interesting question is whether $\mathrm{SbS}$ prompting could produce more diverse solution approaches to solve a problem. 
Intuitively, greater diversity in the output space could improve the odds of finding a solution at higher sample size $k$.
Determining whether two solutions are ``different'' is difficult and subjective, but we approximate this from two angles. First, we use the sequence of {\tt pandas} API calls as a signature of the high-level solution pattern.
Second, since different solutions might have the same functionality (\ie, they execute to the same output), we also measure the diversity in \emph{functionality} by clustering predictions based on their outputs.\footnote{Here, each cluster corresponds to programs that executes to the same exact results (\ie~no fuzzy matching as in \cref{sec:dataset:evaluation_metric}).}
\cref{fig:exp:prompting:solution_diversity_hist} plots the cumulative distributions of the number of unique solution patterns and output clusters on the \newtasks/ split.
Step-by-step prompting results in greater diversity on both metrics compared to the baselines.
Interestingly, prompting with NL explanations yields even more solution patterns.
Having diverse predictions could be potentially useful to handle under-specified intents (\cref{sec:dataset:analysis}), as those predictions could correspond to different interpretations of the ambiguous intents.
Empirically, we find diverse predictions are particularly effective for post-hoc reranking such as self-consistency decoding, where we show the user one prediction from the largest output cluster instead of showing all $k$ predictions.
As shown in \cref{fig:exp:prompting:clustering:pass_rate}, $\mathrm{SbS}$ prompting significantly improves over baselines when using self-consistency decoding ~\cite{wang2022selfconsistency}.
Notably, the $1$-sample accuracy of $\mathrm{SbS}$ with NL explanations is on par with \nbmodel/'s \passat{5} accuracy (last row,  \cref{tab:exp:end2end_results}).
We present further analysis of solution diversity from \nbmodel/ in \cref{sec:app:solution_diversity}.

\section{Error Analysis}
\label{sec:error_analysis}

\begin{figure}[t]
    \centering
    \small
    \includegraphics[width=\columnwidth]{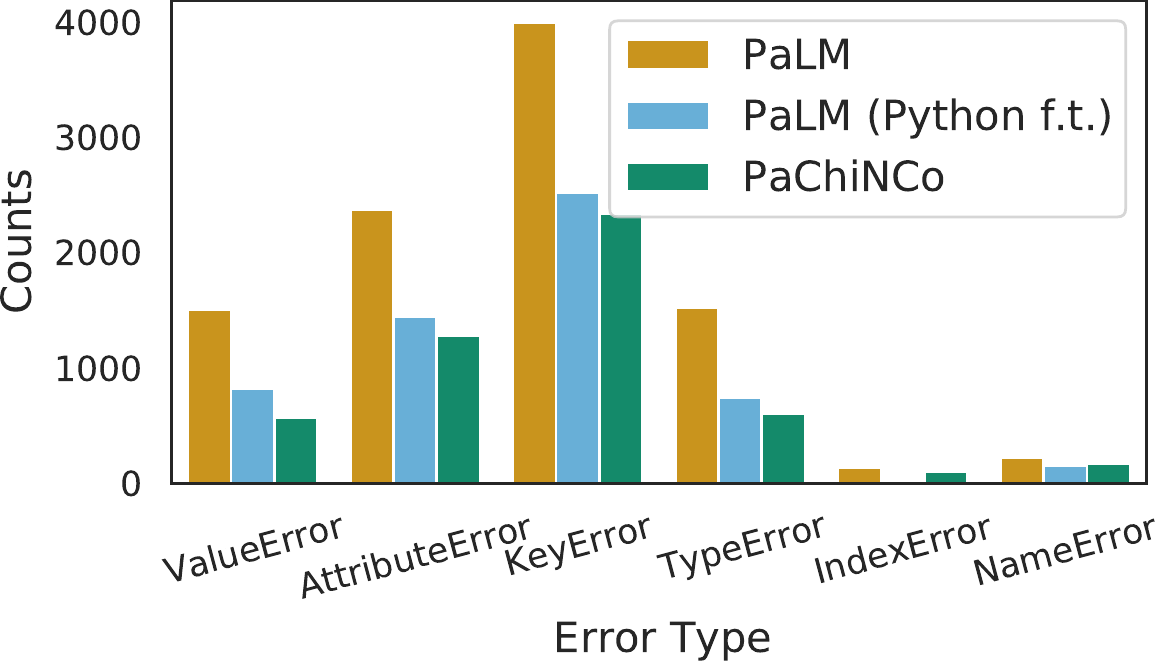}
    \caption{Frequency of execution errors on \newtasks/.}
    \label{fig:exp:runtime_error_distribution}
\end{figure}

To understand the types of errors that LLMs make on \dataset/, especially on challenging problems, we conduct an error analysis on \palm/'s predictions on the \newtasks/ split (\cref{tab:exp:end2end_results}).
Overall, we notice a significant drop in execution errors after two-stage code fine-tuning (\palm/$\mapsto$\nbmodel/, \cref{sec:model}).
Out of all the incorrect predictions from \nbmodel/ under the fuzzy output matching evaluation metric (\cref{sec:dataset:evaluation_metric}), roughly $35\%$ of the samples result in execution errors, while the remaining $65\%$ predictions have executable programs but are functionally incorrect.
First, for un-executable samples, we present an analysis of the distribution of different execution error types, as illustrated in \cref{fig:exp:runtime_error_distribution}.
The primary source of execution error is {\tt KeyError} and {\tt AttributeError} due to reference to non-existing index or columns in {\tt DataFrame}s.
While in the prompts we provide NL descriptions of schema for {\tt DataFrame}s loaded to notebooks (\cref{sec:experiments:prompting_method}), such descriptions for intermediate {\tt DataFrame}s that are later derived in the context are still missing due to limited prompt length, and the model may not be able infer their schema information solely from the source code.
This could be especially problematic for APIs that create compound intermediate {\tt DataFrame}s with complex schema, such as {\tt pd.groupby}, which accounts for that more than $50\%$ of those {\tt KeyError}s and {\tt AttributeError}s.
Similarly, other execution errors such as {\tt ValueError} and {\tt TypeError} are often caused by the insufficient knowledge about the {\tt DataFrame} contents.
For example, {\tt ValueError} occurs when a model tries to calculate the mean of a column which has {\tt NaN} values.
This finding suggests the importance of developing LLMs that could handle longer context \cite{Transformers2022MET} in order to include more {\tt DataFrame} information in prompts. 
We gave a detailed case study on these types of execution errors in \Cref{app:error_examples}.

Next, we conduct a manual analysis on $50$ randomly sampled incorrect predictions (with a focus on executable but incorrect predictions).
The cause of these errors can be grouped into the following categories:
1. Complex problems requiring non-trivial reasoning or data transformation steps ($43\%$); %
2. Errors in interpreting NL intents, such as missing a requirement specified in the intent (\eg~\utterance{round to two decimal places}) in the code solution ($26\%$);
3. Errors caused by under-specified intents (\cref{sec:dataset:analysis}, $19\%$);
4. False-negatives due to limited coverage of the fuzzy-matching evaluation metric  (\cref{sec:dataset:evaluation_metric}, $6\%$);
5. Annotation errors ($6\%$).

The primary source of errors is due to complex problems, which reiterates the motivation of \dataset/ --- evaluating code LLMs on challenging data wrangling and EDA tasks.
The second majority type of errors (misunderstanding intents) suggests room to improve \nbmodel/'s skill in instruction following.
Next, a non-trivial amount of errors are caused by under-specified intents, which are common in the setting of prompting LLMs using short instructions (\cref{sec:dataset:analysis}), calling for future research to specifically address this issue.
Finally, our evaluation metric based on fuzzy output matching seems effective in identifying plausible alternative solutions.
Still, there are non-trivial cases where there are multiple ways of presenting the outputs (\eg~{\tt DataFrame}s with nested columns or different orientations, \cref{fig:errorexample:evalcoverage}). 
We present more detailed examples of these errors in \cref{sec:app:error_breakdown_wrt_ctx_size}.

\section{Case Study: How Useful is Predicted Code with Step-wise Explanations?}
\label{sec:prompting_case_study}

In \cref{sec:experiments} we show how prompting \nbmodel/ to generate code with step-wise explanations is helpful in improving solution diversity (\cref{fig:exp:prompting:solution_diversity_hist}) and also accuracy of self-consistency reranking (\cref{fig:exp:prompting:clustering:pass_rate}).
In this section we conduct a qualitative analysis on model-generated code snippets with explanations to explore how they could be useful for novice data scientists.

First, we observe that NL explanations could help users follow the flow of complex data transformations for programs involving a chain of {\tt pandas} operations.
By decomposing and explaining how data is manipulated after individual transformation steps, it is easier for users to understand the solution and track its dataflow behind the scene, especially when some steps involve complex computation (\cref{fig:explanation_study:complex_single_step}), or the underlying schema is less intelligible (\eg, column names with abbreviations, \cref{fig:explanation_study:meaningfulname_to_schema}).
Additionally, some inline explanations also describe the output of intermediate steps, which is particularly helpful when these steps involve advanced {\tt pandas} functions whose output may not be obvious, such as {\tt pd.unstack} (\cref{fig:explanation_study:explain_complex_api})

Meanwhile, step-wise NL explanations serve as high-level procedural descriptions of code, which enables users to easily browse through and understand different solution approaches without being distracted by nuances in the actual code implementation (\cref{fig:explanation_study:multi_sol}).
Moreover, explanations also help users verify the code solutions by identifying potentially incorrect steps and make necessary corrections (\cref{fig:explanation_study:identify_incorrect_sol}).
The observations presented here offer insight into potential future avenues to improve the utility of code LMs for developers through the use of step-by-step explanations, which we leave as important future work.

\section{Related Works}
\label{sec:related_works}

\paragraph{LLMs for Code}
In recent years there has been a burgeoning of LMs trained on code~\cite{chen2021codex,austin2021lambdacode,nijkamp2022codegen,fried2022incoder,Li2022CompetitionLevelCG,tunstall2022natural,Xu2022ASE}.
Many generalist LMs also have source code as part of their training data mixture~\cite{Thoppilan2022LaMDALM,chowdhery2022palm,gpt-j,gpt-neo}.
Those LLMs have registered impressive performance on a variety of benchmarks for code, such as code translation~\cite{Lachaux2020UnsupervisedTO}, program repair~\cite{Gupta2017DeepFixFC}, and natural language to code generation~\cite{chen2021codex,hendrycksapps2021,austin2021lambdacode,Li2022CompetitionLevelCG}.

\paragraph{Automating Data Science} Given the sheer amount of tasks involved in the lifecycle of data science and the expertise required, it calls for development of systems to automate this lifecycle~\cite{aggarwal2019hicanai,wang2021autods,wang2021howmuchautomationds}.
As an example, automating the process of feature engineering and finding the best-performing models has been the central theme of AutoML research~\cite{he2021automl,Karmaker2020AutoMLTD}, with well-established systems~\cite{feurer2015efficient} and evaluation protocols~\cite{zoller2021benchmark}.
In this paper we focus on automating data wrangling and EDA tasks, which account for nearly the same amount of code and documentations in notebooks as that for feature engineering and building ML models~\cite{agashe-etal-2019-juice,wang2022documentation}.
Along this line of research,
\textsc{AutoPandas}~\cite{Bavishi2019AutoPandasNG} synthesizes {\tt pandas} programs for data transformation given examples of input/output {\tt DataFrame}s.
Other approaches aim to generate {\tt pandas} programs using LLMs using both NL intents and I/O samples \cite{Jain2021JigsawLL} or unit tests \cite{chandel2022msdsp}.
In this paper we consider code generation in notebook environments with multiple rounds of problems and interdependent contexts (refer to \cref{sec:dataset:analysis} for detailed discussions of recent work).
Related to this line, another thread of research focuses on synthesizing problems for visualization plots~\cite{Amar2005LowlevelCO,Narechania2020NL4DVAT,Fu2020QudaNL,Chen2021PlotCoderHD,Wu2022NL2VizNL}.
\dataset/ also includes 57 plotting examples which are not used in this paper. We leave collection and systematic evaluation of plotting tasks as important feature work.

\paragraph{Context-driven Code Generation}
Our work is another application of context-driven code generation, which concerns with mapping a series of contextually dependent utterances to executable programs, such as domain-specific logical forms \cite{Zettlemoyer2009LearningCM,Long2016SimplerCL,Iyyer2017SearchbasedNS}, SQL queries over databases \cite{Suhr2018LearningTM,Yu2019CoSQLAC,Yu2019SParCCS}, or general-purpose languages (PLs) like Python \cite{nijkamp2022codegen}.
Some of those existing works also feature EDA problems, mainly for querying structured databases, while the problems in \dataset/ exhibit more complex query patterns (\eg~string manipulation over semi-structured data, as $u_1$ in \cref{fig:intro:teaser_example}), in addition to extra data wrangling tasks, thanks to the expressiveness of {\tt pandas} API and the general-purpose PL (Python) compared to domain-specific formalisms.
As the first step forward, in this paper we perform evaluation using ground-truth solutions for previous turns (\cref{sec:experiments:prompting_method}), similar to multi-turn program synthesis for task-oriented dialogue in \citet{Andreas2020TaskOrientedDA}.
We leave conditioning on model-predicted solutions as context for evaluation \cite{nijkamp2022codegen} as an important future avenue.

\section{Conclusion}
In this paper we present \dataset/, a code generation benchmark for data wrangling and EDA tasks in computational notebooks.
\dataset/ features problems with realistic NL intents and rich notebook contexts.
We also develop \nbmodel/, a $62\mathrm{B}$ LM tailed for data science, and show that \nbmodel/ outperforms public code LMs on \dataset/, while being effective in few-shot learning to improve code style and solution diversity.

\section*{Acknowledgements}
We are grateful to Meg Risdal and Goeff Thomas from Kaggle for help with dataset collection, and Miltos Allamanis for research discussion.
We thank Jo Chick from the research partnership team, and Rebecca Watson, Ashley Dawe and Kimberly Herrera from Upwork to help with managing the annotation project.
We thank Aroma Mahendru for writing the data card section.
We also thank Cheriskumar Patel, Preet Patel, and Jayendra Parmar for general assistance with the project.

\bibliography{main}

\clearpage
\newpage
\onecolumn
\appendix

\begin{center}
\Large
\textbf{Supplementary Materials}
\end{center}

\section{Outline of \dataset/ Annotation Guideline}
\label{app:annotation_guideline}

In this section we provide a brief summary of the outline in our annotation guideline.

\paragraph{\existtasks/}
The annotators are given a list of Jupyter notebooks. Each notebook uses {\tt pandas} to perform certain data analysis tasks. For each notebook, an annotator is asked to:

\begin{enumerate}
    \item Identify code cells that contain instructive code snippets that perform data wrangling or exploratory data analysis tasks.
    \item Fix the notebook and make them clean and executable.
    \item For each code snippet identified in Step 1, create natural language descriptions of the task.
    Also verify the code solution and fix them as appropriate. Finally, remove any redundant text in the notebook (\eg~solution outline or hints for tutorial notebooks) that could give away to the refernce solution. 
\end{enumerate}

\paragraph{Instruction on Creating Natural Intents}
Specifically, for step 3, in order to collect realistic NL intents, the annotators are given the following high-level description, followed by detailed instructions and examples.

\begin{cite-comment}
Below we share some suggestions to write good intents. \\

Keep it natural without redundant explanations. Imagine an AI programmer can help you accomplish simple data wrangling and EDA tasks, what kind of intents will you send to the system? Our goal is to collect real inputs to such a system from data scientists like you. \\

One idea to write good intents is to keep it concise such that another programmer could quickly understand and implement a solution that executes to the same outputs. You are encouraged to create simple, short intents while describing the desired outputs without much ambiguity. 
\end{cite-comment}

\paragraph{\newtasks/}
For each ML dataset we provided, an annotator creates a Colab notebook with code snippets for some interesting data wrangling and exploratory data analysis tasks using this dataset.
Each code snippet is paired with its natural language intent, simliar to the process of annotating \existtasks/. We ask annotators to feel free to work on any tasks that they may find interesting for the given dataset, as long as the code solution for the task should consist of multiple lines and use different pandas API functions. Different from annotating \existtasks/, we ask them to first create a natural language intent for their task, and then write a code solution in the next cell. 

Below is an excerpt from the annotation guideline describing the types of data wranling and EDA tasks to create.

\begin{cite-comment}
\textbf{What Tasks to Create} \\

In general, you may create whatever exploratory data analysis tasks that you find interesting for the given datasets. To come up with interesting tasks, you can think in this way: before training your ML models for the dataset, what kind of data wrangling or EDA tasks would you like to perform on the dataset? Below are some more concrete descriptions of such wrangling or EDA tasks: \\

\textbf{Data Preprocessing/Wrangling Tasks} which involves modifying existing dataframes or creating new ones. Such as normalizing column names, adding new columns, modifying existing columns (e.g., converting string values to date times), generating new dataframes using ops like {\tt group\_by}, and so on. Some datasets we shared are just raw data without any preprocessing or cleaning. Feel free to . Please also refer to Section: Identify Code Snippets to Annotate  in our previous annotation guideline for more examples. \\

\textbf{Exploratory Data Analysis Tasks that Require Some Wrangling and Preprocessing} Answering interesting EDA questions using the given dataset, but some data wrangling steps are required in order to derive the answer. For example, given a dataframe df of user shopping history and credit card expiration dates in the format of {\tt df.loc[0][‘cc\_exp’] = ‘08/26’}. To answer the EDA question “How many users have a credit card expiring in 2024?”, we need to first convert the expiration year from the string-formatted {\tt cc\_exp} column.

\end{cite-comment}

To encourage the annotators to create more complex tasks, we also provide the following high-level instruction:

\begin{cite-comment}
\textbf{Complexity of Tasks} \\

You should create relatively complex tasks that require multiple steps and also a combination of different pandas APIs to solve them. Avoid problems that can be solved using one-liner code such as {\tt df.group\_by(...).sort\_values(...)}. An ideal task should be reasonably complex and needs to be broken down into multiple smaller steps to solve, and each step may require using one or multiple pandas functions. \\

As a general rule of thumb, you should aim at creating tasks that either have at least 50 tokens or use at least 4 pandas APIs (dataframe/series indexing, like {\tt df[df[‘continent’] == ‘NA’]} is also counted as one API usage). You can find more concrete example tasks at the end of this doc.
\end{cite-comment}

\paragraph{Full Guideline}
Our annotation guideline is 35-pages long in total, which we provide on a per request basis. Please contact \url{pcyin@google.com} to request access.

\section{Details of Fine-tuning \nbmodel/}
\label{app:palm_fine_tuning}
We take the fully trained \palm/ 62B model that has been trained on $1,325\mathrm{B}$ tokens~\cite{chowdhery2022palm} as the base model. To achieve the best performance on , we finetune the base model in two stages: a python only and a ARCADE only finetuning. 
For the two-stage code fine-tuning (\cref{sec:model}), we use the similar training recipe of the base \palm/ 62B model.
More specifically, we apply the learning rate decay scheduling of $0.2 / \sqrt{t}$ where $t$ is the number of steps.
We train the model for 124K steps (1 epoch) with a batch size of 256 on 512 TPU v4 chips.
Afterwards, we reload the optimizer state and continue training on the Jupyter notebooks data ($10\mathrm{B}$ tokens) using the same hyper parameter for 3 epochs ($\sim 572\mathrm{K}$ steps).

\paragraph{Linearize Notebooks to Python Source Code}
We convert computational notebooks for finetuning (\cref{sec:model}) and evaluation (\cref{sec:experiments:prompting_method}) to Python source code using {\tt nbconverter}.
Specifically, markdown and code cells in a notebook are concatenated using the special delimiter `{\tt \# In[]:}', and text in markdown cells is commented out using the `\# ' prefix.
See \cref{lst:prompt:notebook_ctx} for an example of the linearized notebook for \cref{fig:intro:teaser_example} (up to $\cell_3$).
Jupyter notebooks that are converted to Python files in such format are common in GitHub repositories, which mitigates the domain transfer gap between general Python code and notebook-specific data, and also allows us to prompt public code LLMs, which might have seen similar data during pre-training.

\section{Performance of Python-finetuned \palm/ 62B on Code Benchmarks}
\label{app:palm_python_finetuned_performance}

We evaluate the \palm/ $62\mathrm{B}$ model after the first-stage fine-tuning on Python data (\cref{sec:model}) on other code generation benchmarks, namely HumanEval~\cite{chen2021codex}, MBPP~\cite{austin2021lambdacode} and Transcoder~\cite{lachaux2020unsupervised}, following the evaluation setup of  \palm/-\textsc{Coder} in \citet{chowdhery2022palm}.
\cref{table:palm_python_finetuned:eval_results} lists the results.
With $7 \times$ more Python code tokens, our Python fine-tuned \palm/ $62\mathrm{B}$ model outperforms the $8 \times$ larger \palm/-\textsc{Coder} $540\mathrm{B}$ model on all the benchmarks, including the multilingual code translation task.

\begin{table}[h!]
    \centering
    \small
    \begin{tabular}{ l c c c  } 
    \toprule
    \textbf{Dataset} & Human Eval & MBPP & Transcoder \\ \midrule
    \textbf{Metric} & \passat{100} & \passat{80} & \passat{25}  \\
    \palm/-\textsc{Coder} $540\mathrm{B}$ \cite{chowdhery2022palm} & $88.4\%$ & $80.8\%$ & $82.5\%$ \\
    \palm/ $62\mathrm{B}$ Python (\cref{sec:model}) & $\mathbf{91.5\%}$ & $\mathbf{86.0\%}$ & $\mathbf{86.4\%}$ \\ 
    \bottomrule
    \end{tabular}
    \caption{Python fine-tuned PaLM 62B model evaluation results on public code benchmarks.}
    \label{table:palm_python_finetuned:eval_results}
\end{table}

\section{Inference Setup}
\label{app:inference_setup}
For \codegen/, we use the inference script from the official GitHub repository. For \incoder/, we follow the official inference example script and use the release on Huggingface model hub.
We convert each example in our dataset to Python source code to a prompt, as outlined in \cref{sec:experiments:prompting_method}.
Notebooks are linearized using {\tt nbconverter} similar as generating fine-tuning data (\cref{app:palm_fine_tuning}).
One exception is \incoder/, for which we follow \citet{fried2022incoder} and use the Jupyter notebook linearization template used in its pre-training.

At inference time, we left-truncate notebook context up to 900 tokens (measured by \palm/'s vocabulary), which fit in the context window size of all LLMs we evaluated.
We also make sure to always include NL schema descriptions in prompts given their importance in understanding NL intents.
In addition, for few-shot experiments in \cref{sec:exp:few_shot_prompting}, we use additional 1,200 tokens to accommodate the prompt prefix, making the total maximal prompt length to be 2,100.
Due to its excessive length, we only perform few-shot prompting experiments on \nbmodel/ since its rotatory positional embedding \cite{Su2021RoFormerET} could generalize to encode longer contexts at inference time. 
During sampling, we set the maximum target length to be 512 tokens.

\section{\passat{1} for \nbmodel/}
\label{sec:app:pass_at_one}

\begin{table}[h]
    \centering
    \begin{tabular}{cc}
        \toprule
        \existtasks/ & \newtasks/ \\ \midrule
        48.8 & 17.5 \\ \bottomrule
    \end{tabular}
    \caption{\passat{1} for \nbmodel/ (temperature=0.2)}
    \label{tab:pass_at_one}
\end{table}

For reference, we also report \passat{1} of \nbmodel/ on \dataset/. Samples are generated using the same hyper parameters as in \cref{sec:exp:results}, except for the temperature, which is 0.2.

\section{Scaling Curve of \codegen/ on \dataset/}
\label{sec:app:dataset_model_scaling_curve}

\begin{figure}[h!]
    \centering
    \includegraphics[width=0.5 \columnwidth]{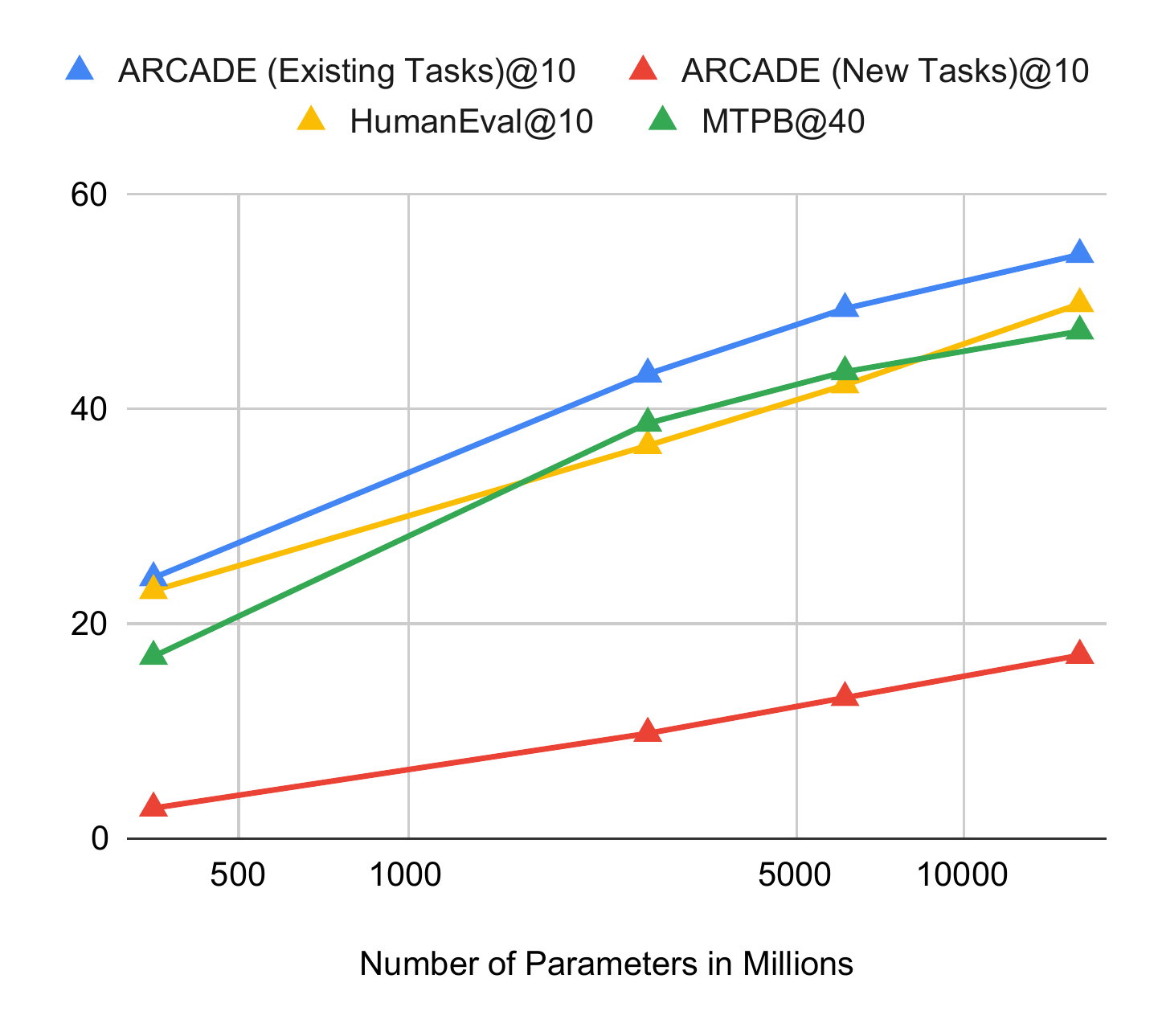}
    \caption{Scaling curve of \codegen/ on \dataset/ and existing code generation benchmarks. Results on HumanEval are selected from the best temperature $t \in \{0.2, 0.6, 0.8\}$ \cite{nijkamp2022codegen}.}
    \label{fig:exp:dataset_model_scaling_curve}
\end{figure}

\cref{fig:exp:dataset_model_scaling_curve} depicts the scaling curve of \dataset/ with respect to the number of parameters for \codegen/$_\textrm{mono}$ models.
The pass rate scales nearly linearly as a function of model size, and the performance has not saturated, especially on the \newtasks/ split.
This shows \dataset/ is a reliable dataset to study the scaling behavior of code LLMs.
The slope of the curve on \newtasks/ is also smaller than on other datasets, suggesting that this problem set is more challenging for \codegen/ models.

\section{Comparing with \textsc{Codex} on \dataset/}
\label{app:codex_results}

\begin{table*}[ht]
    \small
    \centering
    \begin{tabular}{l cccc : cccc}
    \toprule
        \multirow{2}{*}{\passat{k}} & \multicolumn{4}{c}{\bf \textit{Existing Tasks}} & \multicolumn{4}{c}{\bf \textit{New Tasks}} \\
         & 5 & 10 & 20 & 30 & 5 & 10 & 20 & 30 \\ \midrule
        \nbmodel/ ($62\mathrm{B}$) & $64.6$ & $71.0$ & $76.0$ & $78.3$ & $30.6$ & $38.0$ & $45.0$ & $48.6$ \\ \hdashline
        \textsc{Codex}-cushman-001 (12B) & $51.0$ & $59.0$ & $65.9$ & $69.4$ & $14.2$ & $20.1$ & $27.0$ & $31.3$ \\
        \textsc{Codex}-davinci-002 & $66.1$ & $72.6$ & $78.3$ & $81.2$ & $36.0$ & $43.8$ & $50.8$ & $54.8$ \\
    \bottomrule
    \end{tabular}
    \caption{\emph{Pass}$@k$ evaluation on \dataset/ using \textsc{Codex} API.}
\end{table*}

For reference, we also report the results on \dataset/ using the public \textsc{Codex} API, following the same evaluation setting as in \cref{tab:exp:end2end_results}.
\nbmodel/ significantly outperforms the smaller \textsc{Codex}-cushman-001 API, on par with \textsc{Codex}-davinci-002 on \existtasks/, while \textsc{Codex}-davinci-002 is stronger on \newtasks/.
While we cannot gain much insight from the results due to limited knowledge about \textsc{Codex}-davinci-002, through error analysis, we find that \textsc{Codex}-davinci-002 is better at instruction following, especially in understanding NL descriptions of complex {\tt DataFrame} schema (\cref{sec:experiments:prompting_method}).
We leave improving the instruction following skills of \nbmodel/ as interesting future work.

\section{Performance on Problems with Different Complexity}
\label{app:pass_at_k_wrt_task_complexity:ast}

\begin{figure}[h]
    \centering
    \includegraphics[width=0.6 \columnwidth]{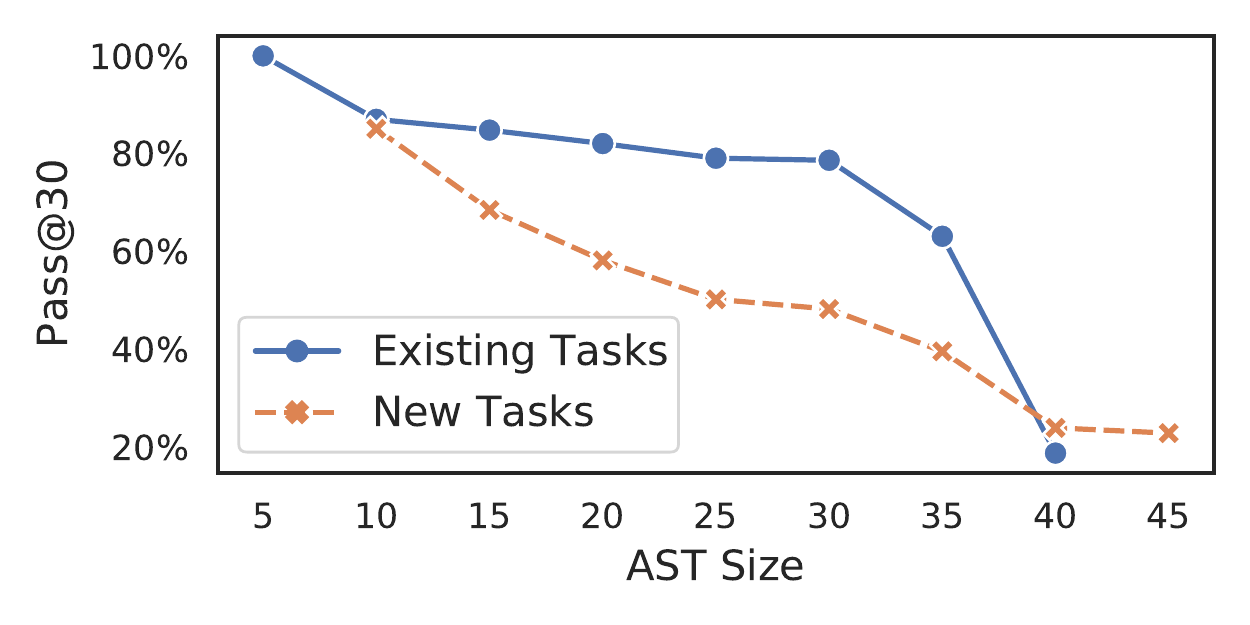}%
    \caption{\passat{k} of \nbmodel/ w.r.t~AST size of reference programs.}
    \label{fig:exp:pass_at_k_wrt_task_complexity:ast}
\end{figure}

\cref{fig:exp:pass_at_k_wrt_task_complexity:ast} plots \passat{30} with respect to the AST size of reference programs. Similar to \cref{fig:exp:pass_at_k_wrt_task_complexity}, results on \newtasks/ are generally lower.
Meanwhile, it seems that AST size correlates better with \passat{k} compared to the number of API usage, while the latter metric offers more intuitive information about the data transformation steps involved. 

\section{Distribution of Execution Error Types w.r.t~Context Size}
\label{sec:app:error_breakdown_wrt_ctx_size}

\begin{figure}
    \centering
    \includegraphics[width=0.8 \textwidth]{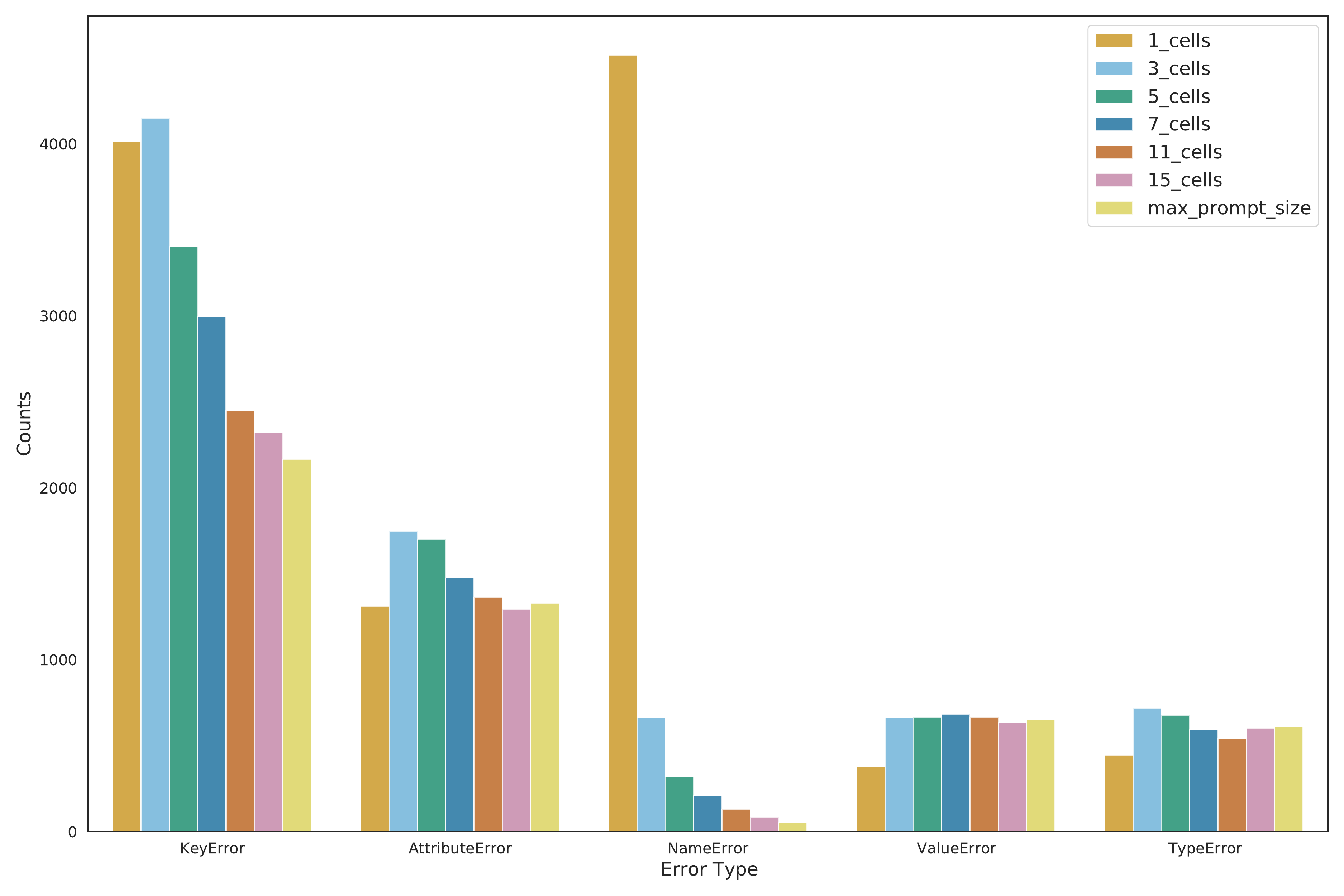}
    \caption{Frequency of runtime errors from \nbmodel/'s predictions aggreated over the two splits with varying amount of notebook context cells.}
    \label{fig:freq_execution_error_with_ctx_size}
\end{figure}

In \cref{fig:exp:ctx_size:split_by_ctx_cell_num} we observe that \passat{k} improves when more notebook context is available.
\cref{fig:freq_execution_error_with_ctx_size} illustrates the distribution of execution error types on failed predictions.
Notably, using more notebook context cells significantly reduces the chance of {\tt NameError}s caused by using undefined variables  in context.
The number of {\tt KeyError}s is also reduced, indicating that the model makes fewer schema understanding errors when referring to columns in {\tt DataFrames}. 

\section{Few-shot Prompting Results on \existtasks/ Split}
\label{sec:app:few_shot_prompting:existing_tasks}

\begin{figure*}[t]
    \centering
    \includegraphics[width=\textwidth]{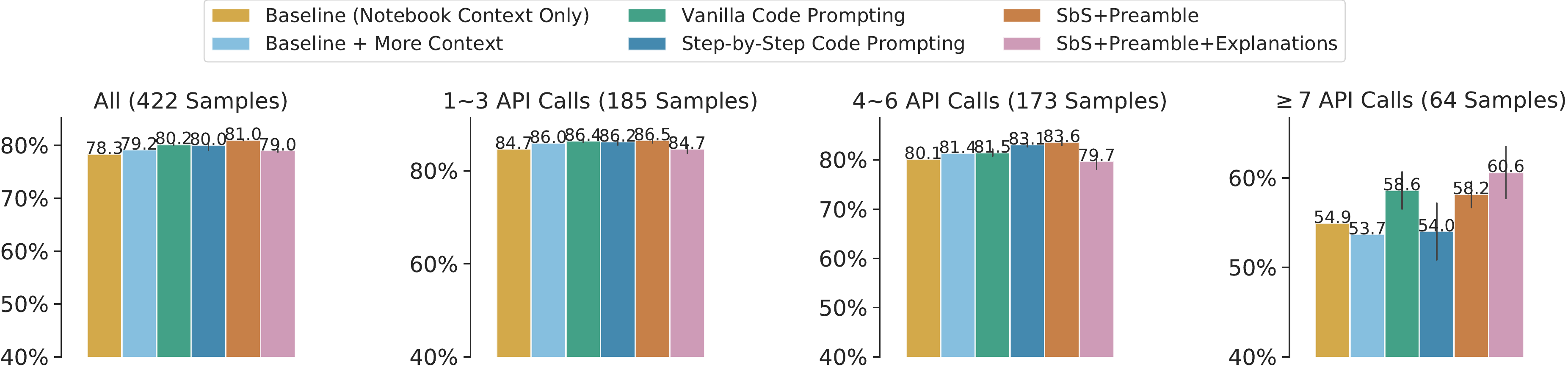}
    \caption{\passat{30} for few-shot prompting on \existtasks/}
    \label{fig:few_shot_prompting:existing_tasks}
\vspace{1em}
    \centering
    \includegraphics[width=\textwidth]{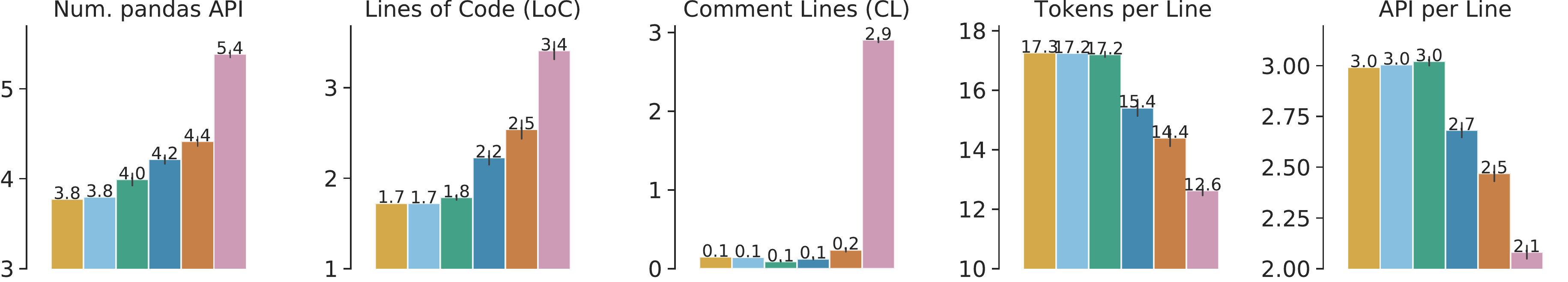}
    \caption{Code style metrics for few-shot prompting on \existtasks/}
    \label{fig:few_shot_prompting:existing_tasks}
\end{figure*}

We report results of prompting \nbmodel/ using few-shot exemplars on the \existtasks/ split in \cref{fig:few_shot_prompting:existing_tasks}.
Compared to the results obtained on \newtasks/ (\cref{fig:exp:prompting:passatk}), while few-shot prompting, especially step-by-step prompting, is still effective compared to the baseline, the gap is not as profound as the results on \newtasks/.
The difference between different prompting methods is also less significant, and using natural language explanations ($\mathrm{SbS}+\mathrm{Explanations}$) seems to hurt performance on simpler tasks, while still beneficial for complex problems with more than 7 {\tt pandas} API calls.
This is likely due to that the model relies on memorization to solve problems in \existtasks/, rendering using extra exemplars to control generation less effective.
Nevertheless, these results reiterate the value of \newtasks/ as a more reliable benchmark to better differentiate different prompting strategies. 

\section{Further Analysis of Solution Diversity}
\label{sec:app:solution_diversity}

\begin{figure}[t]
    \centering
    \subfloat{%
        \includegraphics[width=0.5 \columnwidth]{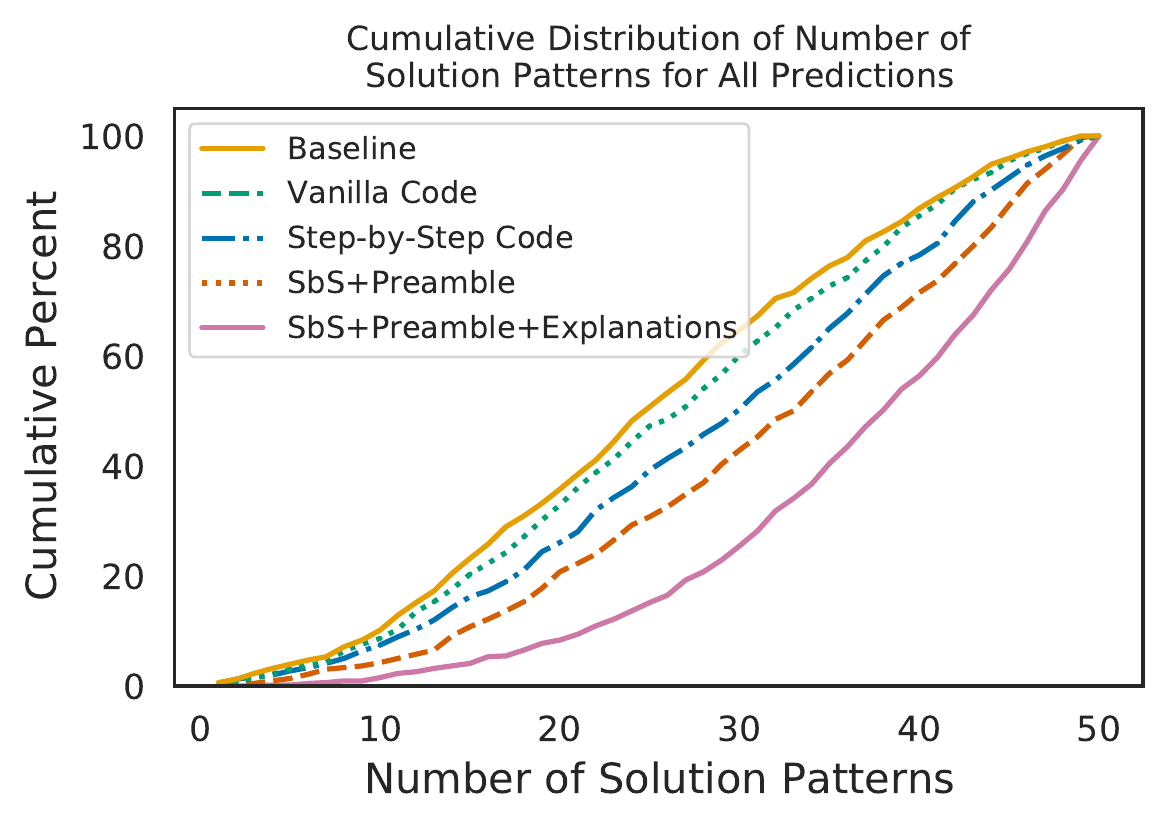}%
    }
    \subfloat{%
        \includegraphics[width=0.5 \columnwidth]{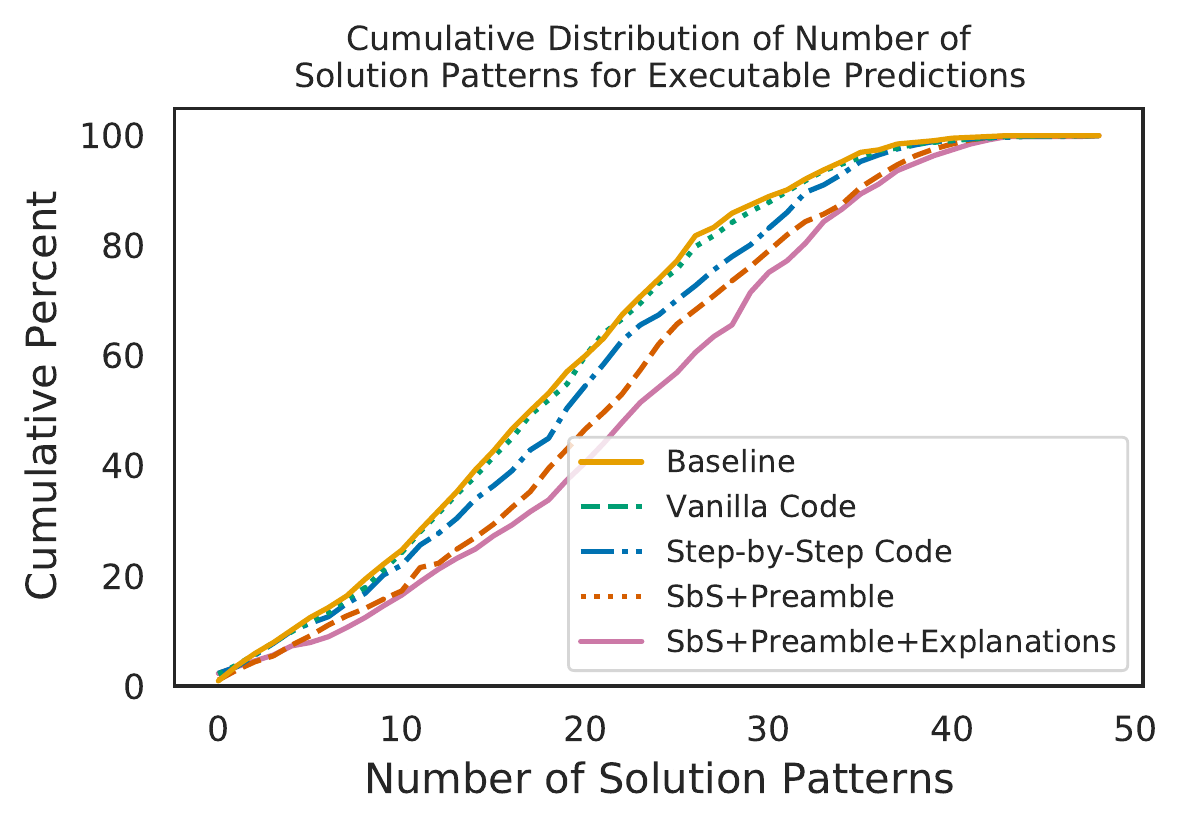}%
    }\\
    \subfloat{%
        \includegraphics[width=0.5 \columnwidth]{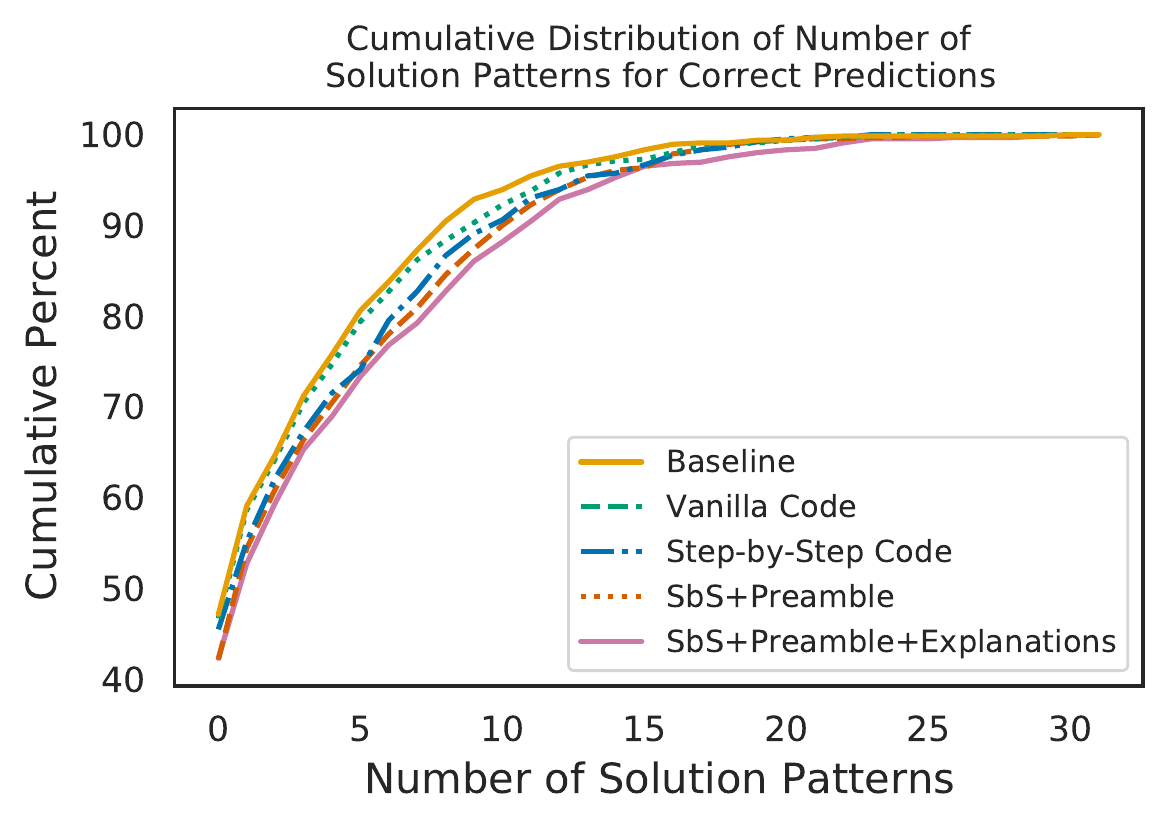}%
    }
    \caption{Cumulative distributions of the number of unique API sequences extracted from \nbmodel/'s $50$ predictions on the \textit{New Tasks} split, for all predictions (top left), only executable predictions (top right), or only correct predictions (bottom). Curves that appear more to the right represent prompting methods with greater diversity in the samples.}
    \label{fig:app:solution_diversity}
\end{figure}

Here we provide further analysis of the diversity in solution patterns, measured by the number of distinct {\tt pandas} API call sequences used in the samples. \cref{fig:app:solution_diversity} shows a cumulative distribution of the number of solution patterns for different subsets of the predictions: all predictions, only those that execute successfully, and only the correct predictions. In each case, we see that step-by-step prompting leads to increased diversity compared to the baselines, and prompting with NL explanations further increases the diversity. While increased diversity is helpful in finding a correct solution at higher sample size $k$, it is also helpful when considering only correct solutions because a user might want to see a variety of solution approaches, whether for educational purposes or to choose the one they like best (which is partially subjective). Refer to \cref{sec:prompting_case_study} for such examples.

\section{Error Analysis Examples}
\label{app:error_examples}
\begin{figure*}[t]
    \centering
    \small
    \begin{minipage}[c]{0.7 \textwidth}
    \vspace{-2mm}
    $\intent$: \textit{What are the average shipping fees for each state, starting from highest to lowest? (rounded to 2 decimal places)?} \\
 \fbox{\includegraphics[width=\textwidth, trim = 1.6cm 9.25cm 2.35cm 0.2cm, clip]{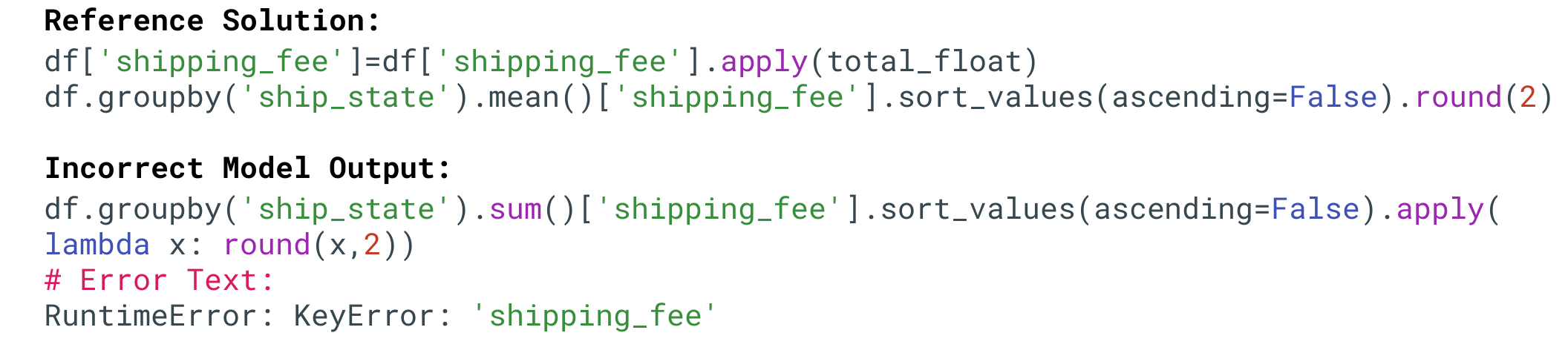}}
    \end{minipage}
    \vspace{-1.5mm}
    \caption{An example of {\tt KeyError}: the model calls a column which is removed after the {\tt pd.groupby().mean()} API call.}
    \label{fig:errorexample:keyerror}
\end{figure*}

\begin{figure*}[t]
    \centering
    \small
    \begin{minipage}[c]{0.7 \textwidth}
    \vspace{-2mm}
    $\intent$: \textit{Show how many new users signed up for every year since 2000} \\
 \fbox{\includegraphics[width=\textwidth, trim = 1.95cm 8.75cm 2.35cm 0.2cm, clip]{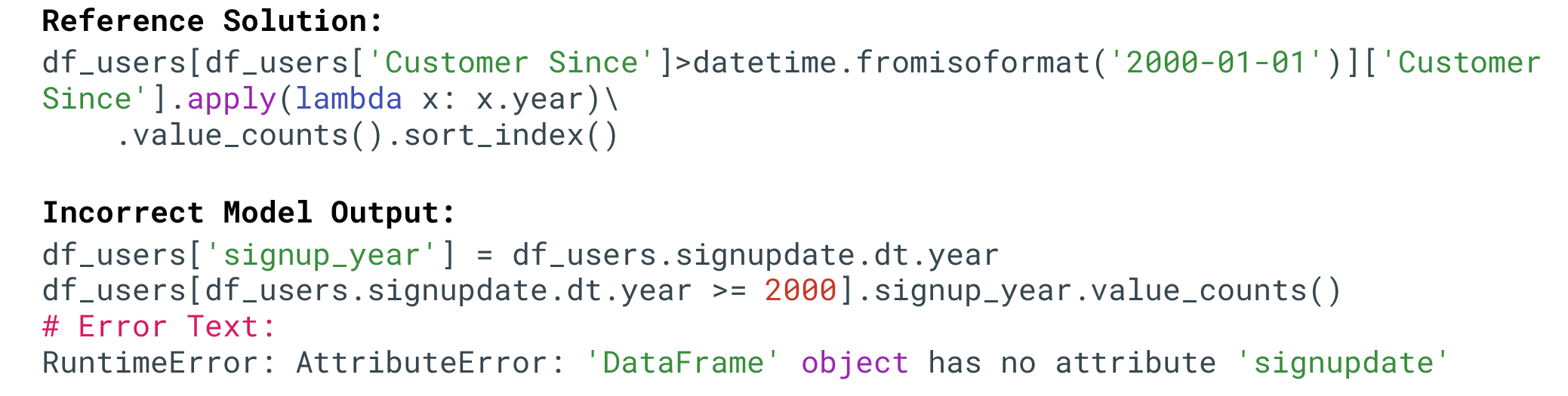}}
    \end{minipage}
    \vspace{-1.5mm}
    \caption{An example of {\tt AttributeError}: the model tries to call a non-existing column.}
    \label{fig:errorexample:attributeerror}
\end{figure*}

\begin{figure*}[t]
    \centering
    \small
    \begin{minipage}[c]{0.7 \textwidth}
    \vspace{-2mm}
    $\intent$: \textit{What are the top five models with most number of bikes having mileage less than 5000 kilometers?} \\
 \fbox{\includegraphics[width=\textwidth, trim = 2.5cm 9.65cm 1.75cm 0.25cm, clip]{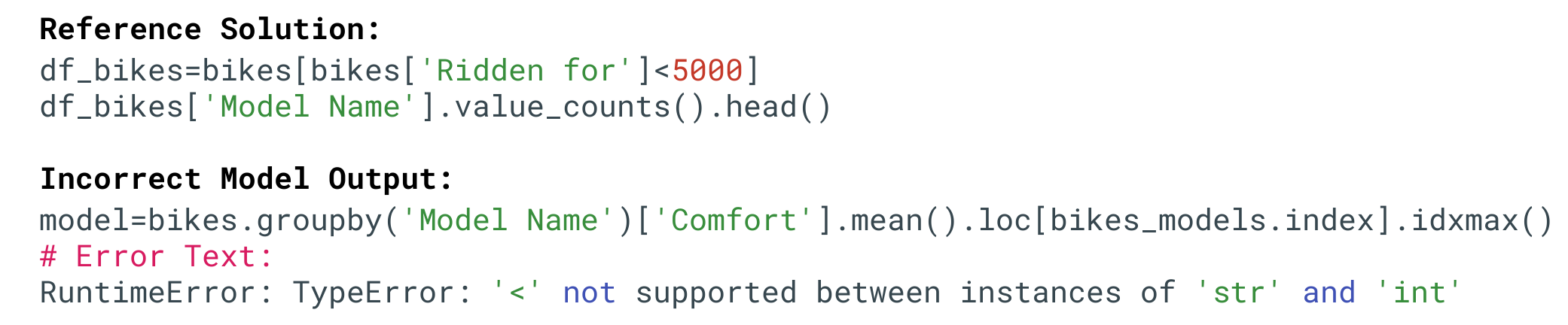}}
    \end{minipage}
    \vspace{-1.5mm}
    \caption{An example of {\tt TypeError}: the model is intent to compare a string-value column to an integer.}
    \label{fig:errorexample:typeerror}
\end{figure*}

\begin{figure*}[t]
    \centering
    \small
    \begin{minipage}[c]{0.7 \textwidth}
    \vspace{-2mm}
    $\intent$: \textit{What is the average number of filed charges for drug related cases?} \\
 \fbox{\includegraphics[width=\textwidth, trim = 1.35cm 9.7cm 1.75cm 0.2cm, clip]{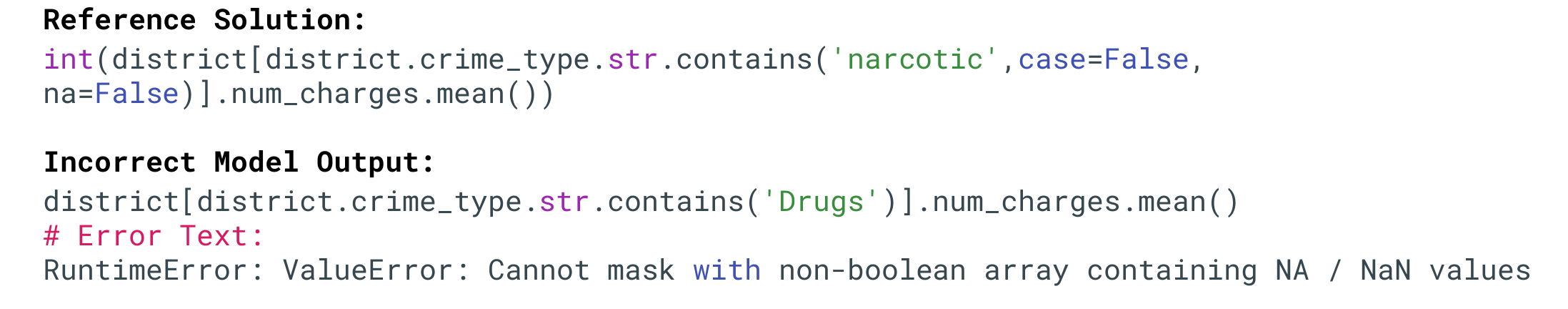}}
    \end{minipage}
    \vspace{-1.5mm}
    \caption{An example program of {\tt ValueError}: the model tries to calculate the mean of a column containing {\tt NaN} values.}
    \label{fig:errorexample:valuerror}
\end{figure*}

In this section, we provided a preliminary study on selected example model errors. We analyzed two types of errors: execution error and semantic error (executable but incorrect). Execution error has error message from the notebook environment. We can classify these errors into more fine-grained categories in \cref{fig:exp:runtime_error_distribution}. As the result shows, {\tt KeyError} is the top error mode in the execution errors. Over 50\% of the {\tt KeyError} are associated with the {\tt pd.groupby} API call. {\tt pd.groupby} API call changes the dataframe schema as the model generates the code. For example, {\tt pd.groupby(*).mean()} will remove non-numeric columns in the dataframe. This requires the model to have a profound understanding of the dataframe schema. We gave an example in \cref{fig:errorexample:keyerror}. The column {\tt shipping\_fee} is string value which will be removed after {\tt df.groupby(ship\_state).sum()}. The secondary source of execution error is {\tt AttributeError}, which shares a similar cause to the {\tt KeyError}. This is because {\tt AttributeError} is often triggered by calling a non-existing column as an attribute of a {\tt pd.dataframe}. An example is given in \cref{fig:errorexample:attributeerror}, where the model tries to call the non-existing column {\tt signupdate} as an attribute of {\tt df\_users}, leading to the {\tt AttributeError}. These two error modes suggest that building a better schema aware language model is a promising future research direction.  We also present \cref{fig:errorexample:typeerror} and \cref{fig:errorexample:valuerror} as the example for {\tt TypeError} and {\tt ValueError} respectively. These two error modes are often caused by insufficient knowledge of the column values. For example, the model tried to compare a string-value column to a integer in \cref{fig:errorexample:typeerror}, which causes {\tt TypeError}. \cref{fig:errorexample:valuerror} showcased that the model tries to apply numeric operation {\tt pd.DataFrame.mean()} on a column with {\tt NaN} values, leading to {\tt ValueError}.

\begin{figure*}[t]
    \centering
    \small
    \begin{minipage}[c]{0.7 \textwidth}
    \vspace{-2mm}
    $\intent$: \textit{What is the number of deaths by accident as a percentage of total deaths in the last ten years?} \\
 \fbox{\includegraphics[width=\textwidth, trim = 1.6cm 5.45cm 2.35cm 1cm, clip]{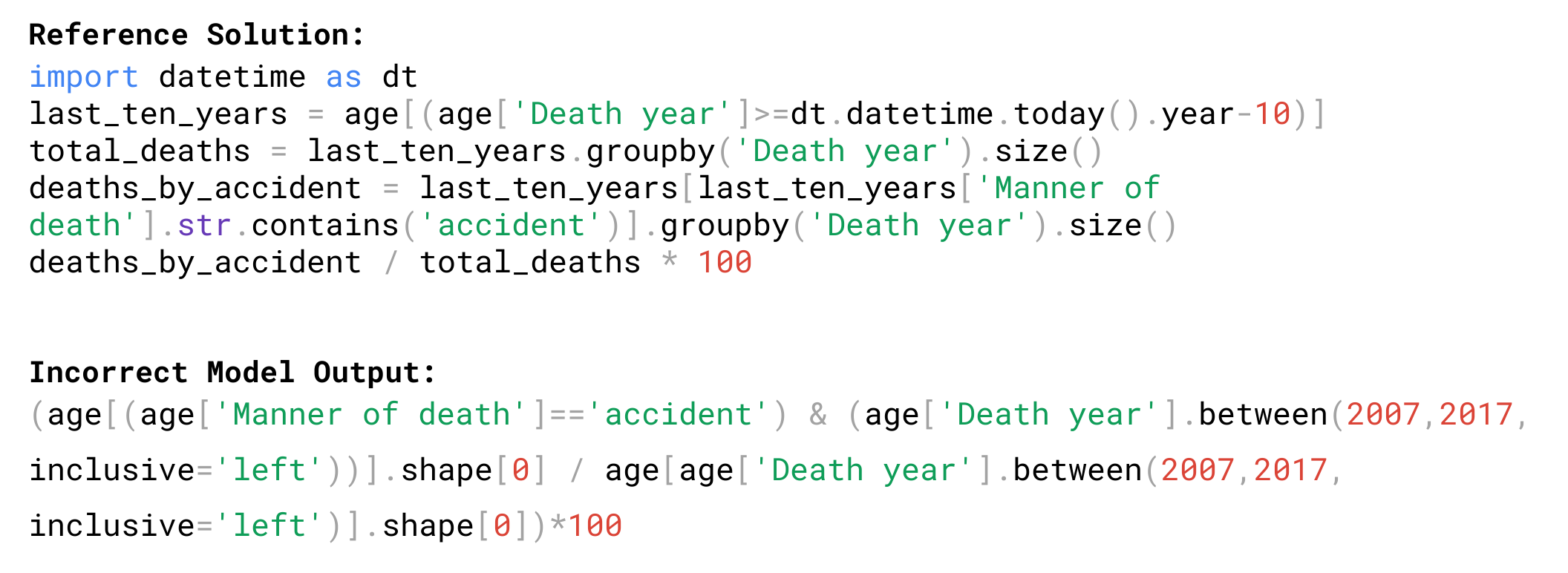}}
    \end{minipage}
    \vspace{-1.5mm}
    \caption{An example of complex reasoning: the model has to infer the API call ({\tt dt.datetime.today().year - 10}) from the last-ten-years constraint in the intent.}
    \label{fig:errorexample:agediff}
\end{figure*}

\begin{figure*}[t]
    \centering
    \small
    \begin{minipage}[c]{0.7 \textwidth}
    \vspace{-2mm}
    $\intent$: \textit{Which hotels had a worse ranking this year than in 2021? Show the hotel name, location and the difference in ranking from last year.} \\
 \fbox{\includegraphics[width=\textwidth, trim = 2.5cm 9.75cm 2.75cm 0.1cm, clip]{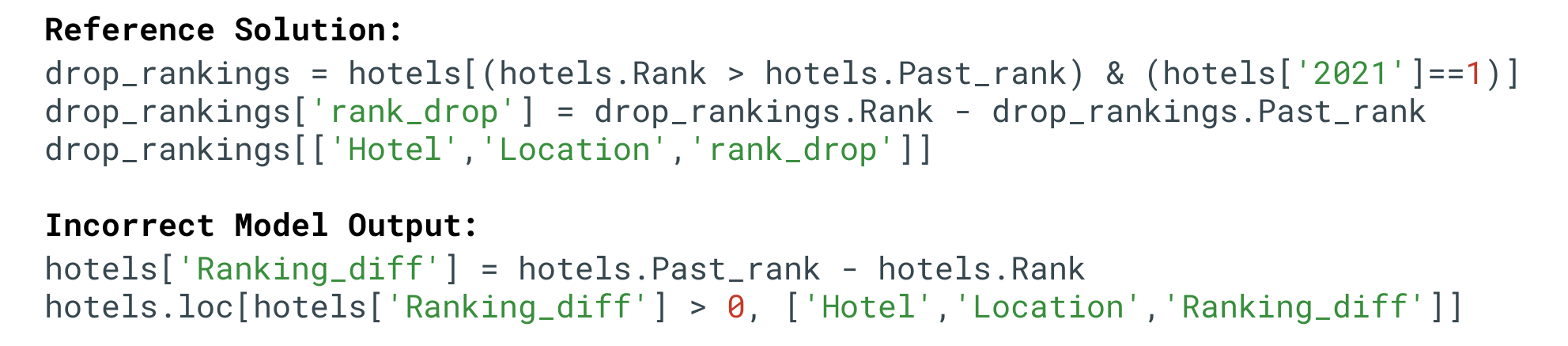}}
    \end{minipage}
    \vspace{-1.5mm}
    \caption{An example of complex reasoning: the model needs to compare the current rank and the past rank while making sure the rank in 2021 exists. }
    \label{fig:errorexample:hoteldiff}
\end{figure*}

\begin{figure*}[t]
    \centering
    \small
    \begin{minipage}[c]{0.72 \textwidth}
    \vspace{-2mm}
    $\intent$: \textit{What was the mean opening rank across all IIT institutes for each program over the years? Show the mean opening rank for each year in columns with program as index and consider only general students.} \\
 \fbox{\includegraphics[width=\textwidth, trim = 0.cm 9.7cm 0.3cm 0.15cm, clip]{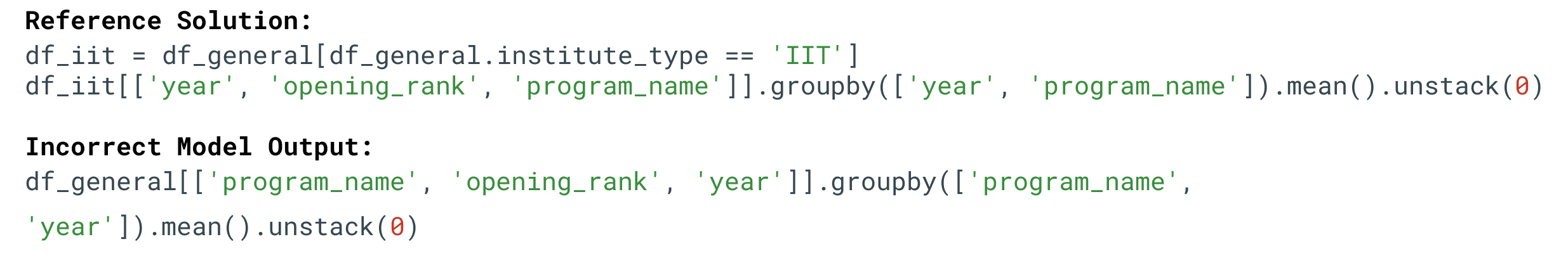}}
    \end{minipage}
    \vspace{-1.5mm}
    \caption{An example of NL misunderstanding: the model does not filter the institute type according to the intent.}
    \label{fig:errorexample:nlmisunderstand}
\end{figure*}

\begin{figure*}[t]
    \centering
    \small
    \begin{minipage}[c]{0.7 \textwidth}
    \vspace{-2mm}
    $\intent$: \textit{How many front and left facing trees were planted in that park?} \\
 \fbox{\includegraphics[width=\textwidth, trim = 1cm 9.85cm 1.25cm 0.1cm, clip]{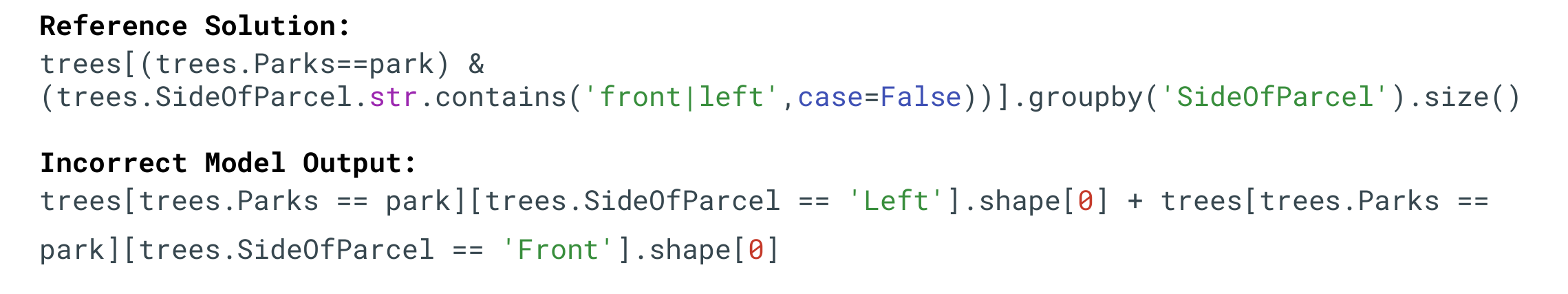}}
    \end{minipage}
    \vspace{-1.5mm}
    \caption{An example of under-specified intent. It does not specify the output should not sum the number of front facing and left facing trees.}
    \label{fig:errorexample:underspecified}
\end{figure*}

\begin{figure}[t]
    \centering
    \subfloat[Intent, reference program and generated program]{
    \begin{minipage}[c]{0.7 \textwidth}
    \small
    \vspace{-2mm}
    $\intent$: \textit{Return a matrix with the average ticket prices to and from all the cities for each ticket class.} \\
     \fbox{\includegraphics[width=\textwidth, trim = 1.25cm 10.6cm 1.55cm 0.1cm, clip]{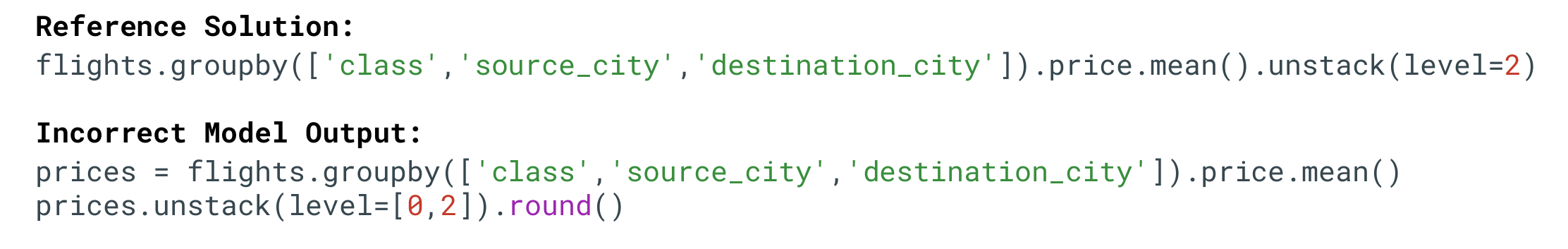}}
    \end{minipage}
    }\\
    \subfloat[Reference output]{%
\resizebox{0.8 \textwidth}{!}{\begin{tabular}{|l|l|l|l|l|l|l|l|}
\hline
& destination\_city & Bangalore    & Chennai      & Delhi        & Hyderabad    & Kolkata      & Mumbai            \\ \hline
class             & source\_city &              &              &              &              &              &              \\ \hline
Business          & Bangalore    & NaN          & 52436.915395 & 48144.337108 & 50395.796948 & 58854.693091 & 58024.618208 \\ \hline
                  & Chennai      & 53113.008692 & NaN          & 52443.367242 & 51559.874283 & 57078.895872 & 56223.838086 \\ \hline
                  & Delhi        & 48576.027921 & 52031.778099 & NaN          & 44457.376775 & 56239.853659 & 44364.442811 \\ \hline
                  & Hyderabad    & 50358.290706 & 51132.155288 & 44250.700281 & NaN          & 53729.157762 & 52184.424666 \\ \hline
                  & Kolkata      & 58681.104437 & 56502.775035 & 55047.492193 & 54732.447908 & NaN          & 57422.551724 \\ \hline
                  & Mumbai       & 57970.544389 & 55703.326197 & 43846.329273 & 51593.643678 & 57106.526385 & NaN          \\ \hline
Economy           & Bangalore    & NaN          & 7105.953850  & 6124.897982  & 6360.141698  & 7375.638594  & 6381.093332  \\ \hline
                  & Chennai      & 7175.020192  & NaN          & 6075.961190  & 5960.788831  & 7547.295815  & 6529.119453  \\ \hline
                  & Delhi        & 6175.622535  & 6102.317245  & NaN          & 6031.164261  & 7045.621678  & 6059.826087  \\ \hline
                  & Hyderabad    & 6234.882649  & 6049.884930  & 6072.296659  & NaN          & 6881.680392  & 5969.259906  \\ \hline
                  & Kolkata      & 7471.621990  & 8011.745229  & 7161.400077  & 7489.144374  & NaN          & 7405.787239  \\ \hline
                  & Mumbai       & 6432.511946  & 6420.917984  & 5889.281400  & 5774.891130  & 7227.971735  & NaN          \\ \hline
\end{tabular}}
    }\\
    \subfloat[Model output]{%
        \resizebox{0.8 \textwidth}{!}{\begin{tabular}{|l|llllll|llllll|}
\hline
class             & \multicolumn{6}{l|}{Business}                                                                                                                                          & \multicolumn{6}{l|}{Economy}                                                                                                                                         \\ \hline
destination\_city & \multicolumn{1}{l|}{Chennai} & \multicolumn{1}{l|}{Delhi}   & \multicolumn{1}{l|}{Hyderabad} & \multicolumn{1}{l|}{Kolkata} & \multicolumn{1}{l|}{Mumbai}  & Bangalore & \multicolumn{1}{l|}{Chennai} & \multicolumn{1}{l|}{Delhi}  & \multicolumn{1}{l|}{Hyderabad} & \multicolumn{1}{l|}{Kolkata} & \multicolumn{1}{l|}{Mumbai} & Bangalore \\ \hline
source\_city      & \multicolumn{1}{l|}{}        & \multicolumn{1}{l|}{}        & \multicolumn{1}{l|}{}          & \multicolumn{1}{l|}{}        & \multicolumn{1}{l|}{}        &           & \multicolumn{1}{l|}{}        & \multicolumn{1}{l|}{}       & \multicolumn{1}{l|}{}          & \multicolumn{1}{l|}{}        & \multicolumn{1}{l|}{}       &           \\ \hline
Bangalore         & \multicolumn{1}{l|}{52437.0} & \multicolumn{1}{l|}{48144.0} & \multicolumn{1}{l|}{50396.0}   & \multicolumn{1}{l|}{58855.0} & \multicolumn{1}{l|}{58025.0} & NaN       & \multicolumn{1}{l|}{7106.0}  & \multicolumn{1}{l|}{6125.0} & \multicolumn{1}{l|}{6360.0}    & \multicolumn{1}{l|}{7376.0}  & \multicolumn{1}{l|}{6381.0} & NaN       \\ \hline
Chennai           & \multicolumn{1}{l|}{NaN}     & \multicolumn{1}{l|}{52443.0} & \multicolumn{1}{l|}{51560.0}   & \multicolumn{1}{l|}{57079.0} & \multicolumn{1}{l|}{56224.0} & 53113.0   & \multicolumn{1}{l|}{NaN}     & \multicolumn{1}{l|}{6076.0} & \multicolumn{1}{l|}{5961.0}    & \multicolumn{1}{l|}{7547.0}  & \multicolumn{1}{l|}{6529.0} & 7175.0    \\ \hline
Delhi             & \multicolumn{1}{l|}{52032.0} & \multicolumn{1}{l|}{NaN}     & \multicolumn{1}{l|}{44457.0}   & \multicolumn{1}{l|}{56240.0} & \multicolumn{1}{l|}{44364.0} & 48576.0   & \multicolumn{1}{l|}{6102.0}  & \multicolumn{1}{l|}{NaN}    & \multicolumn{1}{l|}{6031.0}    & \multicolumn{1}{l|}{7046.0}  & \multicolumn{1}{l|}{6060.0} & 6176.0    \\ \hline
Hyderabad         & \multicolumn{1}{l|}{51132.0} & \multicolumn{1}{l|}{44251.0} & \multicolumn{1}{l|}{NaN}       & \multicolumn{1}{l|}{53729.0} & \multicolumn{1}{l|}{52184.0} & 50358.0   & \multicolumn{1}{l|}{6050.0}  & \multicolumn{1}{l|}{6072.0} & \multicolumn{1}{l|}{NaN}       & \multicolumn{1}{l|}{6882.0}  & \multicolumn{1}{l|}{5969.0} & 6235.0    \\ \hline
Kolkata           & \multicolumn{1}{l|}{56503.0} & \multicolumn{1}{l|}{55047.0} & \multicolumn{1}{l|}{54732.0}   & \multicolumn{1}{l|}{NaN}     & \multicolumn{1}{l|}{57423.0} & 58681.0   & \multicolumn{1}{l|}{8012.0}  & \multicolumn{1}{l|}{7161.0} & \multicolumn{1}{l|}{7489.0}    & \multicolumn{1}{l|}{NaN}     & \multicolumn{1}{l|}{7406.0} & 7472.0    \\ \hline
Mumbai            & \multicolumn{1}{l|}{55703.0} & \multicolumn{1}{l|}{43846.0} & \multicolumn{1}{l|}{51594.0}   & \multicolumn{1}{l|}{57107.0} & \multicolumn{1}{l|}{NaN}     & 57971.0   & \multicolumn{1}{l|}{6421.0}  & \multicolumn{1}{l|}{5889.0} & \multicolumn{1}{l|}{5775.0}    & \multicolumn{1}{l|}{7228.0}  & \multicolumn{1}{l|}{NaN}    & 6433.0    \\ \hline
\end{tabular}}
    }
    \caption{An example of plausible alternative prediction that is labeled as incorrect due to limited coverage of the evaluation metric.}
    \label{fig:errorexample:evalcoverage}
\end{figure}

Semantic errors account for more than 65\% of the incorrect predictions. We cannot get any error messages from these executable but incorrect programs. Therefore, we conducted a manual analysis on 50\% randomly sampled incorrect predictions. We classified the semantic errors into 5 categories: 1. Complex task requiring non-trivial computation ($43\%$); 2. NL misunderstanding ($26\%$); 3. Under-specified intents ($19\%$); 4. False-negatives of the fuzzy-matching evaluation metric ($6\%$); 5. Annotation Errors ($6\%$). To complement the discussion in \cref{sec:error_analysis}, we showcase examples of these types of errors. The primary source of semantic error is \textbf{complex reasoning}. Examples are given in \cref{fig:errorexample:agediff} and \cref{fig:errorexample:hoteldiff}. In \cref{fig:errorexample:agediff}, the model need to infer that last 10 year can be computed by {\tt dt.datetime.today().year - 10} then apply string operation {\tt str.contains(accident)} to select the required contents. \cref{fig:errorexample:hoteldiff} is another example of complex data wrangling steps. To generate the correct program, the model need to compare the current rank to the past rank, ensure the 2021 rank column exists and then aggregate the information. \textbf{NL misunderstanding} is the secondary source of semantic errors. In \cref{fig:errorexample:nlmisunderstand}, the generated output does not reflect the institute constraint in the intent. Another source of semantic errors is \textbf{under-specified intent}, which happens occasionally in natural language instructions. In \cref{fig:errorexample:underspecified}, the intent does not specify the output format. The program generated by the model calculates the sum of front and left facing trees while the reference calculates them respectively. The evaluation fails but the model output is not necessarily incorrect in this case. \cref{fig:errorexample:evalcoverage} illustrates another type of semantic error: \textbf{evaluation coverage}. In this example, the model gives an acceptable answer which only differs from the reference in the column order. It does not pass the evaluation due to its limited coverage.

\begin{figure*}[t]
    \centering
    \small
    \begin{minipage}[c]{0.75 \textwidth}
 $\intent$: \textit{What are the five most common genres for shows released during the year having the \\highest percent increase in votes?} \\
 \fbox{\includegraphics[width=\linewidth]{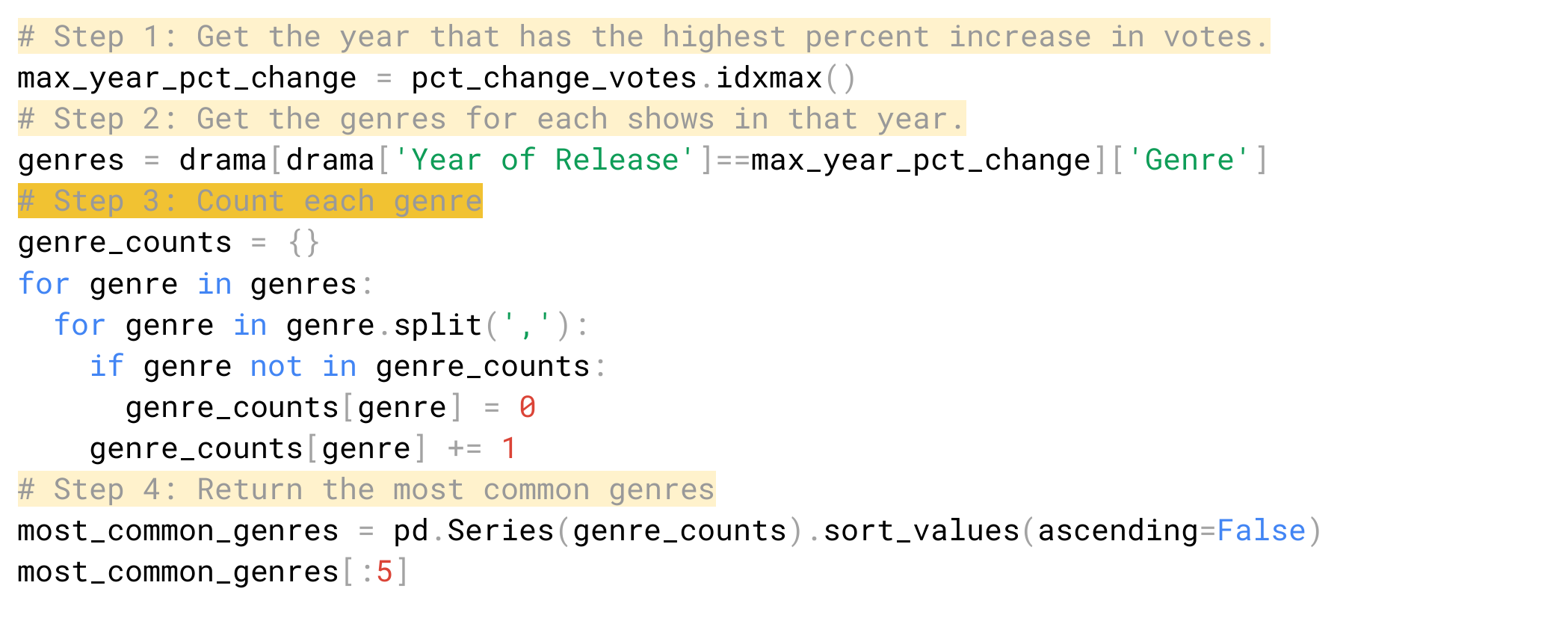}}
    \end{minipage}
    \caption{An example prediction from \nbmodel/ with a complex single step.}
    \label{fig:explanation_study:complex_single_step}
\end{figure*}

\begin{figure*}[t]
    \centering
    \small
    \begin{minipage}[c]{0.7 \textwidth}
    $\intent$: \textit{Convert crash dates to datetime and show the total number of vehicles involved in crashes over the years.} \\
 \fbox{\includegraphics[width=\linewidth]{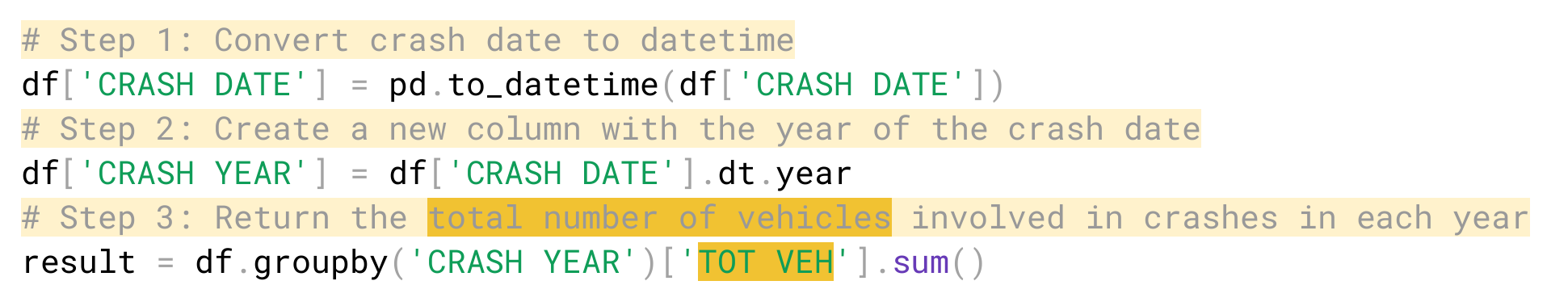}}
    \end{minipage}
    \caption{An example prediction from \nbmodel/ that explains the semantics of a column.}
    \label{fig:explanation_study:meaningfulname_to_schema}
\end{figure*}

\begin{figure*}[t]
    \centering
    \small
    \begin{minipage}[c]{0.7 \textwidth}
    $\intent$: \textit{What is the distribution of student adaptivity level across each age group? \\Return a {\tt DataFrame} with age groups as an index and adaptivity levels as columns.} \\
\fbox{\includegraphics[width=\linewidth]{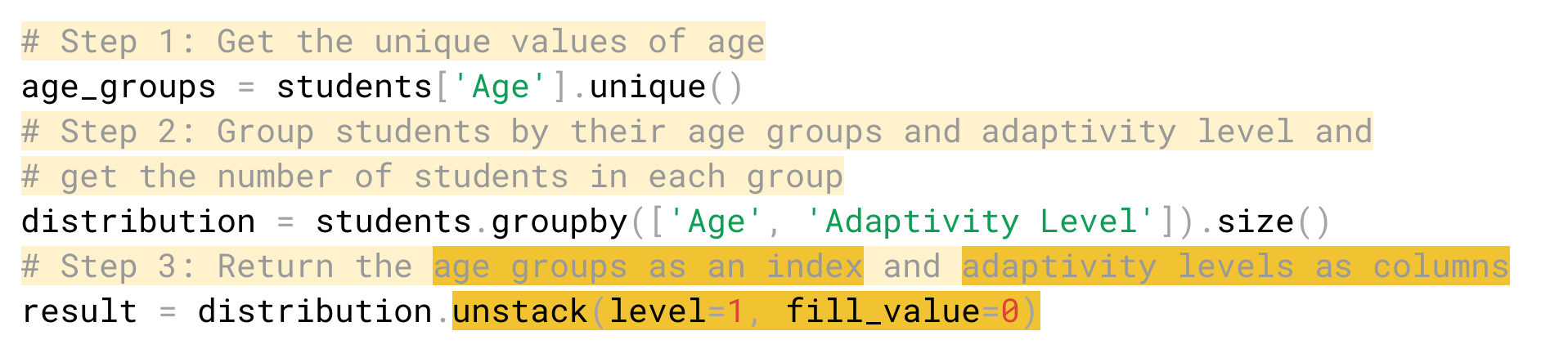}}
    \end{minipage}
    \caption{An example prediction from \nbmodel/ that explains the return value from a complex API.}
    \label{fig:explanation_study:explain_complex_api}
\end{figure*}

\begin{figure*}[t]
    \centering
    \small
    \begin{minipage}[c]{0.7 \textwidth}
    $\intent$: \textit{Drop columns with more than 70 percent null values} \\
\fbox{\includegraphics[width=\linewidth]{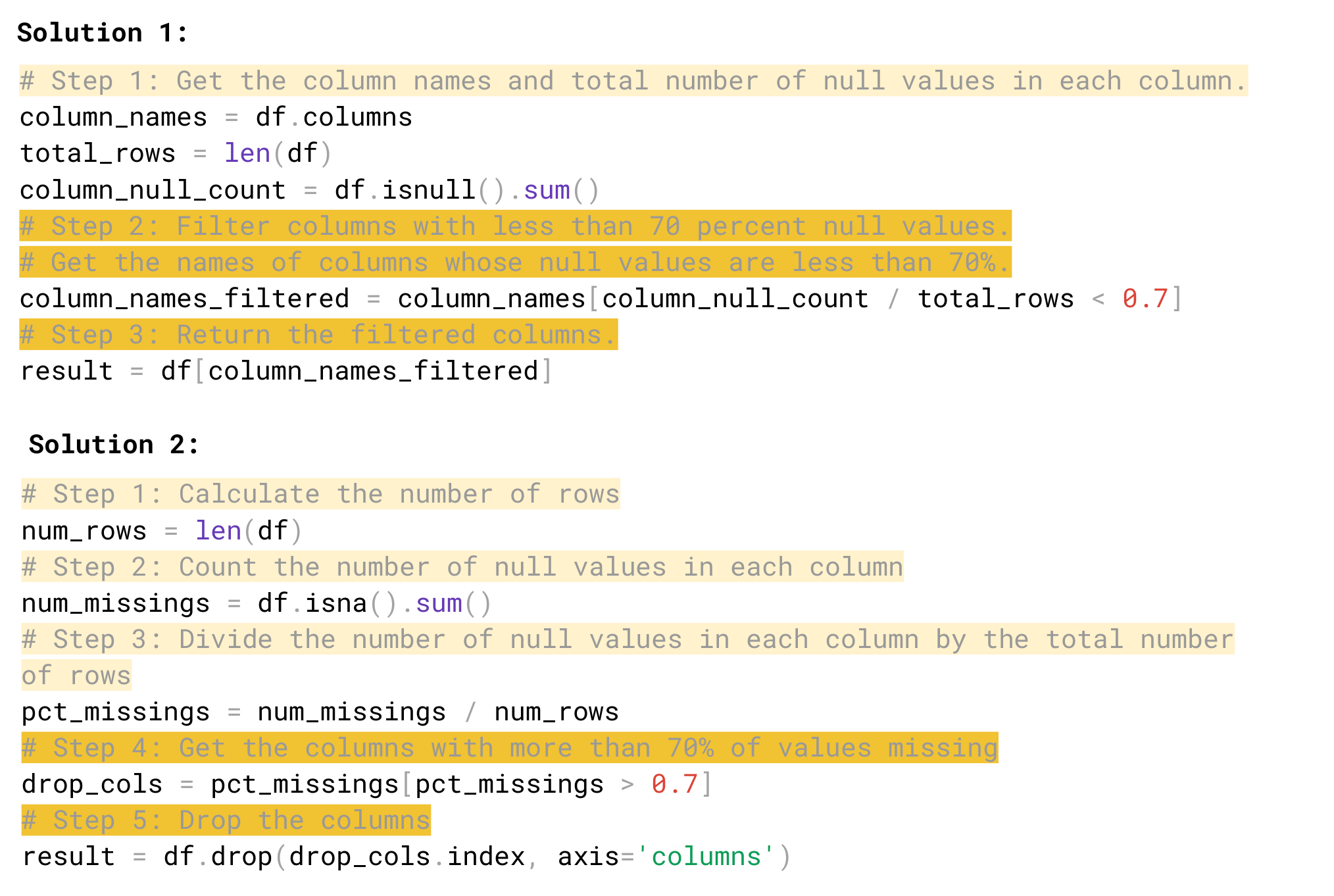}}
    \end{minipage}
    \caption{Two predictions from \nbmodel/ with different solution approaches.}
    \label{fig:explanation_study:multi_sol}
\end{figure*}

\begin{figure*}[t]
    \centering
    \small
    \begin{minipage}[c]{0.7 \textwidth}
    $\intent$: \textit{In which year, within the last ten years, did Israel receive the highest amount of financial aid, in constant amount? Show the year and amount received.} \\
 \fbox{\includegraphics[width=\linewidth]{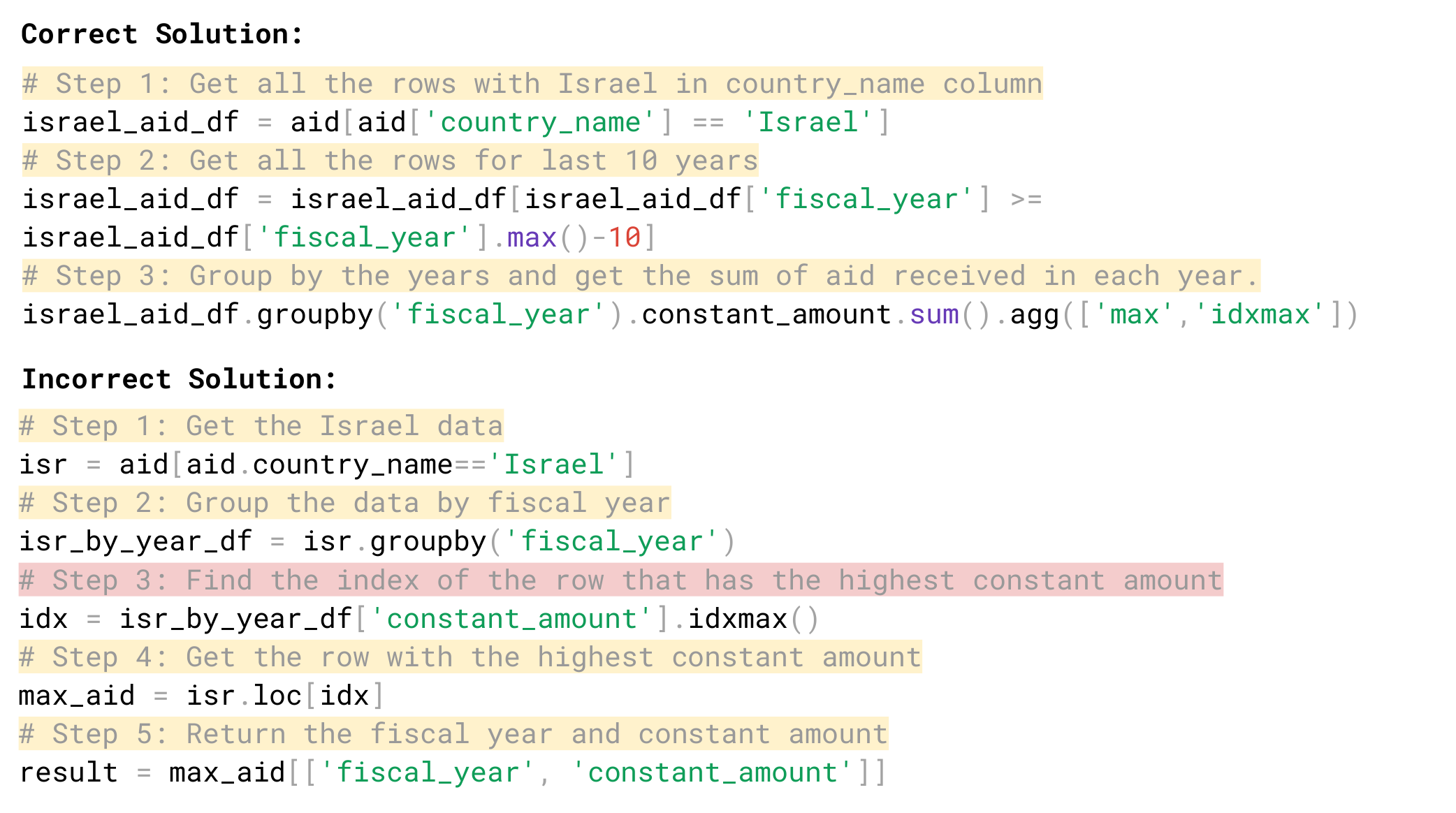}}
    \end{minipage}
    \caption{Two predictions from \nbmodel/. NL explanations help users identify incorrect steps.}
    \label{fig:explanation_study:identify_incorrect_sol}
\end{figure*}

\clearpage
\newpage

\section{Data Card for the Training Data of \nbmodel/}
\label{app:datacard}

We provide a data card for the training data of \nbmodel/ as outlined in \cref{sec:model}, following the structure presented in~\citet{chowdhery2022palm}.
We also report training data composition in~\cref{table:data-composition}.

\begin{longtable}[c]{ p{.32\textwidth} p{.62\textwidth} } 
\toprule \\
\multicolumn{2}{c}{\textbf{Motivation}} \\ \toprule
For what purpose was the dataset created? Who created the dataset? Who funded the creation of the dataset? & The dataset was created for training code and language models by a team of researchers at Google. \\ \midrule
\multicolumn{2}{c}{\textbf{Composition}}      \\ \toprule
What do the instances that comprise the dataset represent (e.g., documents, photos, people, countries)? & Dataset comprises of Python source code files and Jupyter notebooks from GitHub, filtered by license so as to exclude code with restrictive licenses. \\ \midrule
How many instances are there in total (of each type, if appropriate)? & The data makeup is given in Table~\ref{table:data-composition}. \\ \midrule
Does the dataset contain all possible instances or is it a sample (not necessarily random) of instances from a larger set? & The dataset is a small (random) subset of a larger set. \\ \midrule
What data does each instance consist of? & Each instance is encoded content of a source code file. \\ \midrule
Is there a label or target associated with each instance? & No, there are no labels associated with each instance. \\ \midrule
Is any information missing from individual instances? & No. \\ \midrule
Are relationships between individual instances made explicit? & No. \\ \midrule
Are there recommended data splits? & We use random splits for the training, validation, and test. \\ \midrule
Are there any errors, sources of noise, or redundancies in the dataset? & \begin{itemize}
    \item Python files were near deduplicated at the file level using a custom implementation of minhash algorithm, so lower level redundancies (lines, code blocks) may still exist.
    \item Some files were misclassified in the license tagging and filtration process given that license classification algorithm can have false positives and negatives.
\end{itemize}  \\ \midrule
Is the dataset self-contained, or does it link to or otherwise rely on external resources? & The dataset is self-contained. \\ \midrule
Does the dataset contain data that might be considered confidential? & No. \\ \midrule
Does the dataset contain data that, if viewed directly, might be offensive, insulting, threatening, or might otherwise cause anxiety? & Given the dataset contains source code, it is not likely there is any offensive text in it, however no explicit measures are in place to eliminate such data if it were present.  \\
\toprule
\multicolumn{2}{c}{\textbf{Collection Process}}      \\ \toprule
How was the data associated with each instance acquired? & The data was collected from publicly available sources. \\ \midrule
What mechanisms or procedures were used to collect the data? & The data was collected using a variety of software programs to extract and clean source code files. \\ \midrule
If the dataset is a sample from a larger set, what was the sampling strategy? & The dataset is small subset of publicly available code from Github, sampled randomly.\\ \midrule
Who was involved in the data collection process? & A team of researchers at Google. \\ \midrule
Over what timeframe was the data collected? & April - July 2022\\ \midrule
Were any ethical review processes conducted? & No. \\ \toprule
\multicolumn{2}{c}{\textbf{Preprocessing, cleaning, and labeling}}      \\ \toprule
Was any preprocessing, cleaning, or labeling of the data done (e.g., discretization or bucketing, tokenization, part-of-speech tagging, SIFT feature extraction, removal of instances, processing of missing values)? & License filtration, quality filtration and deduplication were applied to the source code files.
\begin{itemize}
    \item License classification was done using \href{https://github.com/google/licenseclassifier}{Google License Classifier} library. Source code files with restricted licenses were filtered out.
    \item Python files were deduplicated at the file level using a custom variant of minhash algorithm. Locality sensitive hashes of file content were used to create partitions of potentially duplicate files based on collisions in the hash buckets. For each pair in the partitions, Jaccard Similarity and Edit Distance scores were calculated to create an "edge" for a pair whenever the scores are higher than the specified threshold. This was followed by application of connected components algorithm to return the sets of duplicates.
    \item Jupyter notebooks were first deduplicated following the same procedure as deduplicating Python files, and then deduplicated at individual cell level against the evaluation dataset (\cref{sec:model}).
\end{itemize}\\ \midrule
Is the software used to preprocess, clean, or label the instances available? & No. \\ \toprule
\multicolumn{2}{c}{\textbf{Uses}}      \\ \toprule
Has the dataset been used for any tasks already? & Yes, we use the dataset for pre-training other code and language models. \\ \midrule
Is there a repository that links to any or all papers or systems that use the dataset? & No. \\ \midrule
What (other) tasks could the dataset be used for? & The dataset can be used for training of other code and language models. \\ \midrule
Is there anything about the composition of the dataset or the way it was collected and pre-processed/cleaned/labeled that might impact future uses? & The dataset is static in nature and thus will become progressively more “stale”. It will not include any new source code repositories that were created/updated later on Github.\\ \midrule
Are there tasks for which the dataset should not be used? & This should not be used for any unacceptable code or language modeling use cases e.g. generating code or language with toxic/biased connotations.\\ \toprule
\multicolumn{2}{c}{\textbf{Distribution}}      \\ \toprule
Will the dataset be distributed to third parties outside of the entity (e.g., company, institution, organization) on behalf of which the dataset was created? & No. \\
\bottomrule
\label{table:datasheet}
\end{longtable}

\begin{table}[H]
\centering
{
\begin{tabular}{lcc}
\toprule
\textbf{Language} & \textbf{Tokens} & \textbf{Source Files}\\
\midrule
\textrm{Python} & 63,786,481,126 & 60,397,107\\
\textrm{Jupyter Notebooks} &  9,613,648,619 & 3,796,713 \\
\bottomrule
\end{tabular}}
\caption{Data Composition of the fine-tuning data for \nbmodel/.}\label{table:data-composition}
\end{table}

\newpage
\clearpage

\section{Detailed Prompting Examples}
\label{app:prompts}

In this section we provide detailed examples of prompts used in our experiments.
As in \cref{sec:experiments:prompting_method}, there are two categories of experiments in \cref{sec:experiments}, namely prompting using notebook context (\cref{sec:exp:results}) and few-shot prompting with extra exemplars pre-pended to  notebook context (\cref{sec:exp:few_shot_prompting}).
Here, we list the prompts for $\intent_2$ in \cref{fig:intro:teaser_example} in these two types of prompting experiments.

\paragraph{Prompting using Notebook Context}
In the basic setup without extra few-shot exemplars, the prompt basically consist of all the prior notebook context, including  NL descriptions of schema information and previous rounds of problems.
\cref{lst:prompt:notebook_ctx} shows the complete prompt for $\intent_2$ in \cref{fig:intro:teaser_example} in this setup.\footnote{This prompt is not exactly the same as the one in our dataset. It is adapted to align with the illustrative example in \cref{fig:intro:teaser_example}}
At inference time, a code LM will complete the last code cell after the cell delimiter `{\tt \# In[ ]:}'.
Note that for \incoder/ we follow \citet{fried2022incoder} and use a special template to linearize notebooks (\cref{app:inference_setup}).

\paragraph{Prompting using Additional Few-shot Exemplars}
We have four prompting styles for few-shot experiments.
Here, we show the prompt prefix (\cref{sec:experiments:prompting_method}) for \vc/ and \sbs/$+\mathrm{Explanations}$ prompting, as the remaining two styles are just simplified version of the latter by removing inline explanations ($\mathrm{SbS}+\mathrm{Preamble}$) and preambles (\sbs/).

A prompt in this setup is the concatenation of a prompt prefix (with few-shot exemplars) and the notebook context (with prior rounds of problems and NL schema descriptions).
The part of a prompt that corresponds to notebook context as the same as the previous setting (\eg~\cref{lst:prompt:notebook_ctx}), except that we insert the preamble {\tt \# Solution:~Let's solve this problem step-by-step.} as appropriate after the last cell delimiter.
For prompt prefix, \cref{lst:prompt:sbs:e1} gives an example prompt prefix for \sbs/ prompting, while \cref{lst:prompt:vc:e1} shows the same set of few-shot exemplars for \vc/ prompting.

As mentioned in \cref{sec:experiments:prompting_method}, we created three prompt prefixes for each of the four different styles, and report results averaged over these three restarts.
\cref{lst:prompt:sbs:e1,lst:prompt:sbs:e2,lst:prompt:sbs:e3} show the three groups of prompt prefixes for \sbs/, and \cref{lst:prompt:vc:e1,lst:prompt:vc:e2,lst:prompt:vc:e3} show those for \vc/ prompting.
Each prompt prefix has four exemplars, and some exemplars are shared across different prefixes.
Note that some prompt prefixes in \sbs/ also contain one simple problem that does not require decomposition and explanation (\eg~Exercise 3, \cref{lst:prompt:sbs:e1}).
We find this to be useful to not bias a model from generate overly complex code solutions for simpler problems.
We did not put much effort in prompting engineering. Actually, those prompt prefixes were created before we collected $70\%$ of the dataset,

\begin{code}
\caption{\sbs/ Prompt Prefix (Group 1)}
\label{lst:prompt:sbs:e1}
\begin{minted}[linenos, breaklines, firstnumber=last]{python}

# In[ ]:


import pandas as pd
import matplotlib.pyplot as plt


# In[ ]:


# You are a professional data scientist. Answer the following questions using pandas and matplotlib.


# In[ ]:


# # Exercise 1


# In[ ]:


df = pd.read_csv('employee.csv')


# In[ ]:


# Schema of Dataframes:
# Columns in df with example values:
# name (Peter), gender (m), DOB (1992/01/17)


# In[ ]:


# Problem: How many male and female employees are born in 1992?


# In[ ]:


# Solution: Let's solve this problem step-by-step.
# Step 1: convert date of birth in to datetime
df['DOB'] = pd.to_datetime(df['DOB'])
# Step 2: get the number of male born in 1992
num_male_students = len(df[(df['DOB'].dt.year == 1992) & (df['gender'] == 'm')])
# Step 3: get the number of female born in that year
num_female_students = len(df[(df['DOB'].dt.year == 1992) & (df['gender'] == 'f')])


# In[ ]:


# # Exercise 2


# In[ ]:


df = pd.read_csv('scores.csv')


# In[ ]:


# Schema of Dataframes:
# Columns in df with example values:
# Stu_Name (Mike), Engineering (90), English (89), Math (92)


# In[ ]:


# Problem: Get the students with an averaged score above 90 for science subjects.


# In[ ]:


# Solution: Let's solve this problem step-by-step.
# Step 1: Create a new column with the average score of engineering and math
df['Science_Avg'] = (df['Engineering'] + df['Math']) / 2
# Step 2: Get the rows whose average score is above 90
df_score_above_90 = df[df['Science_Avg'] > 90]
# Step 3: Return the student name and average scores
result = df_score_above_90[['Stu_Name', 'Science_Avg']]


# In[ ]:


# # Exercise 3


# In[ ]:


df = pd.read_csv('geo.csv')


# In[ ]:


# Schema of Dataframes:
# Columns in df with example values:
# state (WA), capital (Seattle), population (1.4 millon)


# In[ ]:


# Problem: What is the population of California?


# In[ ]:


# Solution: Let's solve this problem step-by-step.
result = df[df['state'] == 'CA']['population']


# In[ ]:


# # Exercise 4


# In[ ]:


df = pd.read_csv('phones.csv')


# In[ ]:


# Schema of Dataframes:
# Columns in df with example values:
# model (Pixel 6), brand (Google), price (387), release (2022)


# In[ ]:


# Problem: What is the most expensive phone in each brand.


# In[ ]:


# Solution: Let's solve this problem step-by-step.
# Step 1: Group models by their brands.
model_by_brand_df = df.groupby('brand')
# Step 2: Find the index of rows that have the highest price in each group
idx = model_by_brand_df['price'].idxmax()
# Step 3: Get the rows using the index
expensive_models_df = df.loc[idx]
# Step 4: Return the brand name, model and price.
result = expensive_models_df[['brand', 'model', 'price']]


# In[ ]:


# # Exercise 5

\end{minted}
\end{code}

\begin{code}
\caption{\sbs/ Prompt Prefix (Group 2)}
\label{lst:prompt:sbs:e2}
\begin{minted}[linenos, breaklines, firstnumber=last]{python}
# In[ ]:


import pandas as pd
import matplotlib.pyplot as plt


# In[ ]:


# You are a professional data scientist. Answer the following questions using pandas and matplotlib.


# In[ ]:


# # Exercise 1


# In[ ]:


df = pd.read_csv('employee.csv')


# In[ ]:


# Schema of Dataframes:
# Columns in df with example values:
# name (Peter), gender (m), DOB (1992/01/17)


# In[ ]:


# Problem: How many male and female employees are born in 1992?


# In[ ]:


# Solution: Let's solve this problem step-by-step.
# Step 1: convert date of birth in to datetime
df['DOB'] = pd.to_datetime(df['DOB'])
# Step 2: get the number of male born in 1992
num_male_students = len(df[(df['DOB'].dt.year == 1992) & (df['gender'] == 'm')])
# Step 3: get the number of female born in that year
num_female_students = len(df[(df['DOB'].dt.year == 1992) & (df['gender'] == 'f')])


# In[ ]:


# # Exercise 2


# In[ ]:


df = pd.read_csv('scores.csv')


# In[ ]:


# Schema of Dataframes:
# Columns in df with example values:
# Stu_Name (Mike), Engineering (90), English (89), Math (92)


# In[ ]:


# Problem: Get the students with an averaged score above 90 for science subjects.


# In[ ]:


# Solution: Let's solve this problem step-by-step.
# Step 1: Create a new column with the average score of engineering and math
df['Science_Avg'] = (df['Engineering'] + df['Math']) / 2
# Step 2: Get the rows whose average score is above 90
df_score_above_90 = df[df['Science_Avg'] > 90]
# Step 3: Return the student name and average scores
result = df_score_above_90[['Stu_Name', 'Science_Avg']]


# In[ ]:


# # Exercise 3


# In[ ]:


df = pd.read_csv('geo.csv')


# In[ ]:


# Schema of Dataframes:
# Columns in df with example values:
# state (WA), capital (Seattle), population (1.4 millon)


# In[ ]:


# Problem: What is the population of California?


# In[ ]:


# Solution: Let's solve this problem step-by-step.
result = df[df['state'] == 'CA']['population']


# In[ ]:


# # Exercise 4


# In[ ]:


df = pd.read_csv('phones.csv')


# In[ ]:


# Schema of Dataframes:
# Columns in df with example values:
# model (Pixel 6), brand (Google), price (387), release (2022)


# In[ ]:


# Problem: What is the most expensive phone in each brand.


# In[ ]:


# Solution: Let's solve this problem step-by-step.
# Step 1: Group models by their brands.
model_by_brand_df = df.groupby('brand')
# Step 2: Find the index of rows that have the highest price in each group
idx = model_by_brand_df['price'].idxmax()
# Step 3: Get the rows using the index
expensive_models_df = df.loc[idx]
# Step 4: Return the brand name, model and price.
result = expensive_models_df[['brand', 'model', 'price']]


# In[ ]:


# # Exercise 5
\end{minted}
\end{code}

\begin{code}
\caption{\sbs/ Prompt Prefix (Group 3)}
\label{lst:prompt:sbs:e3}
\begin{minted}[linenos, breaklines, firstnumber=last]{python}
# In[ ]:


import pandas as pd
import matplotlib.pyplot as plt


# In[ ]:


# You are a professional data scientist. Answer the following questions using pandas and matplotlib.


# In[ ]:


# # Exercise 1


# In[ ]:


df = pd.read_csv('olympics.csv')


# In[ ]:


# Schema of Dataframes:
# Columns in df with example values:
# Year (1896), City (Athens), Country (Greece), Nations (14)


# In[ ]:


# Problem: Which countries host at least two olympic games?


# In[ ]:


# Solution: Let's solve this problem step-by-step.
# Step 1: Count the number of times each country hosted olympics
count_df = df['Country'].value_counts()
# Step 2: Find entries with more than 2 counts
filtered_df = count_df[count_df >= 2]
# Step 3: Get the country names as a list
filtered_df.index.tolist()


# In[ ]:


# # Exercise 2


# In[ ]:


df = pd.read_csv('employee.csv')


# In[ ]:


# Schema of Dataframes:
# Columns in df with example values:
# name (Peter), gender (m), DOB (1992/01/17)


# In[ ]:


# Problem: How many male and female employees are born in 1992?


# In[ ]:


# Solution: Let's solve this problem step-by-step.
# Step 1: convert date of birth in to datetime
df['DOB'] = pd.to_datetime(df['DOB'])
# Step 2: get the number of male born in 1992
num_male_students = len(df[(df['DOB'].dt.year == 1992) & (df['gender'] == 'm')])
# Step 3: get the number of female born in that year
num_female_students = len(df[(df['DOB'].dt.year == 1992) & (df['gender'] == 'f')])


# In[ ]:


# # Exercise 3


# In[ ]:


df = pd.read_csv('score.csv')


# In[ ]:


# Schema of Dataframes:
# Columns in df with example values:
# name (John), score (97)


# In[ ]:


# Problem: Make a new column "grade" for letter grades (A: 90+, B: 70-90, C: <70) and plot the number of students in each grade.


# In[ ]:


# Solution: Let's solve this problem step-by-step.
# Step 1: Define a function to convert scores to letter grades.
def get_grade(score):
  if score >= 90:
    return 'A'
  elif 70 <= score < 90:
    return 'B'
  else:
    return 'C'
# Step 2: Convert scores to letter grades.
df['grade'] = df.score.apply(get_grade)
# Step 3: Count the number of students by grade.
count_df = df['grade'].value_counts()
# Step 4: Visualize in a bar chart.
count_df.plot(kind='bar')


# In[ ]:


# # Exercise 4


# In[ ]:


df = pd.read_csv('phones.csv')


# In[ ]:


# Schema of Dataframes:
# Columns in df with example values:
# model (Pixel 6), brand (Google), price (387), release (2022)


# In[ ]:


# Problem: What is the most expensive phone in each brand.


# In[ ]:


# Solution: Let's solve this problem step-by-step.
# Step 1: Group models by their brands.
model_by_brand_df = df.groupby('brand')
# Step 2: Find the index of rows that have the highest price in each group
idx = model_by_brand_df['price'].idxmax()
# Step 3: Get the rows using the index
expensive_models_df = df.loc[idx]
# Step 4: Return the brand name, model and price.
result = expensive_models_df[['brand', 'model', 'price']]


# In[ ]:


# # Exercise 5
\end{minted}
\end{code}

\begin{code}
\caption{\vc/ Prompt Prefix (Setup 1)}
\label{lst:prompt:vc:e1}
\begin{minted}[linenos, breaklines, firstnumber=last]{python}
# In[ ]:


import pandas as pd
import matplotlib.pyplot as plt


# In[ ]:


# You are a professional data scientist. Answer the following questions using pandas and matplotlib.


# In[ ]:


# # Exercise 1


# In[ ]:


df = pd.read_csv('employee.csv')


# In[ ]:


# Schema of Dataframes:
# Columns in df with example values:
# name (Peter), gender (m), DOB (1992/01/17)


# In[ ]:


# Problem: How many male and female employees are born in 1992?


# In[ ]:


# Solution:
df['DOB'] = pd.to_datetime(df['DOB'])
num_male_students = len(df[(df['DOB'].dt.year == 1992) & (df['gender'] == 'm')])
num_female_students = len(df[(df['DOB'].dt.year == 1992) & (df['gender'] == 'f')])


# In[ ]:


# # Exercise 2


# In[ ]:


df = pd.read_csv('scores.csv')


# In[ ]:


# Schema of Dataframes:
# Columns in df with example values:
# Stu_Name (Mike), Engineering (90), English (89), Math (92)


# In[ ]:


# Problem: Get the students with an averaged score above 90 for science subjects.


# In[ ]:


# Solution:
df['Science_Avg'] = (df['Engineering'] + df['Math']) / 2
df[df['Science_Avg'] > 90][['Stu_Name', 'Science_Avg']]


# In[ ]:


# # Exercise 3


# In[ ]:


df = pd.read_csv('geo.csv')


# In[ ]:


# Schema of Dataframes:
# Columns in df with example values:
# state (WA), capital (Seattle), population (1.4 millon)


# In[ ]:


# Problem: What is the population of California?


# In[ ]:


# Solution:
result = df[df['state'] == 'CA']['population']


# In[ ]:


# # Exercise 4


# In[ ]:


df = pd.read_csv('phones.csv')


# In[ ]:


# Schema of Dataframes:
# Columns in df with example values:
# model (Pixel 6), brand (Google), price (387), release (2022)


# In[ ]:


# Problem: What is the most expensive phone in each brand.


# In[ ]:


# Solution:
df.loc[df.groupby('brand')['price'].idxmax()][['brand', 'model', 'price']]


# In[ ]:


# # Exercise 5
\end{minted}
\end{code}

\begin{code}
\caption{\vc/ Prompt Prefix (Setup 2)}
\label{lst:prompt:vc:e2}
\begin{minted}[linenos, breaklines, firstnumber=last]{python}
# In[ ]:


import pandas as pd
import matplotlib.pyplot as plt


# In[ ]:


# You are a professional data scientist. Answer the following questions using pandas and matplotlib.


# In[ ]:


# # Exercise 1


# In[ ]:


df = pd.read_csv('employee.csv')


# In[ ]:


# Schema of Dataframes:
# Columns in df with example values:
# name (Peter), gender (m), DOB (1992/01/17)


# In[ ]:


# Problem: How many male and female employees are born in 1992?


# In[ ]:


# Solution:
df['DOB'] = pd.to_datetime(df['DOB'])
num_male_students = len(df[(df['DOB'].dt.year == 1992) & (df['gender'] == 'm')])
num_female_students = len(df[(df['DOB'].dt.year == 1992) & (df['gender'] == 'f')])


# In[ ]:


# # Exercise 2


# In[ ]:


df = pd.read_csv('scores.csv')


# In[ ]:


# Schema of Dataframes:
# Columns in df with example values:
# Stu_Name (Mike), Engineering (90), English (89), Math (92)


# In[ ]:


# Problem: Get the students with an averaged score above 90 for science subjects.


# In[ ]:


# Solution:
df['Science_Avg'] = (df['Engineering'] + df['Math']) / 2
df[df['Science_Avg'] > 90][['Stu_Name', 'Science_Avg']]


# In[ ]:


# # Exercise 3


# In[ ]:


df = pd.read_csv('geo.csv')


# In[ ]:


# Schema of Dataframes:
# Columns in df with example values:
# state (WA), capital (Seattle), population (1.4 millon)


# In[ ]:


# Problem: What is the population of California?


# In[ ]:


# Solution:
result = df[df['state'] == 'CA']['population']


# In[ ]:


# # Exercise 4


# In[ ]:


df = pd.read_csv('phones.csv')


# In[ ]:


# Schema of Dataframes:
# Columns in df with example values:
# model (Pixel 6), brand (Google), price (387), release (2022)


# In[ ]:


# Problem: What is the most expensive phone in each brand.


# In[ ]:


# Solution:
df.loc[df.groupby('brand')['price'].idxmax()][['brand', 'model', 'price']]


# In[ ]:


# # Exercise 5
\end{minted}
\end{code}

\begin{code}
\caption{\vc/ Prompt Prefix (Setup 3)}
\label{lst:prompt:vc:e3}
\begin{minted}[linenos, breaklines, firstnumber=last]{python}
# In[ ]:


import pandas as pd
import matplotlib.pyplot as plt


# In[ ]:


# You are a professional data scientist. Answer the following questions using pandas and matplotlib.


# In[ ]:


# # Exercise 1


# In[ ]:


df = pd.read_csv('olympics.csv')


# In[ ]:


# Schema of Dataframes:
# Columns in df with example values:
# Year (1896), City (Athens), Country (Greece), Nations (14)


# In[ ]:


# Problem: Which countries host at least two olympic games?


# In[ ]:


# Solution:
count_df = df['Country'].value_counts()
count_df[count_df >= 2].index.tolist()


# In[ ]:


# # Exercise 2


# In[ ]:


df = pd.read_csv('employee.csv')


# In[ ]:


# Schema of Dataframes:
# Columns in df with example values:
# name (Peter), gender (m), DOB (1992/01/17)


# In[ ]:


# Problem: How many male and female employees are born in 1992?


# In[ ]:


# Solution:
df['DOB'] = pd.to_datetime(df['DOB'])
num_male_students = len(df[(df['DOB'].dt.year == 1992) & (df['gender'] == 'm')])
num_female_students = len(df[(df['DOB'].dt.year == 1992) & (df['gender'] == 'f')])


# In[ ]:


# # Exercise 3


# In[ ]:


df = pd.read_csv('score.csv')


# In[ ]:


# Schema of Dataframes:
# Columns in df with example values:
# name (John), score (97)


# In[ ]:


# Problem: Make a new column "grade" for letter grades (A: 90+, B: 70-90, C: <70) and plot the number of students in each grade.


# In[ ]:


# Solution:
df['grade'] = df.score.apply(lambda x: 'A' if x >= 90 else ('B' if 70 <= x < 90 else 'C'))
df.grade.value_counts().plot(kind='bar')


# In[ ]:


# # Exercise 4


# In[ ]:


df = pd.read_csv('phones.csv')


# In[ ]:


# Schema of Dataframes:
# Columns in df with example values:
# model (Pixel 6), brand (Google), price (387), release (2022)


# In[ ]:


# Problem: What is the most expensive phone in each brand.


# In[ ]:


# Solution:
df.loc[df.groupby('brand')['price'].idxmax()][['brand', 'model', 'price']]


# In[ ]:


# # Exercise 5
\end{minted}
\end{code}

\begin{code}
\caption{The notebook context part of the prompt for $\intent_2$ in \cref{fig:intro:teaser_example}}
\label{lst:prompt:notebook_ctx}
\begin{minted}[linenos, breaklines, firstnumber=last, escapeinside=||]{python}
# In[ ]:


import pandas as pd

df=pd.read_csv('dataset/Gamepass_Games_v1.csv')


# In[ ]:


# Schema of Dataframes:
# Columns in df with example values:
# GAME (Mass Effect Legendary Edition), RATIO (1.87), GAMERS (84,143), COMP %


# In[ ]:


# Extract min and max (if applicable) hours as two columns


# In[ ]:


def get_avg(x):
      try:
            return float(x[0]) , float(x[1])
      except:
        return 0,0
df['min'],df['max']=zip(*df['TIME'].str.replace(" hours",'').str.strip('+').str.split("-").apply(get_avg))


# In[ ]:


df['ADDED']=pd.to_datetime(df['ADDED'],format="%


# In[ ]:


# In which year was the most played game added?


# In[ ]:
|$\triangleright$ \textcolor{red}{\textrm{Model starts prediction}}|
\end{minted}
\end{code}

\end{document}